# A Review of 315 Benchmark and Test Functions for Machine Learning Optimization Algorithms and Metaheuristics with Mathematical and Visual Descriptions

MZ Naser*[1], Mohammad Khaled al-Bashiti[1], Arash Teymori Gharah Tapeh[1], Arrsin Dadras Eslarslou[1], Ahmed Naser[2], Venkatesh Kodur[3], Rami Hawileeh[4], Jamal Abdalla[4], Nima Khodadadi[5], and Amir H. Gandomi[6]

[1]Clemson University, USA
[2]University of Guelph, Canada
[3]Michigan State University, USA
[4]American University of Sharjah, UAE
[5]University of Miami, USA
[6]University of Technology Sydney, Australia

May 2024

**Abstract**

In the rapidly evolving optimization and metaheuristics domains, the efficacy of algorithms is crucially determined by the benchmark (test) functions. While several functions have been developed and derived over the past decades, little information is available on the mathematical and visual description, range of suitability, and applications of many such functions. To bridge this knowledge gap, this review provides an exhaustive survey of more than 300 benchmark functions used in the evaluation of optimization and metaheuristics algorithms. This review first catalogs benchmark and test functions based on their characteristics, complexity, properties, visuals, and domain implications to offer a wide view that aids in selecting appropriate benchmarks for various algorithmic challenges. This review also lists the 25 most commonly used functions in the open literature and proposes two new, highly dimensional, dynamic and challenging functions that could be used for testing new algorithms. Finally, this review identifies gaps in current benchmarking practices and suggests directions for future research.

*Email: mznaser@clemson.edu; Website: `http://www.mznaser.com`

# 1 Introduction

Optimization algorithms are the backbone of modern engineering, economics, logistics, artificial intelligence (AI), and many other fields where decision-making and resource allocation under constraints are required. These are computational units within AI systems tasked with discovering the best solution among various potential answers within a given problem space. This set of potential solutions is typically represented by objective functions [1]. Here, the best is defined according to one or more criteria.

Broadly speaking, optimization can be divided into deterministic and stochastic optimization algorithms [2]. Deterministic optimization involves a set of predetermined procedures and steps aimed at finding the best solution based on the problem's inherent nature and structure. Examples of deterministic optimization methods include Newton's method, conjugate gradient methods, and gradient-based algorithms such as the gradient descent algorithm.

On the other hand, stochastic approaches involve methods that leverage randomness or probabilistic techniques to explore solutions within problem spaces with flexibility and adaptability. Examples of stochastic optimization include evolutionary algorithms, swarm intelligence, and metaheuristic algorithms. These methods introduce randomness in their search processes, allowing for exploration of a broader solution space and potentially finding novel solutions that deterministic methods might overlook.

The performance of these algorithms, however, is inherently tied to the benchmarks against which they are tested. Benchmark functions serve not only as a means to measure efficiency (in terms of computational time and resources) but also effectiveness (in terms of solution quality and reliability). They enable the comparative evaluation of algorithms under standardized conditions, fostering a deeper understanding of strengths and weaknesses. Therefore, the selection of appropriate benchmarks is crucial for the development of optimization algorithms that are both powerful and practical.

One significant challenge in the field of optimization is the lack of standardization in the selection of benchmark functions. For example, one may opt to choose benchmarks that best demonstrate their algorithm's advantages, leading to biased assessments and comparisons. This issue is compounded by the diverse nature of optimization problems, which vary greatly in terms of complexity, constraints, and dimensions. Furthermore, many benchmarks are historically derived or adopted because of their mathematical convenience rather than their relevance to practical problems.

This inconsistency in benchmark selection can lead to algorithms being ill-prepared for real-world applications, where the characteristics of problems are often far removed from those in standard benchmark tests. For example, an algorithm optimized for speed over unimodal differentiable functions might perform poorly on a complex, noisy, and discontinuous real-world problem. The misalignment between theoretical benchmarking and practical needs diminishes the utility of research outcomes, making it imperative to critically assess and categorize the available benchmark functions.

Evaluating optimization algorithms is a procedure that assesses their effectiveness and correctness. A key tool for appraising optimization algorithms is the benchmark function. Benchmark functions are mathematical expressions with specific features and optima that make them suitable for assessing the performance of different algorithms.

This paper builds on existing collective works as seen in [3]-[10] and then starts by describing over 300 benchmark and test functions and shedding light on their historical significance, mathematical and visual representation, and relevance to practical problems. Our review also explores how different benchmark functions are applied across various domains to emphasize the connection between function properties and domain-specific needs/challenges. Then, we also list the 25 most commonly used functions, propose two new complex functions, discuss current gaps in benchmarking practices, and suggest areas for future research. We hope this paper can serve as a catalog and a valuable resource for interested readers involved in developing or utilizing optimization algorithms.

# 2 Technical Definitions

Several categories can be used to classify benchmark functions[11, 12]. While a number of these categories are presented below, we first start with a catalog of technical definitions and details that will serve as a primer to this section/review. Thus, the following lists a series of notable definitions that are thought to be necessary to showcase and complement this survey. These are presented alphabetically. The readers are encouraged to review other sources for a more comprehensive discussion [13, 14].

## 2.1 Benchmark Function

A mathematically defined expression used to evaluate the performance of optimization algorithms. Benchmark functions simulate various problem landscapes and challenges (i.e. sharp peaks, deep valleys, plateaus, and deceptive directions that mislead search heuristics) that algorithms may face in real-world applications.

## 2.2 Continuity

A function is continuous if one can draw its graph without lifting the pen from the paper. In mathematical terms, a function $f : R^n \to R$ is continuous at a point $x$ if small changes in $x$ result in small changes in $f(x)$. Continuity across the function's domain means no abrupt jumps or breaks in the function's value. This feature is important because it implies that a small change in input leads to a small change in output, which in turn helps in understanding the behavior of the function near optimal points.

## 2.3 Convergence

Describes how quickly an optimization algorithm approaches the optimum solution. True rapid convergence is generally desirable (however, one must be wary of premature optimization if an algorithm converges to a local optimum too quickly without adequate exploration).

## 2.4 Differentiability

A function is differentiable if it has a derivative at each point in its domain. For functions of several variables, this extends to having partial derivatives. Differentiability can be crucial for optimization methods, especially those that rely on gradient information because the derivative provides direction and rate of change information that guides the optimization algorithm toward a minimum or maximum.

## 2.5 Dimensionality

Indicates the number of independent variables (e.g. parameters) in the function. Higher dimensionality exponentially increases the size of the search space and makes it computationally expensive to explore the space search thoroughly.

## 2.6 Fitness Landscape

A conceptual model representing the quality (fitness) of solutions across the search space.

## 2.7 Global Optimum

The best possible solution in the entire search space of the function. This is the point at which the benchmark function achieves its minimum or maximum value, depending on whether the task minimizes or maximizes the function.

## 2.8 Local Optimum

A solution better than neighboring solutions within a certain region/radius, but not necessarily the best overall. For example, the existence of multiple local optima can make it challenging for algorithms to identify the true global optimum, particularly in large, complex search spaces.

## 2.9 Modality

The number of peaks (maxima) or valleys (minima) in a function, wherein a higher modality increases the complexity of the optimization landscape to test an algorithm's ability to differentiate between various optima. For example, a function is multimodal if it has multiple local minima and maxima. This is in contrast to a unimodal function, which has only one local minimum or maximum. Multimodal functions can be particularly challenging for optimization algorithms because they present many potential traps in the form of local optima, making it difficult for algorithms to find the global optimum. This property tests an algorithm's ability to escape local minima and effectively explore the search space.

### 2.10 Noise and Stochasticity

Represents random fluctuations that affect the evaluation of a function's output. For example, noise can mimic operational or measurement errors and requires robust algorithms that distinguish between genuine improvements in solution quality and random variations.

### 2.11 Optimization Algorithm

A procedure or formula for solving a problem. In the context of this survey, an optimization algorithm specifically refers to algorithms that attempt to find the best solution from a set of possible solutions based on defined criteria for a given problem.

### 2.12 Scalability

A scalable function is one whose domain can be effectively expanded to higher dimensions while maintaining its general characteristics. For instance, a function that behaves consistently as it scales provides a reliable testbed for evaluating and comparing the performance and robustness of different algorithms.

### 2.13 Search Space

The set of all possible candidate solutions over which an optimization algorithm operates. Similar to scalability, the shape and size of the search space can significantly influence the efficiency of optimization strategies.

### 2.14 Separability

A function is separable if it can be expressed as the sum of functions, each depending on a single independent variable. That is, a function $f(x_1, x_2, \ldots, x_n)$ is separable if it can be written as $f_1(x_1) + f_2(x_2) + \cdots + f_n(x_n)$. Separable functions can be optimized on each variable independently as a means to simplify the problem and potentially reduce the computational cost.

## 3 Classification of Benchmark Functions

Benchmark functions can be classified based on their mathematical properties and the type of challenges they present to optimization algorithms. These func-tions can be classified based on their modality, continuity, separability, scalability, differentiability, and dimensionality.

## 3.1 Unimodal Functions

Unimodal functions have a single global optimum. As such, and for the most part, these functions tend to be simple and more predictable (i.e. represent problems where the path to the optimum is straightforward albeit possibly steep or flat). Unimodal functions become crucial for testing the efficiency of an algorithm in converging to a global optimum without distractions from local optima or traps [15]. These functions are often used in the early stages of algorithmic development, calibration tasks, and certain real-world problems like cost minimization, wherein the relationship between variables and the objective is direct and well-understood.

## 3.2 Multimodal Functions

Unlike unimodal functions, multimodal functions contain multiple optima as a means to introduce complexity and the challenge of local minima traps [16]. This additional complexity is adopted to examine how algorithms can differentiate between local and global optima and explore and exploit the search space adaptively. Therefore, these functions can test the robustness and versatility of algorithms. Multimodal functions are particularly useful in testing scenarios that involve metaheuristic algorithms where exploration and exploitation capabilities are critical to identifying multiple satisfactory solutions and finding the best among them.

## 3.3 Fixed-dimension and Scalable Functions

Fixed-dimension functions are those designed for optimization problems with a set number of variables. They come in handy in testing how algorithms handle dimension-specific characteristics such as variable interactions and scaling. Fixed-dimension functions can be suitable for optimization problems in fixed-configuration systems such as circuit design and parameter tuning in engineering applications [17]. On the other hand, scalable functions can adjust their complexity and dimensionality to test how algorithms perform as the size of the problem increases (and pinpoint their algorithmic limitations). This feature is fundamental for applications in machine learning, data mining, and other fields.

## 3.4 Continuous and Discrete Functions

Continuous functions allow any real number value within a range as input and are vital for problems requiring fine-tuned adjustments such as fluid dynamics simulations [18]. Discrete functions, which accept only specific separate values, are essential in scenarios like scheduling and routing where decisions are categorical (e.g., yes/no, on/off). Differentiable functions are necessary for gradient-based optimization methods that rely on derivatives to guide the search direction and non-differentiable functions challenge algorithms to find optima without this guidance (which becomes beneficial in real-world applications where abrupt changes and discontinuities occur).

### 3.5 Dynamic and Constrained Functions

The open literature also notes the category of dynamic and constrained functions [19, 20]. The former type of functions change over time or as a function of the algorithm's state, thereby mimicking environments like stock markets or evolving ecosystems. On the contrary, constrained functions include additional rules or conditions that solutions must satisfy (i.e. reflecting real-world limits such as resource constraints, physical dimensions, etc.). Thus, these functions increase the complexity and realism of the optimization challenge by testing an algorithm's ability to find optimal solutions within a defined set of constraints [21]-[23].

## 4 Benchmark and Test Functions

This section documents and reviews 315 collected benchmark and test functions (see Fig. 1). We would like to note that some of these functions include variants that have been refined or customized by various researchers for specific tasks. A few of such variants are included herein for completion, and we advise interested readers to review the following sources for more general information on such variants [3]-[11]. Toward the end of this section, we list the top 25 functions commonly seen in the reviewed works. It is worth noting that the Appendix contains a visual representation of many of the reviewed functions.

The reviewed functions are listed alphabetically. Where possible, the general forms are presented, and dedicated dimension-based forms are also presented (especially those simplified to 2D). For visualization purposes, the description of funcations' landscapes was presented in terms of 2D (i.e., $x$ and $y$ as opposed to the $x_n$ notation such as $x_0$ and $x_1$). The mathematical descriptions favored the same notation - with slight variations to comply with the style of the original documents.

### 4.1 Ackley N.1 [24, 25]

This is a nonlinear multimodal function that is formed by the superposition of an exponential component and an amplified cosine component. The function has the appearance and shape of a rough, non-linear curved surface which can be clearly regarded as a hole or peak formed by a cosine function. The Ackley function has many local optimal values and the global minimum is located at 0 and $f(x^*) = 0$. The range for this function is often used at $[-32.768, 32.768]$, $[-15, 30]$, or $[-55]$. This function is Continuous, Differentiable, Non-separable, Scalable, and Multimodal.

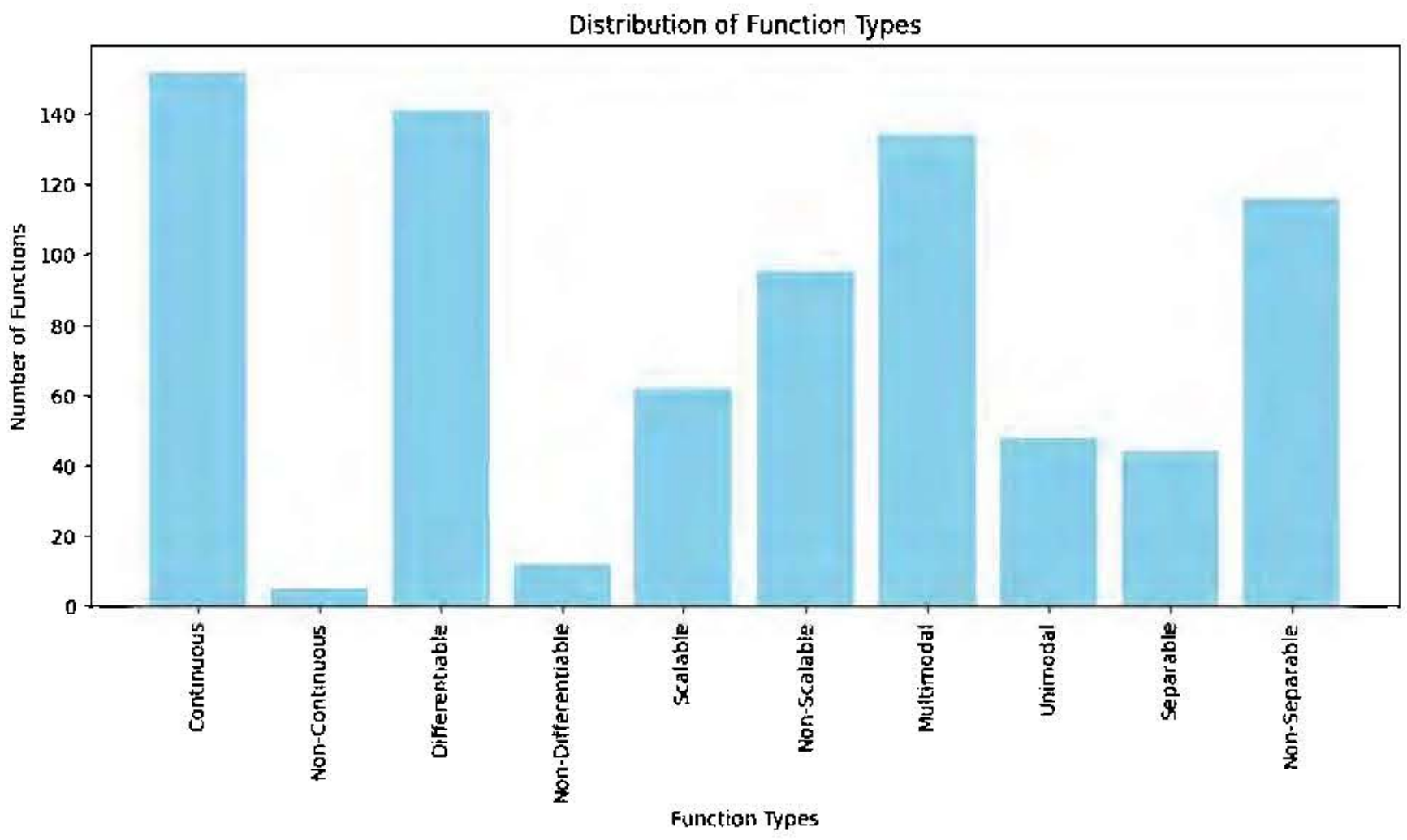

Figure 1: Classification of some of the reviewed functions

$$f(\mathbf{x}) = -a \cdot \exp\left(-b \cdot \sqrt{\left(\frac{1}{n}\sum_{i=1}^{n}\mathbf{x}_i\right)}\right) - \exp\left(\frac{1}{n}\sum_{i=1}^{n}\cos(c \cdot \mathbf{x}_i)\right) \tag{1}$$

with a = 20, b = 0.2 and c = $2\pi$

## 4.2 Ackley N.2 [23]

This is one of the commonly seen Ackley variants. This particular variant features an exponential decay influenced directly by the radial distance from the origin in a two-dimensional space. The Ackley N.2 exhibits a smooth, monotonically increasing shape as one moves away from the origin, reaching a maximum decay rate at the origin. This function has the global minimum located at 0 and $f(x^*) = -200$. The range for this function is often used at $[-32, 32]$ and $[-55]$. This function is Continuous, Differentiable, Non-separable, Scalable, and Unimodal.

$$f(\mathbf{x}) = -200e^{-0.2\sqrt{x_0^2+x_1^2}} \tag{2}$$

## 4.3 Ackley N.3

This variant combines an exponential decay based on distance with an exponential term altered by trigonometric functions (as a means to add more complexity). As one can see, the first term provides a radial decay similar to the Ackley N.2, but contributes positively to the function's value while the second term introduces additional complexity and variability that become apparent near the origin. At this point, the trigonometric values significantly influence the outcome. This function has the global minimum located at $(0.6825, -0.3607)$ and $f(x^*) = -195.6290$. This function has a range of $[-32, 32]$ and is Continuous, Differentiable, Separable, Scalable, and Unimodal.

$$f(\mathbf{x}) = -200e^{-0.2\sqrt{x_0^2+x_1^2}} - 5e^{\sin(3x_1)+\cos(3x_0)} \tag{3}$$

## 4.4 Ackley's Path [7]

Ackley's path function combines an exponential decay term based on the Euclidean distance from the origin with another exponential term that depends on the cosine of the coordinates (and happens to be adjusted by a constant shift). This combination provides a complex landscape due to the interplay between the rapidly changing cosine terms and the smooth decay of the exponential term based on distance. This variant is known for its highly multimodal roughness and multiple local minima. This function comes in various forms (including Ackley's N.1) and has a range of $[-32, 32]$.

$$f(\mathbf{x}) = -20\exp\left(-0.2\sqrt{\frac{1}{n}\sum_{i=1}^{n} x_i^2}\right) - \exp\left(\frac{1}{n}\sum_{i=1}^{n}\cos(2\pi x_i)\right) + 20 + e \tag{4}$$

## 4.5 Ackley (Modified)

A variant of the Ackley function with a range of $[-5, 5]$.

$$f(\mathbf{x}) = -20\cdot\exp\left(-0.2\cdot\sqrt{0.5\cdot\left(\left(x_1^2+x_2^2\right)^{0.5} + 0.3\cdot\cos(2\pi x_1) + 0.3\cdot\cos(2\pi x_2)\right)}\right) - \exp\left(0.5\cdot\left(\cos(2\pi x_1)+\cos(2\pi x_2)\right)\right) + e + 20 \tag{5}$$

## 4.6 Adjiman [24, 25]

This function consists of an interaction between a trigonometric product and a rational term that inversely depends on $y$. As such, this function exhibits a complex behavior with oscillations arising from the trigonometric terms and a damping effect from the rational division. This function has the global minimum located at $(2.0, 1.0)$ and $f(x^*) = 0$ and ranges between $[(-1, 2), (-1, 1)]$ and $[-5, 5]$. This function is Continuous, Differentiable, Separable, Non-scalable, and Multimodal.

$$f(\mathbf{x}) = \cos(x_0)\sin(x_1) - \frac{x_0}{x_1^2+1} \tag{6}$$

## 4.7 Alpine N.1 [47]

This function combines linear growth with sinusoidal modulation. This combination makes the function's response dependent on both the magnitude and the sinusoidal properties of each element. The Alpine N.1 function is periodic due to the sine component and generally increases with increasing $x$ (due to the linear term) [26]. This function has the global minimum located at 0 and $f(x^*) = 0$. This function has a search area range of $[10, 10]^D$ and is Continuous, Differentiable, Separable, Non-scalable, and Multimodal.

$$f(\mathbf{x}) = \sum_{i=0}^{n} |x_i \sin(x_i) + 0.1x_i| \tag{7}$$

## 4.8 Alpine N.2 [27]

This is an Alpine variant with a global minimum located at 7.9171 and $f(x^*) = 2.8081n$. This function is Continuous, Differentiable, Separable, Scalable, and Multimodal.

$$f(\mathbf{x}) = \prod_{i=1}^{n} \sqrt{|x_i|} \sin(x_i) \tag{8}$$

## 4.9 Aluffi-Pentini [28]

The Aluffi-Pentini function features a quartic polynomial in $x$ and a simple quadratic term in $y$ with additional linear adjustments. The $x$ component has a more complex polynomial behavior with the potential for multiple local minima and maxima due to the quartic and quadratic interplay. The $y$ component is simpler, with a single parabolic curve. This function has the global minimum located at $(-1.0465, 0)$ and $f(x^*) = -0.3523$ and ranges from $[-10, 10]$. This function is also known as the Zirilli's function.

$$f(\mathbf{x}) = 0.25x_0^4 - 0.5x_0^2 + 0.1x_0 + 0.5x_1^2 \tag{9}$$

## 4.10 Attractive Sector [29, 30]

This unimodal function selects a piecewise function with distinct behaviors depending on the values. The search area for this area ranges between $[-5, 5]$.

$$f(\mathbf{x}) = T_{osz}\left(\sum_{i=1}^{D}(s_i z_i)^2\right)^{0.9} + f_{opt} \tag{10}$$

where:

$$z = QA^{10R(x - x^{opt})}$$

$$s_i = \begin{cases} 10^2 & \text{if } x_i > x_i^{opt} \\ 1 & \text{otherwise} \end{cases}$$

## 4.11 Axis Parallel Hyper-Ellipsoid

This is a simple unimodal radial function with a constant downward shift. This function computes the sum of the squares of $x$ and $y$ and subtracts a constant which is the product of the squares of the first two positive integers (i.e. $1^2 \times 2^2 = 4$). This function is also known as the weighted sphere model function. This function has the global minimum located at 0 and $f(x^*) = 0$ and ranges from $[-5, 5]$.

$$f(\mathbf{x}) = \sum_{i=1}^{n} (i \cdot x_i^2) \tag{11}$$

## 4.12 Bartels Conn [48]

This function combines an absolute quadratic form with the absolute values of $x$ and $y$ sinusoidal functions. In this function, all terms contribute positively to algebraic growth moderated by the linear and periodic trigonometric elements. The complexity and variation in this function arise near the origin. The minimum occurs at points where the quadratic term is minimized and the trigonometric terms are at their minimum. The search space is at $[-500, 500]$. This function is Continuous, Non-differentiable, Separable, Scalable, and Multimodal.

$$f(\mathbf{x}) = \left|x_0^2 + x_1^2 + x_0 x_1\right| + \left|\sin(x_0)\right| + \left|\cos(x_1)\right| \tag{12}$$

## 4.13 Beale [7, 8]

The Beale function features a series of polynomials that increase in complexity by incorporating higher powers of $y$. The function's value is driven by the deviations of $x$ and $y$ from conditions that nullify each squared term. The global minimum is located at $(3, 0.5)$. The search space is noted at $[-4.5, 4.5]$ and $[-10, 10]$. This function is Continuous, Differentiable, Non-separable, Non-scalable, and Multimodal.

$$f(\mathbf{x}) = \left(x_0 x_1 - x_0 + \frac{3}{2}\right)^2 + \left(x_0 x_1^2 - x_0 + \frac{9}{4}\right)^2 + \left(x_0 x_1^3 - x_0 + \frac{21}{8}\right)^2 \tag{13}$$

## 4.14 Beale with Noise

A variant of Beale that introduces additional noise. The search space is noted at $[-4.5, 4.5]$.

$$f(\mathbf{x}) = 2.25\left(0.66667x_0x_1 - 0.66667x_0 + 1\right)^2 + 5.0625\left(0.44444x_0x_1^2 - 0.44444x_0 + 1\right)^2 + 6.89062\left(0.38095x_0x_1^3 - 0.38095x_0 + 1\right)^2 + \frac{1}{2} \quad (14)$$

## 4.15 Becker Lago with decay [31]

The Becker Lago function combines the square roots of the absolute values of $x$ and $y$ with an exponential decay term dependent on the squares of $x$ and $y$. For this, this function shows a mixture of mild growth near the origin (due to the root terms) and rapid decay (as $x$ and $y$ move away from the origin due to the exponential term). The search space is reported at $[-10, 10]$ and the global minimum is located at $(\pm 5, \pm 5)$ and $f(x^*) = 0$.

$$f(\mathbf{x}) = ((|x_0| - 5)^2 + (|x_1| - 5)^2)e^{-x_0^2 - x_1^2} \quad (15)$$

## 4.16 Bent Cigar [32]

This is a unimodal and nonseparable function. The function scales $y^2$ term significantly higher than $x^2$ which creates a steep valley along the $y$-axis. The landscape is dominated by the large coefficient on the $y^2$ term (which also makes it extremely sensitive to changes in $y$). The function has a parabolic shape and a minimum valley running along the $x$-axis with its minimum located at $(x, y) = (0, 0)$. The search space is reported at $[-100, 100]$.

$$f(\mathbf{x}) = x_1^2 + 1000000.0\sum_{i=1}^{n}(x_i^2) \quad (16)$$

## 4.17 Bent Identity

This Bent Identity function adds a linear radial distance term to a quadratic term based on the same variables as a means to balance the linear and quadratic growth rates. This allows the function to increase as a square root and quadratically from the origin, leading to a rapid increase in function values with increasing the distance from the origin. The minimum value of the function is at the origin where it equals zero and increases in all directions away from the origin. A possible search space can be identified as $[-5, 5]$.

$$f(\mathbf{x}) = \left(x_0^2 + x_1^2\right)^{0.5} + x_0^2 + x_1^2 \quad (17)$$

## 4.18 Biggs EXP2 [33, 34]

This exponential decay function is modulated by a time variable $t$ to model time series data. The terms involve fitting $x$ and $y$ to minimize the deviation from a given time series $y_i$. The function tests how well $x$ and $y$ can be tuned to match the decay rates in $y_i$. The search space for this function can be $[0, 20]$ and the global minimum is located at $(1, 10)$ and $f(x^*) = 0$. This function is Continuous, Differentiable, Non-separable, Non-scalable, and Multimodal.

$$f(\mathbf{x}) = \sum_{i=1}^{10} \left(e^{-t_i x_1} - 5e^{-t_i x_2} - y_i\right)^2 \tag{18}$$

where $\mathrm{t}_i = 0.1i,\ y_i = e^{-t_i} - 5e^{-10t_i}$

## 4.19 Biggs EXP3 [34]

This is Biggs EXP variant with a search area of $[0, 20]$ and the global minimum is located at $(1, 10, 5)$ and $f(x^*) = 0$. This function is Continuous, Differentiable, Non-separable, Non-scalable, and Multimodal.

$$f(\mathbf{x}) = \sum_{i=1}^{10} \left(e^{-t_i x_1} - 3x_i e^{-t_i x_2} - y_i\right)^2 \tag{19}$$

where $\mathrm{t}_i = 0.1i,\ y_i = e^{-t_i} - 5e^{-10t_i}$

## 4.20 Biggs EXP4 [34]

This is a Biggs EXP variant with a search area of $[0, 20]$ and the global minimum is located at $(1, 10, 1, 5)$ and $f(x^*) = 0$. This function is Continuous, Differentiable, Non-separable, Non-scalable, and Multidimensional.

$$f_{13}(\mathbf{x}) = \sum_{i=1}^{10} \left(x_3 e^{-t_i x_1} - x_4 e^{-t_i x_2} - y_i\right)^2 \tag{20}$$

where $\mathrm{t}_i = 0.1i,\ y_i = e^{-t_i} - 5e^{-10t_i}$

## 4.21 Biggs EXP5 [34]

This is a Biggs EXP variant with a search area of $[0, 20]$ and the global minimum is located at $(1, 10, 1, 5, 4)$ and $f(x^*) = 0$. This function is Continuous, Differentiable, Non-separable, Non-scalable, and Multimodal.

$$f(\mathbf{x}) = \sum_{i=1}^{13} \left(x_3 e^{-t_i x_1} - x_4 e^{-t_i x_2} + x_6 e^{-t_i x_5} - y_i\right)^2 \tag{21}$$

where $\mathrm{t}_i = 0.1i,\ y_i = e^{-t_i} - 5e^{10t_i} + 3e^{-4t_i}$

## 4.22 Bird [8]

This function combines sinusoidal and exponential components sensitive to each other's outputs to further create a complex interaction between $x$ and $y$. Such non-linear interdependencies with a squared term make the function's landscape potentially rich with local minima and maxima. A possible search area is $[-2\pi, 2\pi]$ and the global minimum is located at $(4.7010, 3.1529)$ and $f(x^*) = -106.7645$. This function is Continuous, Differentiable, Non-separable, Non-scalable, and Multimodal.

$$f(\mathbf{x}) = (x_0 - x_1)^2 + e^{(1-\sin(x_0))^2}\cos(x_1) + e^{(1-\cos(x_1))^2}\sin(x_0) \tag{22}$$

## 4.23 Bird with Noise

This is a variant of the Bird function that adds noise.

$$f(\mathbf{x}) = (x_0 - x_1)^2 + e^{(1-\sin(x_0))^2}\cos(x_1) + e^{(1-\cos(x_1))^2}\sin(x_0) + \frac{1}{2} \tag{23}$$

## 4.24 Bohachevsky N.1 [68, 35]

The following three functions come from the Bohachevsky family, wherein all have a similar bowl shape. The quadratic components dictate the overall growth while the cosine terms create oscillations that may lead to multiple local minima depending on the frequency and amplitude of these terms. The search area for the Bohachevsky N.1 is reported at $[-100, 100]$, $[-50, 50]$, and $[-15, 15]$ and the global minimum is located at 0 and $f(x^*) = 0$. This function is Continuous, Differentiable, Separable, Non-scalable, and Multimodal.

$$f(\mathbf{x}) = \sum_{i=1}^{n-1}\left(x_i^2 + 2x_{i+1}^2 - 0.3\cos(3\pi x_i) - 0.4\cos(4\pi x_{i+1}) + 0.7\right) \tag{24}$$

## 4.25 Bohachevsky N.2 [35]

A second variant from the Bohachevsky family. This function has a search area at $[-100, 100]$ and $[-50, 50]$ and the global minimum is located at 0 and $f(x^*) = 0$. This function is Continuous, Differentiable, Non-separable, Non-scalable, and Multimodal.

$$f(\mathbf{x}) = x_1^2 + 2x_2^2 - 0.3\cos(3\pi x_1)\cos(4\pi x_2) + 0.3 \tag{25}$$

## 4.26 Bohachevsky N.3 [35]

A third variant from the Bohachevsky family. This function has a search area at $[-100, 100]$ and $[-50, 50]$ and the global minimum is located at 0 and $f(x^*) = 0$.

This function is Continuous, Differentiable, Non-separable, Non-scalable, and Multimodal.

$$f(\mathbf{x}) = x_1^2 + 2x_2^2 - 0.3\cos(3\pi x_1 + 4\pi x_2) + 0.3 \tag{26}$$

### 4.27 Booth [36]

The Booth function belongs to the plate family. The Booth function forms a quadratic basin where the function is minimized when both expressions are zero, leading to a clear global minimum. This function sums the squares of two linear expressions to find points that minimize linear distances or errors (as typical of optimization problems). This function has a search area at $[-10, 10]$ and has its global minimum located at $(1, 3)$ and $f(x^*) = 0$. This function is Continuous, Differentiable, Non-separable, Non-scalable, and Unimodal.

$$f(\mathbf{x}) = (x_0 + 2x_1 - 7)^2 + (2x_0 + x_1 - 5)^2 \tag{27}$$

### 4.28 Boothby

This function has high-degree polynomials and combines quadratic and cubic terms in a sum of squares format as a mensa to enhance its landscape's complexity with steep valleys and possibly multiple local minima. This function has a possible search space at $[-5, 5]$ and has a global minimum located at $(x, y) \approx (3.40, 0.005)$ and $f(x^*) = 13.273$.

$$f(\mathbf{x}) = \left(x_0 + x_1^3 - 7\right)^2 + \left(x_0^2 + 11x_1^2 - 11\right)^2 \tag{28}$$

### 4.29 Box Betts [8]

The Box Betts function involves differences between exponential decays of the elements of $x$, particularly focusing on the first two elements and modulation by the third element. This allows a behaviour that can be described as a series of differences between scaled squared exponential terms that highlight discrepancies. This behaviour could help test the sensitivity of algorithms to small differences in inputs. This function has a possible search space at $[-5, 5]$ and has a global minimum located at $(1, 10, 1)$ and $f(x^*) = 0$ (where $0.9 \leq x_1, x_3 \leq 1.2$, $9 \leq x_2 \leq 11.2$). This is a Multimodal function.

$$f(\mathbf{x}) = \sum_{i=0}^{D-1} g(x_i)^2, \tag{29}$$

where $\mathrm{g}(\mathrm{x}_i) = e^{-0.1(i+1)x_1} - e^{-0.1(i+1)x_2} - e^{-\left(0.1(i+1) - e^{-(i+1)}\right)x_3}$

## 4.30 Box Betts Quadratic Sum

This is a variant of the Box Betts function that includes a 2D Gaussian function over a meshgrid. The computations of this function generate a 2D matrix where each element represents the product of Gaussian evaluations at grid points, simulating a heat map or a topographical surface. This function has a possible search space at $[-5, 5]$ and has a global minimum located at $f(1, 10, 1) = 0$ (where $0.9 \leq x_1 \leq 1.2$, $9 \leq x_2 \leq 11.2$, $0.9 \leq x_3 \leq 1.2$). This function is Continuous, Differentiable, Non-separable, Non-Scalable, and Multimodal.

$$f(\mathbf{x}) = \sum_{j=1}^{10} \left(\left(e^{-0.1jx_1} - e^{-0.1jx_2}\right)\left(e^{-0.1j} - e^{-j}\right)x_3\right)^2 \tag{30}$$

## 4.31 Bradford

The Bradford function combines a power-based term (that increases with distance from the origin) with a product of trigonometric sine terms to introduce complex oscillations. This combination provides a mix of radial growth and complex oscillatory. This translates to a rich landscape with local minima and maxima. This function has a search area of $[0, 1]$.

$$f(\mathbf{x}) = \left(x_0^2 + x_1^2\right)^{0.5} + \sin^2\left(x_0^3\right)\sin^2\left(x_1^3\right) \tag{31}$$

## 4.32 Branin [8, 37]

This function includes a polynomial (quadratic and linear) term influenced by cosine modulation. The polynomial term forms valleys dependent on the values of $x$ and $y$ while the cosine term adds periodicity that impacts the locations of minima and maxima. The Branin has multiple irregularly spaced minima at $(-\pi, 12.275)$, $(\pi, 2.275)$, $(3\pi, 2.475)$ where $f(x^*) = 0.3978$ (where $-5 \leq x_1 \leq 10$ and $0 \leq x_2 \leq 15$, $x_1 \in [-5, 10]$, $x_2 \in [0, 15]$). This function is Continuous, Differentiable, Non-separable, Non-Scalable, and Multimodal.

$$f(\mathbf{x}) = \left(x_2 - \frac{5.1x_1^2}{4\pi^2} + \frac{5x_2}{\pi} - 6\right)^2 + 10\left(1 - \frac{1}{8\pi}\right)\cos(x_1) + 10 \tag{32}$$

## 4.33 Branin RCOS2 [38]

This variant combines a polynomial term that is influenced by trigonometric and logarithmic adjustments. This combination produces complex behavior especially near the origin that comprises valleys (shaped by the quadratic polynomial) modulated by the cosine and logarithmic. Similar to the Branin function, this variant has potentially multiple local minima (notably at $(-3.2, 12.53)$ where $f(x^*) = 5.55904$). This function is Continuous, Differentiable, Separable, Non-scalable, and Multimodal.

$$f(\mathbf{x}) = \left(x_2 - \frac{5.1x_1^2}{4\pi^2} + \frac{5x_1}{\pi} - 6\right)^2 + 10\left(1 - \frac{1}{8\pi}\right)\cos(x_1)\cos(x_2)\left(\ln(x_1^2 + x_2^2 + 1)\right) + 10 \quad (33)$$

## 4.34 Brent [8, 39]

The function comprises quadratic terms that construct a parabolic bowl, while an exponential term introduces a localized bump at the origin. The quadratic element primarily shapes the function and guides it towards the lower bowl's center. On the other hand, the exponential term introduces a slight peak on a smoothly curving surface. This function has its global minimum near $(-10, -10)$. The function operates within the range of $x_1, x_2 \in [-10, 10]$ achieving a minimum value of $f(x^*) = 0$. This function is Continuous, Differentiable, Non-separable, Non-scalable, and Unimodal.

$$f(\mathbf{x}) = (x_1 + 10)^2 + (x_2 + 10)^2 + e^{-x_1^2 - x_2^2} \quad (34)$$

## 4.35 Brown [8, 40]

The Brown function exhibits an extremely sensitive behavior to changes in $x$ and $y$ due to the power terms (which are also dependent on both variables). This function is symmetric and grows more complex and larger as values move away from the origin. The Brown function operates within the range of $[-1, 4]$, achieving a minimum value of $f(x^*) = 0$. This function is Continuous, Differentiable, Non-separable, Scalable, and Unimodal.

$$f(\mathbf{x}) = \sum_{i=1}^{n-1} \left(x_i^2\right)^{\left(x_{i+1}^2 + 1\right)} + \left(x_{i+1}^2\right)^{\left(x_i^2 + 1\right)} \quad (35)$$

## 4.36 Brown Almost Linear [41]

This is a linear variant of the Brown function. This variant has a search range of $[-0.5, 0.5]$ achieving a minimum value of $f(x^*) = 0$. This function is Continuous, Differentiable, Non-separable, Scalable, and Unimodal.

$$f(\mathbf{x}) = (x_0 - x_1 + 1)^2 + (x_0 + x_1 - 1)^2 \quad (36)$$

## 4.37 Brown and Dennis [54]

This function has a minima of 85822.2 at (25, 5, -5, -1).

$$f(\mathbf{x}) = (x_1 + t_{x_2}\exp\left[t_{x_3}\right] + (x_3 + x_4\sin(t_{x_5}) - \cos(t_{x_5})))^2 \quad (37)$$

where $t_{x_i} = \frac{u}{5}$ for $i = 2, 3, 5$.

## 4.38 Broyden Tridiagonal [54]

This function has a minima of zero at (-1, ..., -1).

$$f(\mathbf{x}) = (3 - 2x_i)x_{i-1} - 2x_{i+1} + 1 \tag{38}$$

where $x_0 = x_{n+1} = 0$.

## 4.39 Bukin N.2 [8]

The Bukin function contains two quadratic terms wherein one is heavily weighted and the other is lightly weighted. The heavily weighted term shapes a narrow valley while the lighter term slightly adjusts the shape and tilting of the valley. This function has a search range of $[-10, 10]$ achieving a minimum value of $f(x^*) = 0$ (where $x_1 \in [-15, -5]$, $x_2 \in [-3, 3]$). This function is Continuous, Differentiable, Non-separable, Non-scalable, and Multimodal.

$$f(\mathbf{x}) = 0.01\left(x_0 + 10\right)^2 + 100\left(x_0 - 0.01x_1^2 + 1\right)^2 \tag{39}$$

## 4.40 Bukin N.4 [8]

This is a variant of the Bukin function. This function has a search range of $[-10, 10]$ achieving a minimum value of $f(x^*) = 0$ (where $x_1 \in [-15, -5]$, $x_2 \in [-3, 3]$). This function is Continuous, Differentiable, Non-separable, Non-scalable, and Multimodal.

$$f(\mathbf{x}) = 0.01\left|x_1 + 10\right| + 100x_0^2 \tag{40}$$

## 4.41 Bukin N.6 [8]

This is a variant of the Bukin function. This function has a search range of $[-10, 10]$ achieving a minimum value of $f(x^*) = 0$ (where $x_1 \in [-15, -5]$, $x_2 \in [-3, 3]$). This function is Continuous, Differentiable, Non-separable, Non-scalable, and Multimodal.

$$f(\mathbf{x}) = 0.01\left|x_0 + 10\right| + 100\sqrt{\left|0.01x_0^2 - x_1\right|} \tag{41}$$

## 4.42 Camel Three Hump [37]

This function comprises polynomial expressions for both $x$ and $y$ to feature biquadratic and cross-product terms that yield a landscape complete with multiple local minima. The function operates within the domain $x_1, x_2 \in [5, 5]$ with minima where $f(x^*) = 0$. This function is Continuous, Differentiable, Non-separable, Non-scalable, and Multimodal.

$$f(\mathbf{x}) = \frac{x_0^6}{6} - 1.05x_0^4 + 2x_0^2 + x_0x_1 + x_1^2 \tag{42}$$

## 4.43 Camel Six Hump [33, 37]

This function integrates a higher order (sixth-degree) polynomial in $x$ coupled with a product interaction between $x$ and $y$ and complemented by a quadratic term in $y$. The substantial polynomial component in $x$ shapes a convoluted terrain with multiple local minima and maxima. The quadratic term in $y$ stabilizes the function's vertical fluctuations. The function's domain spans $x_1 \in [3, 3]$ and $x_2 \in [2, 2]$ with global minima located at $(\pm 0.0898, \pm 0.7126)$ where $f(x^*)$ reaches 1.0316. In addition, this function has two global and four local partially regular minima [42]. This function is Continuous, Differentiable, Non-separable, Non-scalable, and Multimodal.

$$f(\mathbf{x}) = \left(4x_1^2 - 4\right) x_1^2 + \left(\frac{x_0^4}{3} - 2.1x_0^2 + 4\right) x_0^2 + x_0 x_1 \tag{43}$$

## 4.44 Carrom Table [8, 39]

The Carrom Table function is characterized by a cosine variation attenuated by an exponential decay contingent on the radial distance (accompanied by a normalization constant factor). The behavior of the function can be summed by sharp fluctuations (governed by the cosine component) interconnected with an exponential decay (which intensifies towards a radius $\pi$ from the origin). Once the distance from the origin surpasses $\pi$, the function flattens as it approaches zero. The function operates within the domain $x_1$ and $x_2 \in [10, 10]$ with a minima located at $(9.6461, 9.6461)$ where $f(x^*) = -24.15682$ and another at $(0, 0)$ where $f(x^*) = -0.09061$. This function is Continuous, Differentiable, Non-separable, Non-scalable, and Multimodal.

$$f(\mathbf{x}) = \frac{1}{30} e^{-2\left|(1 - \frac{\sqrt{x_1^2 + x_2^2}}{\pi}\right|} \cos^2(x_1) \cos^2(x_2) \tag{44}$$

## 4.45 Chameleon

This function integrates a quartic term based on the radial distance from a circle of radius 2, supplemented by cosine waves in the $x$ and $y$ directions. The quartic term in this function determines a pronounced valley centered at a radius of 2 from the origin. At the same time, the cosine terms introduce periodic dynamic fluctuations along the $x$ and $y$ axes. The function is defined within the range $x_1$ and $x_2 \in [0, 16.49]$ with a notable minima located at $(0.005, 2.0)$ where $f(x^*) = 0.0003$.

$$f(\mathbf{x}) = \left(x_0^2 + x_1^2 - 4\right)^2 - 0.3 \cos(3\pi x_0) - 0.4 \cos(4\pi x_1) + 0.7 \tag{45}$$

## 4.46 Chen Bird [25]

The Chen Bird function utilizes inverse quadratic functions that create a sharp peak where the denominator is minimized. This function has two peaks deter-mined by the expressions within the squared terms. These sharp peaks occur when the linear combinations of $x$ and $y$ when $x0.4+y0.1 \approx 0$ and $2x+y1.5 \approx 0$. This function has a search area of $[-500, 500]$ and $f(-0.3888, -0.7222) = 2000$. This function is Continuous, Differentiable, Non-separable, Non-scalable, and Multimodal.

$$f(\mathbf{x}) = -\frac{0.001}{((0.001)^2 + (x_1 - 0.4x_2 - 0.1)^2)} - \frac{0.001}{((0.001)^2 + (2x_1 + x_2 - 1.5)^2)} \tag{46}$$

## 4.47 Chen V [25]

This function also features inverse quadratic terms dependent on both the sum and difference of squares of $x$ and $y$ which yield multiple focal points in the function landscape. The function has several peaks around the circles defined by $x^2 + y^2 \approx 1$ and $x^2 + y^2 \approx 0.5$ and along the lines defined by $x^2 = y^2$. This function has a search area of $[-500, 500]$ and $f(0.3888, 0.7222) = 2000$. This function is Continuous, Differentiable, Non-separable, Non-scalable, and Multimodal.

$$f(\mathbf{x}) = -\frac{0.001}{((0.001)^2 + (x_1^2 - x_2^2 + 1)^2)} - \frac{0.001}{((0.001)^2 + (x_1^2 + x_2^2 - 0.5)^2)} - \frac{0.001}{((0.001)^2 + (x_1^2 - x_2^2)^2)} \tag{47}$$

## 4.48 Chichinadze [8]

This function combines a polynomial in $x$ with cosine and sine functions to add complexity with a decaying exponential in $y$. The presence of these elements allows the generation of periodic behavior in $x$ combined with a narrow exponential well in $y$. This function has a search area of $[-30, 30]$ and $(6.1898, 0.5)$ and $f(x^*) = -42.9443$. This function is Continuous, Differentiable, Separable, Non-scalable, and Multimodal.

$$f(\mathbf{x}) = 8\sin\left(\frac{5\pi x_0}{2}\right) + 10\cos\left(\frac{\pi x_0}{2}\right) + x_0^2 - 12x_0 - 0.2^{0.5}e^{-0.5(x_1-0.5)^2} + 11 \tag{48}$$

## 4.49 Chung Reynolds [43]

This function represents a simple radial function where values grow quartically from the origin and accelerate as the distance from the origin increases. This behaviour forms a steep bowl-shaped landscape. This function has a range of $[-10, 10]$ and a global minimum at $(x, y) = (0, 0)$ with a value of 0. This function is Continuous, Differentiable, Separable, Scalable, and Unimodal.

$$f(\mathbf{x}) = \sum_{i=1}^{n-1} (x_i^2)^2 \tag{49}$$

$$f(x, for2D) = \left(x_0^2 + x_1^2\right)^2$$

### 4.50 Chung Reynolds N.2

This is a variant of the original Chung Reynolds function.

$$f(\mathbf{x}) = \left(x_0^2 + x_1^2\right)^2 - \cos(x_0)\cos(x_1) \tag{50}$$

### 4.51 Modified Cigar [8]

The Cigar function has a slight variation to the Bent Cigar function by introducing a sin. Visually, this function forms a multimodal elongated parabolic valley along the $x$-axis with very steep sides along the $y$-axis. This function has a range of $[-10, 10]$ and a global minimum at $(x, y) = (0, 0)$ with a value of 0.

$$f(\mathbf{x}) = x_1^2 + 1000000.0 \sum_{i=1}^{n} (x_i^2) + sin(x_1) \tag{51}$$

### 4.52 Clunar

This function reaches its minimum close to the origin, which aligns with the expected minimizing behavior of the polynomial term $(x^2 + y^2)^2$. The Clunar has a minimum value of approximately $8.92 \times 10^{-5}$, which occurs near the coordinates $(0.005, 0.005)$. This function has a range of $[0, 64]$.

$$f(\mathbf{x}) = \left(x_0^2 + x_1^2\right)^2 + \frac{\sin^2(3\pi x_0)}{50} + \frac{\sin^2(3\pi x_1)}{50} \tag{52}$$

### 4.53 Cola [8]

This function involves a Gaussian profile centered at the origin. The Gaussian component predominately molds the landscape, creating a depression around the origin with values escalating as one moves outward (in response to the Gaussian decay). This profile of this function is adjusted by subtracting a constant to shift its baseline. The Cola function exhibits a global minimum just outside the origin where the Gaussian value approximates 0.5. Operating within a range of $[-4, 4]$, this function has a global minimum $f(x^*) = 11.7464$. This function is Continuous, Differentiable, Separable, Non-scalable, and Multimodal.

$$f(\mathbf{x}) = \sum_{j<i} (r_{ij} - d_{ij})^2$$

where $r_{ij} = \sqrt{(x_i - x_j)^2 + (y_i - y_j)^2}$

$$\text{and } d = \begin{bmatrix} 1.27 & & & & & & & & \\ 1.69 & 1.43 & & & & & & & \\ 2.04 & 2.35 & 2.43 & & & & & & \\ 3.09 & 3.18 & 3.26 & 2.85 & & & & & \\ 3.20 & 3.22 & 3.27 & 2.88 & 1.55 & & & & \\ 2.86 & 2.56 & 2.58 & 2.59 & 3.12 & 3.06 & & & \\ 3.17 & 3.18 & 3.18 & 3.12 & 1.31 & 1.64 & 3.00 & & \\ 3.21 & 3.18 & 3.18 & 3.17 & 1.70 & 1.36 & 2.95 & 1.32 & \\ 2.38 & 2.31 & 2.42 & 1.94 & 2.85 & 2.81 & 2.56 & 2.91 & 2.97 \end{bmatrix} \tag{53}$$

## 4.54 Colville [7, 8]

This is an enhanced version of the standard Rosenbrock function and includes additional variables and cross-terms to significantly augment the complexity of its optimization landscape (with multiple minima, valleys, and ridges). The inclusion of multiple terms fosters intricate interactions that serve to penalize deviations from these terms. The function's operational range is $x_1, x_2 \in [-10, 10]$ with the global minimum posited at a value where $f(x^*) = 0$. This function is Continuous, Differentiable, Non-separable, Scalable, and Multimodal.

$$\begin{aligned} f(\mathbf{x}) = {} & 100(x_1 - x_2^2)^2 + (1 - x_1)^2 + 90(x_4 - x_3^2)^2 \\ & + (1 - x_3)^2 + 10.1((x_2 - 1)^2 + (x_4 - 1)^2) + 19.8(x_2 - 1)(x_4 - 1) \end{aligned} \tag{54}$$

$$f(x, for2D) = (1 - x_0)^2 + 100\left(x_0 - x_1^2\right)^2 + 10.1\left(x_1 - 1\right)^2$$

## 4.55 Composite Griewank Rosenbrock F8F2 [44]

This variant of the classic Rosenbrock function is modified through scaling down and the inclusion of a cosine term to modulate its landscape. By scaling down, the function's steepness (which is a resembleness of the Rosenbrock function) is diminished while the introduced cosine element produces periodic fluctuations that significantly influence the overall topography of the function. This interplay between the parabolic Rosenbrock elements and the oscillatory nature of the cosine element complicates the identification of exact minima. This multimodal function operates within the domain of $[-5, 5]$ and has multiple minima.

$$f(\mathbf{x}) = 10\left(\frac{\sum_{i=1}^{D-1}\left(\frac{s_i}{4000} - \cos(s_i)\right)}{D - 1}\right) + 10 + f_{\text{opt}} \tag{55}$$

where

$$\begin{aligned} z &= \max\left(1, \frac{1}{\sqrt{8}}\right) Rx + 0.5, \\ s_i &= 100(z_i^2 - z_{i+1})^2 + (z_i - 1)^2 \text{ for } i = 1, \ldots, D, \\ z_{\text{opt}} &= 1. \end{aligned}$$

$$f(x, for2D) = \frac{(1 - x_0)^2}{4000} + \frac{(-x_0^2 + x_1)^2}{40} - \cos\left(\frac{\sqrt{2}x_1}{2}\right)\cos(x_0) + 1$$

### 4.56 Composition Function N.1

This function joins three benchmark functions—Sphere, Rastrigin, and Weierstrass—through specified weights. The smoothness of the Sphere function establishes a baseline of simplicity in the overall topology. Then the Rastrigin function introduces its notorious highly local minima that aim to significantly complicate the search landscape with its sharp, frequent oscillations. The Weierstrass function adds another layer of complexity. The resultant combination of these functions creates a diverse and challenging optimization landscape where the specific composition and weighting of the constituent functions highly influence the nature of the minima. Operating within the domain $[-5, 5]$, the exact location and value of the minima vary, reflecting the dynamic interplay and balance among the integrated functions' smooth, rugged, and complex features.

$$f(\mathbf{x}) = 0.3 \cdot \text{Sphere}(\mathbf{x}) + 0.4 \cdot \text{Rastrigin}(\mathbf{x}) + 0.3 \cdot \text{Weierstrass}(\mathbf{x}) \tag{56}$$

### 4.57 Composition Function N.2

This function builds on the aforenoted Composition function N.1 but with adjusted weights and the incorporation of the Griewank function instead of the Sphere. This function provides a different balance, slightly tilting towards the fractal nature of the Weierstrass function with a search space of $[-5, 5]$.

$$f(\mathbf{x}) = 0.3 \cdot \text{Griewank}(\mathbf{x}) + 0.3 \cdot \text{Rastrigin}(\mathbf{x}) + 0.4 \cdot \text{Weierstrass}(\mathbf{x}) \tag{57}$$

### 4.58 Corana [22, 45]

This function is dominated by the sixth power of both variables, which forms a very steep multimodal landscape. The addition of the linear term in $x$ breaks symmetry along the $x$-axis. The Corana function is Scalable and Multimodal. It has a range of $[-100, 100]$ and is minimized near zero (i.e. 0 and $f(x^*) = 0$).

$$f(\mathbf{x}) = \begin{cases} 0.15(z_i - 0.05\,\mathrm{sgn}(z_i^2))d_i & \text{if } |v_i| < A \\ d_i x_i^2 & \text{otherwise} \end{cases} \tag{58}$$

where

$$v_i = |x_i - z_i|, \qquad A = 0.05$$
$$z_i = 0.2\left(\left\|\frac{x_i}{0.2}\right\| + 0.49999\right)\mathrm{sgn}(x_i)$$
$$d_i = (1, 1000, 10, 100)$$

## 4.59 Cosine Envelope Sine Wave

This function combines cosine elements with simple quadratic terms in order to form a complex oscillating landscape. The superposition of high-frequency cosine elements with a quadratic growth leads to a challenging landscape with many local minima. The range of this function is $[-10, 10]$ and has minima at $(0.005, 0.005)$ and $f(x^*) = -0.199$.

$$f(\mathbf{x}) = -0.1\cos(5\pi x_0) - 0.1\cos(5\pi x_1) + x_0^2 + x_1^2 \tag{59}$$

## 4.60 Cosine Function

The minimum value of the Cosine function is approximately 0.9999, which occurs very close to the point $(\pi, \pi)$. This is where both cosine terms are maximized (since $\cos(\pi) = 1$) and the exponential term is near its peak.

$$f(\mathbf{x}) = -e^{-(x_0-\pi)^2-(x_1-\pi)^2}\cos(x_0)\cos(x_1) \tag{60}$$

## 4.61 Cosine Mixture [8]

This function evaluates a sum of cosine functions and quadratic terms across an array $x$. The combination of negative cosines and quadratic terms creates a landscape that balances between periodic local minima and a parabolic rise. This function has a range of $[-1, 1]$ and $f(x^*) = -0.1N$. This function is Non-continuous, Non-differentiable, Non-separable, and Multimodal.

$$f(\mathbf{x}) = -0.1\sum_{i=0}^{n-1}\cos(5\pi x_i) - \sum_{i=0}^{n-1} x_i^2 \tag{61}$$

## 4.62 Cosine Root

The minimum value of this function is approximately 6.283, which occurs at $(-6.2831, -6.2831)$. This point represents a location where both $x$ and $y$ are at significant values and both cosine terms are positive or negative (which maximizes the product before negation).

$$f(\mathbf{x}) = -\cos(x_0)\cos(x_1)\sqrt{|x_0 x_1|} \tag{62}$$

### 4.63 Cosinee Mixture [6]

This function incorporates a sum of cosine functions, each with a frequency multiplier to induce periodic oscillations. This function also couples a sum of squared terms to impose a penalty on larger values of $x$. The minima of this function are linked to the alignment of the peaks of the cosine functions with smaller values of $x$ (one should note that the quadratic component exerts its influence by reducing function values progressively further from zero). This function operates within the range of $[1, 1]$ and has specific minima at $f(x^*) = 0.2$ when $n = 2$ and $f(x^*) = 0.4$ when $n = 4$.

$$f(\mathbf{x}) = -0.1\sum_{i=1}^{n}\cos(5\pi x_i) - \sum_{i=1}^{n} x_i^2 \tag{63}$$

### 4.64 Cross in Tray [8, 46]

This function applies an exponential decay influenced by the radial distance from the origin modulated by the sine functions' product. A power transformation is also included to moderate the influence of large values and ensure a smoother transition in the function's landscape. The behavior of this function is characterized by gradual changes due to the smoothing effect of the exponential component. However, this function also has significant peaks or valleys emerging from the interactions between the trigonometric elements and the exponential decay. The range and behavior of the minima are predominantly governed by the negative exponential decay, which is scaled to reduce the impact of peaks typically associated with exponential functions. The function operates within the domain $[15, 15]$ and exhibits multiple local minima (e.g. $f(x^*) = -2.06261$ at $x = 1.34941, y = 1.34941$). This function is Continuous, Non-differentiable, Non-separable, Non-Scalable, and Multimodal.

$$f(\mathbf{x}) = -0.0001\left(\sin(x_1 x_2)\exp\left(100 - \sqrt{x_1^2 + x_2^2}\right)/\pi\right)^{0.1} + 1 \tag{64}$$

### 4.65 Cross Leg

This function has an extremely large and negative minimum value close to $= -9.56 \times 10^{42}$ that occurs very close to the coordinates $(0.752, 0.752)$. At these coordinates, the trigonometric components typically reduce the magnitude and the exponential growth due to the large constant term (100) in the exponent dominates, leading to an extremely negative value.

$$f(\mathbf{x}) = -\left|\sin(x_1)\cos(x_2)\exp\left(\left|100 - \frac{\sqrt{x_1^2 + x_2^2}}{\pi}\right|\right)\right| \tag{65}$$

## 4.66 Cross Leg Table [8, 46]

This multimodal function builds upon the previous function but omits the additional power scaling, which results in more pronounced effects from both the exponential and sinusoidal components. The absence of power scaling allows the exponential term to more significantly amplify the outcomes of the sinusoidal functions. Thus, this function can be characterized by sharp peaks where the interaction of these components is most intense. The inclusion of the absolute value operation ensures that all function values remain non-positive (which leads to less distinct minima). This function operates within [10, 10] with the minima being particularly challenging to isolate (i.e. $f(x^*) = -1$).

$$f(\mathbf{x}) = -\left|\sin(x_1)\sin(x_2)\exp\left(\left[100 - \left(\sqrt{x_1^2 + x_2^2}\right)^{0.5}\right]/\pi\right) + 1\right|^{-0.1} \tag{66}$$

## 4.67 Crowned Cross [8, 46]

This multimodal function integrates exponential decay modulated by sinusoidal terms, which are then scaled and offset before computing the absolute difference from a constant. The landscape of this function is portrayed by scaling and power transformations within the interaction between the trigonometric and exponential components. Due to the absolute difference term, identifying minima becomes challenging as the function targets values close to 2 after the transformation. The search space for this function resides in [10, 10]. The function presents a minimization scenario where the global minimum is close to zero, specifically $f(x^*) = 0.0001$.

$$f(\mathbf{x}) = 0.0001\left(e^{\left|100 - \frac{\sqrt{x_0^2 + x_1^2}}{\pi}\right|}\left|\sin(x_0)\sin(x_1)\right| + 1\right)^{0.1} \tag{67}$$

## 4.68 Csendes [8, 47]

This function applies a sixth power polynomial to each element in $x$ and introduces a sinusoidal perturbation that inversely depends on each element. The behavior of the function varies significantly near $x = 0$ as the sine function becomes highly oscillatory (implying that this function becomes particularly challenging for algorithms to optimize around zero). This function is also known for its multimodal nature and challenges presented by the sixth power term, making it sensitive to changes and small variations in $x$. This function has a range of $[-1, 1]$ and a global minimum at $(x, y) = (0, 0)$ with $f(x^*) = 0$.

$$f(\mathbf{x}) = \sum_{i=0}^{n-1}\left(\sin\left(\frac{1}{x_i}\right) + 2\right)x_i^6 \tag{68}$$

## 4.69 Cube

The Cube function integrates a cubic term that introduces non-linear growth in all directions from the origin. The search space is $[-10, 10]$ with $f(x^*) = 0$ at $(-1, 1)$.

$$f(\mathbf{x}) = 100\left(x_1 - x_0^3\right)^2 + (1 - x_0)^2 \tag{69}$$

## 4.70 Cubic ten

This function evaluates the cubic power of the absolute difference between the variables and 10. The function is symmetric about $x = 10$ and $y = 10$, resulting in a function that is mirrored across these values. This function has a search space of $[-10, 10]$, with $(10, 10)$ and $f(x^*) = 0$.

$$(|x_0| - 10)^3 + (|x_1| - 10)^3 \tag{70}$$

## 4.71 Damavandi [48]

The Damavandi function has a combination of power and sinusoidal terms with the search space of $[-5, 5]$ and $[0, 14]$.

$$f(\mathbf{x}) = \left(1 - |\frac{\left|\frac{\sin(\pi(x_0-2))\sin(\pi(x_1-2))}{(x_0-2)(x_1-2)}\right|}{\pi^2}|\right)^5 \left((x_0 - 7)^2 + 2(x_1 - 7)^2 + 2\right) \tag{71}$$

## 4.72 Damavandi N.2

This is a variant of the Damavandi function.

$$f(\mathbf{x}) = \left(1 - \frac{\sin(\pi(x_0 - 2))\sin(\pi(x_1 - 2))}{\pi^2(x_0 - 2)(x_1 - 2)}\right)^2 \left((x_0 - 7)^2 + 2(x_1 - 7)^2 + 2\right) \tag{72}$$

## 4.73 De Jong [3]

There are a series of functions developed by De Jong. This particular function is based on yields a paraboloid shape arising from integrating individual squared terms of $x$ and $y$. This function is designed to have a global minimum at the origin, $(x, y) = (0, 0)$, where the function value reaches 0. This function has a commonly used range of $[-5.12, 5.12]$. This is a continuous and unimodal function. This function is also known as the Sphere.

$$f(\mathbf{x}) = \sum_{i=1}^{n} x_i^2 \tag{73}$$

## 4.74 De Jong modified

This particular variant of De Jong is based on yields a paraboloid shape with an interaction term $xy$, embodying a simple bilinear form. The inclusion of the $xy$ term introduces a twist in the otherwise standard parabolic surface and results in a symmetric configuration about the origin in a rotated frame.

$$f(\mathbf{x}) = x_0^2 + x_0 x_1 + x_1^2 \tag{74}$$

## 4.75 De Jong N.5 [22]

One of De Jong's variants. This particular multimodal variant has a range of $[-65.536, 65.536]$. This function is also known as Shekel's fox.

$$f(\mathbf{x}) = \sum_{i=0}^{24} \frac{1}{i + \left( x_0 - \begin{bmatrix} -32 & -16 \\ -16 & -16 \\ -32 & -8 \\ -16 & -8 \\ -32 & 0 \\ -16 & 0 \\ -32 & 8 \\ -16 & 8 \\ -32 & 16 \\ -16 & 16 \\ 0 & -16 \\ 0 & -8 \\ 0 & 0 \\ 0 & 8 \\ 0 & 16 \\ 16 & -16 \\ 16 & -8 \\ 16 & 0 \\ 16 & 8 \\ 16 & 16 \\ 32 & -16 \\ 32 & -8 \\ 32 & 0 \\ 32 & 8 \\ 32 & 16 \end{bmatrix}_{i,0} \right)^6 + \left( x_1 - \begin{bmatrix} -32 & -16 \\ -16 & -16 \\ -32 & -8 \\ -16 & -8 \\ -32 & 0 \\ -16 & 0 \\ -32 & 8 \\ -16 & 8 \\ -32 & 16 \\ -16 & 16 \\ 0 & -16 \\ 0 & -8 \\ 0 & 0 \\ 0 & 8 \\ 0 & 16 \\ 16 & -16 \\ 16 & -8 \\ 16 & 0 \\ 16 & 8 \\ 16 & 16 \\ 32 & -16 \\ 32 & -8 \\ 32 & 0 \\ 32 & 8 \\ 32 & 16 \end{bmatrix}_{i,1} \right)^6} + 0.002 \tag{75}$$

## 4.76 Deb N.1 [4, 8]

This function features a high-power sinusoidal function and applies it to each element of $x$ and then averages the results. The use of a heightened power in the sine function leads to pronounced oscillations, which accentuate the peaks more sharply compared to the troughs. As one can see, the behavior of the function is largely influenced by the properties of $\sin(5\pi x)$, which reaches its maximum values when all elements of $x$ align. The identification of the global minimum is heavily dependent on the specific distribution of $x$ values. This function operates within $[-1, 1]$ and can exhibit multiple local minima, with a global minimum where $f(x^*) = -1$. This function is Continuous, Differentiable, Separable, Scalable, and Multimodal.

$$f(\mathbf{x}) = -\frac{\sum_{i=0}^{n} \sin^6\left(5\pi x_i\right)}{n} \tag{76}$$

## 4.77 Deb N.3 [8]

This is a variant of the Deb function with a range of $[0, 1]$ and multiple minima (0 and $f(x^*) = 0$). This function is Continuous, Differentiable, Separable, Scalable, and Multimodal.

$$f(\mathbf{x}) = \frac{\sum_{i=0}^{n-1} \sin^6\left(\pi\left(5x_i^{0.75} - 0.25\right)\right)}{n} \tag{77}$$

## 4.78 Deb N.4

This is a variant of the Deb function with a range of $[0, 1]$ and multiple minima.

$$f(\mathbf{x}) = -e^{0.5 \sin\left(\frac{\pi x_1}{16}\right) \cos\left(\frac{\pi x_0}{12}\right)} + e + 20 - 20e^{-0.2\sqrt{0.06x_0^2 + 0.015x_1^6 + 1.0}} \tag{78}$$

## 4.79 Deckkers Aarts [6]

This function is constructed with polynomial terms ranging from quadratic to more aggressive quartic, each scaled significantly to shape the optimization landscape. It begins with a shallow quadratic rise, and as the values of $x$ and $y$ increase, the higher-power quartic terms introduce complexity by creating multiple wells and barriers. The range of the function extends over $[-20, 20]$. The local minima are influenced by how the positive and negative powers in the polynomial terms interact. A particularly notable minimum occurs at $(0, 15)$ with $f(x^*) = -24777$. This function is Continuous, Differentiable, Non-separable, Non-scalable, and Multimodal.

$$f(\mathbf{x}) = 1.0 \cdot 10^{-5}\left(x_0^2 + x_1^2\right)^4 - \left(x_0^2 + x_1^2\right)^2 + 100000x_0^2 + x_1^2 \tag{79}$$

### 4.80 Deflected Corrugated Spring [25]

This function integrates a radial distance measure, which is then adjusted by a cosine function designed to influence the decay across the function's landscape. Specifically, while the overall value of the function decreases with increasing distance from the origin, it also exhibits oscillations due to the cosine squared term. This leads to a dynamic landscape where the decay is not smooth. The function has a range within the domain $[-5, 5]$. A notable minimum is at $f(x^*) = -1$.

$$f(\mathbf{x}) = 0.1 \sum_{i=1}^{n} (x_i - \alpha)^2 - \cos\left(K \sqrt{\sum_{i=1}^{n} (x_i - \alpha)^2}\right) \tag{80}$$

where, *alpha* and $K = 5$ (by default)

### 4.81 Devillers Glasser

This function has a peak at the origin and decreases as $x$ or $y$ increases. The value of the function approaches zero as $x$ and $y$ go towards infinity. A typical range is $[-5, 5]$.

$$f(\mathbf{x}) = \frac{1}{25x_0^2 + 25x_1^2 + 1} \tag{81}$$

### 4.82 Devilliers Glasser N.1 [8]

A variant of the Devilliers Glasser with a range of $[-500, 500]$ and $f(x^*) = 0$. This function is Continuous, Differentiable, Non-separable, Non-scalable, and Multimodal.

$$f(\mathbf{x}) = \sum_{i=1}^{24} \left[x_1 x_2^{t_i} \sin(x_3 t_i + x_4) - y_i\right]^2 \tag{82}$$

where $\quad t_i = 0.1(i-1) \quad$ and $\quad y_i = 60.137(1.371^i) \sin(3.112t_i + 1.761)$

### 4.83 Devilliers Glasser N.2 [8]

A variant of the Devilliers Glasser with a range of $[1, 60]$ and $f(x^*) = 0$. This function is Continuous, Differentiable, Non-separable, Non-scalable, and Multimodal.

$$f(\mathbf{x}) = \sum_{i=1}^{24} \left[x_1 x_2^{t_i} \tanh\left(x_3 t_i + \sin(x_4 t_i)\right) \cos\left(t_i e^{x_5}\right) - y_i\right]^2 \tag{83}$$

where $\quad t_i = 0.1(i-1) \quad$ and
$y_i = 53.81(1.27t_i) \tanh(3.012t_i + \sin(2.13t_i)) \cos(e^{0.507t_i})$

## 4.84 Discus [49]

The Discus function is a quadratic unimodal function with smooth local irregularities where a single direction in the search space can be much more sensitive than others. The suggested search area falls within the hypercube $[-100, 100]$D. The global minimum is $f(x^*) = 0$.

$$f(\mathbf{x}) = 10^6 z_1^2 + \sum_{i=2}^{D} z_i^2 + f_{\text{opt}} \tag{84}$$

where $\mathrm{z} = \mathrm{T}_{\mathrm{osz}}(R(x - x^{\mathrm{opt}}))$

## 4.85 Dixon Coles

This function combines shifts and nonlinear transformations of $x$ and $y$, with each term squared to ensure non-negativity. The function also penalizes deviations from specific configurations, particularly near $x = 1$ and $y = -1$, as these lead to minima near these specific points. This function operates within the domain $[-10, 10]$.

$$f(\mathbf{x}) = (x_0 - 1)^2 + \left(-x_0^2 + x_0\right)^2 + (x_1 + 1)^2 + \left(x_1^2 + x_1\right)^2 \tag{85}$$

## 4.86 Dixon Price [8, 50]

This function is an extended version of the Rosenbrock function. This function comprises a sequence of quadratic terms, each influenced by the preceding terms in the vector, creating a highly dependent and interconnected topography. The behavior of this function can be described as a narrow, curved valley that leads to the global minimum. The function's range encompasses $[-10, 10]$, within which the global minimum is located at $x = (1, 1, \ldots, 1)$, where the function value reaches 0. This function is Continuous, Differentiable, Non-separable, Scalable, and Unimodal.

$$f(\mathbf{x}) = (x_0 - 1)^2 + \sum_{i=2}^{n} i\left(2x_i^2 - x_{i-1}\right)^2 \tag{86}$$

## 4.87 Dixon Price N.2

A variant of the Dixon Price with a range of $[-10, 10]$.

$$f(\mathbf{x}) = 2\left(-x_0 + 2x_1^2\right)^2 + (x_0 - 1)^2 \tag{87}$$

## 4.88 Dixon Price N.3

A variant of the Dixon Price with a range of $[-10, 10]$.

$$f(\mathbf{x}) = 2\left(-x_0 + 2x_1^2\right)^2 + (x_0 - 1)^2 \tag{88}$$

## 4.89 Dixon Price N.4

A variant of the Dixon Price with a range of $[-10, 10]$.

$$f(\mathbf{x}) = 3\left(-x_0 + 2x_1^2\right)^2 + (x_0 - 1)^2 + (x_1 - 1)^2 \tag{89}$$

## 4.90 Dixon Price N.5

A variant of the Dixon Price with a range of $[-10, 10]$ and a minima at 0.

$$f(\mathbf{x}) = \left(-x_0 + 2x_1^2\right)^2 + (x_0 - 1)^2 \tag{90}$$

## 4.91 Dixon-Price-Rosenbrock's Function

A variant of the Dixon Price with a range of $[-30, 30]$.

$$f(\mathbf{x}) = \sum_{i=1}^{n-1} \left[100(x_{i+1} - x_i^2)^8 + (x_i - 1)^8\right] \tag{91}$$

## 4.92 Dolan [8, 25]

This function combines sine and cosine functions with other elements' amplitude and overall behavior. The integration of hyperbolic tangents with the oscillatory nature of sine and cosine results in a function that exhibits complex behavior with many local extrema. The function's domain extends across $[-100, 100]$. These may include $(8.3904, 4.8142, 7.3457, 68.8824, 3.8547)$ with $f(x^*) = -1$ and $(7.8102)$ with $f(x^*) = -703.72878$. This function is Continuous, Differentiable, Non-separable, Non-Scalable, and Multimodal.

$$f(\mathbf{x}) = (x_1 + 1.7x_2)\sin(x_1) - 1.5x_3 - 0.1x_4\cos(x_4 + x_5 - x_1) + 0.2x_5^2 - x_2 - 1 \tag{92}$$

## 4.93 Drop Wave [8]

The Drop Wave is a multimodal function that incorporates a cosine function that modulates based on the radial distance from the origin and is divided by a quadratic expression in $x$ and $y$. This results in a function with a periodic component that decays as the distance from the origin increases, with the valleys and peaks of the cosine function being moderated by the growing denominator. The minimum value of the function occurs at points where the cosine term reaches its minimal value, typically where the cosine function aligns with the valleys in its periodic pattern. The suggested search area for this function is the hypercube $[-5.12, 5.12]$D, and the global minimum, $f(0) = 0$.

$$f(\mathbf{x}) = \frac{-1 + \cos\left(12\sqrt{x_0^2 + x_1^2}\right)}{0.5(x_0^2 + 0.5x_1^2) + 2} \tag{93}$$

### 4.94 Drop Wave N.2

A variant of the Drop Wave function with a range of $[-5.12, 5.12]$.

$$f(\mathbf{x}) = \frac{-1 + \cos\left(12\sqrt{x_0^2 + x_1^2}\right)}{0.5x_0^2 + 0.5x_1^2 + 2} \tag{94}$$

### 4.95 Dynamic Deceptive Basin

This is a newly proposed function to be discussed in detail in the next section.

$$f(\mathbf{x}) = \sin(x_0 + \theta_0) \cdot \cos(x_1 + \theta_1) \cdot e^{-\|\mathbf{x}-\boldsymbol{\theta}\|^2} + \text{noise} \tag{95}$$

Where:

x=[x$_0, x_1, \ldots, x_{n-1}$] represents the input vector.
$\theta = [\theta_0, \theta_1, \ldots, \theta_{n-1}]$ represents the dynamically updating state parameters.

Noise is a normally distributed random variable, noise $\sim \mathcal{N}(0, 0.1)$.

### 4.96 Complex Dynamic Deceptive Basin

This is a newly proposed function to be discussed in detail in the next section.

$$f(\mathbf{x}) = \sum_{i=0}^{n-1} \theta_i^3 x_i^3 + \sum_{i=0}^{n-1} \sin(x_i + \theta_i) + \sum_{i=0}^{n-1} \theta_i x_i^{\theta_i + n} + \sum_{i=0}^{n-1} 0.1|x_i|^{2\theta_i} + \text{noise} \tag{96}$$

Where:

x=[x$_0, x_1, \ldots, x_{n-1}$] represents the input vector.
$\theta = [\theta_0, \theta_1, \ldots, \theta_{n-1}]$ represents the dynamically updating state parameters.

Noise has an amplitude dependent on the norm of x - sin(recent effect) and is normally distributed, noise $\sim \mathcal{N}(0, \text{noise amplitude})$.

### 4.97 Easom [7,51]

This function combines cosine functions with an exponential function centered at $(\pi, \pi)$. The cosine function's periodic nature is modulated by the Gaussian envelope, which is most concentrated at its center. At this point, both the cosine components reach their peak values simultaneously. The range of the function spans multiple domains: $[-100, 100]$, $[-40, 40]$, and $[-10, 10]$. As $x$ and $y$ deviate from $\pi$, the function's value approaches zero. The function's maximum, or the least negative value, is strategically located at $(\pi, \pi)$ with $f(x^*) = -1$. This function is Continuous, Differentiable, Separable, Scalable, and Multimodal.

$$f(\mathbf{x}) = -e^{-(x_0-\pi)^2-(x_1-\pi)^2} \cos(x_0) \cos(x_1) \tag{97}$$

## 4.98 Easom with Noise

A variant of the Easom function with noise that contains noise and has a range of $[-100, 100]$.

$$f(\mathbf{x}) = -e^{-(x_0-\pi)^2-(x_1-\pi)^2} \cos(x_0) \cos(x_1) + \frac{1}{2} \tag{98}$$

## 4.99 Egg Box

This function involves a highly nonlinear transformation incorporating periodic elements, power transformation, and inversion to create a challenging landscape. The periodic nature of the function features repeating patterns across the plane, characterized by sharp troughs that correspond to the maximal values of the product of cosine terms. The range of the function is specified as $[-243, -1]$, and the minima occur at points where $x/2 = y/2 = 0, 2\pi, 4\pi$, etc., with the function value reaching as low as $-243$.

$$f(\mathbf{x}) = -\left(\cos\left(\frac{x_0}{2}\right) \cos\left(\frac{x_1}{2}\right) + 2\right)^5 \tag{99}$$

## 4.100 Egg Crate [25]

This function merges standard quadratic terms with sinusoidal penalties, where the penalties increase the function value based on the sine squared values of $x$ and $y$. This integration creates a landscape that exhibits continuous, quadratic growth as the baseline, while the sinusoidal penalties introduce periodic fluctuations. The function is defined over the domain $[-5, 5]$, with the global minimum occurring at $x = (0, 0)$, where the function value reaches $f(x^*) = 0$. This minimum is reached when the sinusoidal components contribute minimally to the function's value, thus reverting to the pure quadratic behavior at the origin.

$$f(\mathbf{x}) = 25 \sin^2(x_0) + 25 \sin^2(x_1) + x_0^2 + x_1^2 \tag{100}$$

## 4.101 Egg Holder [46]

The Eggholder function is recognized for its highly non-linear and oscillatory nature, which stems from its formulation that creates an intricate terrain. It is particularly challenging due to the vast number of local minima that can trap optimization algorithms, complicating the search for the global minimum. The function operates within a wide range of $[-512, 512]$ for both $x$ and $y$ dimensions. The significant global minimum is located at $x = 512$ and $y = 404.2319$, where the function value reaches its lowest point at $-959.6407$. This function is Continuous, Differentiable, Non-separable, Scalable, and Multimodal.

$$f(\mathbf{x}) = \sum_{i=1}^{m-1} \left[ -(x_{i+1} + 47) \sin\left(\sqrt{|x_{i+1} + x_i/2 + 47|}\right) \right.$$
$$\left. - x_i \sin\left(\sqrt{|x_i - (x_{i+1} + 47)|}\right) \right] \quad (101)$$

## 4.102 Eggholder with Noise

A variant of the Eggholder function with noise that contains noise and has a range of $[-512, 512]$.

$$f(\mathbf{x}) = \sum_{i=1}^{m-1} \left[ -(x_{i+1} + 47) \sin\left(\sqrt{x_{i+1} + x_i/2 + 47}\right) - \right.$$
$$\left. x_i \sin\left(\sqrt{x_i - (x_{i+1} + 47)}\right) + 0.5 \right] \quad (102)$$

## 4.103 El Attar Vidyasagar Dutta [4,8]

This function is composed of several polynomial terms, including cubic, quadratic, and linear elements, creating a complex landscape with the potential for many local minima. Each of these elements adds its own influence, with higher-order terms introducing more pronounced variations and potential valleys and peaks within the search space of $[-500, 500]$. The notable global minimum for this function is located at $(3.409, -2.171)$ with a function value of $f(x^*) = 1.7127$. This function is Continuous, Differentiable, Non-separable, Non-scalable, and Unimodal.

$$f(\mathbf{x}) = \left(x_0^2 + x_1 - 10\right)^2 + \left(x_0 + x_1^2 - 7\right)^2 + \left(x_0^2 + x_1^3 - 1\right)^2 \quad (103)$$

## 4.104 Elliptic

This function incorporates two quadratic terms, with the $x^2$ term scaled up significantly by $10^6$, making the function disproportionately sensitive to changes in $x$ compared to $y$. This large scaling factor causes the function to increase very rapidly as $x$ moves away from zero, indicating a steep gradient in the $x$-direction and a relatively flatter gradient along the $y$-axis. This function is symmetric and mirrors itself across both axes. A typical domain of the function spans $[-5, 5]$ with a minimum value of $f(x^*) = 0$ at the origin, where the influence of the highly scaled $x^2$ term is neutralized, and $y$ contributes minimally to the function. This point represents the global minimum within a landscape characterized by rapid increases and notable asymmetry between its two principal axes.

$$f(\mathbf{x}) = 1000000.0 x_0^2 + x_1^2 \quad (104)$$

### 4.105 Elliptic N.2

Unlike the previous function, this variant is not scaled in one direction nor symmetric around the origin due to constant shifts in both the $x$ and $y$ components. This function has a range of $[-5, 5]$ and minimizes at $(0, 1)$ and $f(x^*) = 5$.

$$f(\mathbf{x}) = \left(x_1^2 + 1\right)^2 + \left(x_0^2 - 2x_0 + 3\right)^2 \tag{105}$$

### 4.106 Exp 2

This function is a sum of squares of a combination of exponential decay with different rates. As $x$ increases, all exponential terms decay towards zero, but the decay rates differ, which affects their contribution to the sum differently over $x$. This function has a range of $[0, 20]$ and minimizes at 0 and $f(x^*) = 0$.

$$f(\mathbf{x}) = \sum \left(e^{-\frac{x_i}{10}} - 5e^{-\frac{x_i}{2}} + 5e^{-\frac{x_i}{10}} + 5e^{-x_i}\right)^2 \tag{106}$$

### 4.107 Exponential [8]

This function represents a multidimensional Gaussian function, which decays exponentially from the origin. The function peaks at the origin and decays as the Euclidean distance from the origin increases in any direction. This function has a range of $[-1, 1]$ and minimizes at 0 and $f(x^*) = -1$. This function is Continuous, Differentiable, Non-separable, Scalable, and Multimodal.

$$f(\mathbf{x}) = -\exp\left(-0.5\sum_{i=1}^{n} x_i^2\right) \tag{107}$$

### 4.108 Exponential Noise

A variant of the Exponential function with noise that contains noise and has a range of $[-1, 1]$.

$$f(\mathbf{x}) = x_0^2 + x_1^2 + \frac{1}{2} \tag{108}$$

### 4.109 Flexus

This function can be described in polar coordinates, with $r = x^2 + y^2$, and simplifies to $r^2\cos(r) - 1$. Here, $r^2$ modulates the amplitude of the cosine function, resulting in oscillations whose amplitude increases with the radial distance from the origin. Although the cosine function is naturally bounded, the multiplication by $r^2$ causes the term to oscillate increasingly wildly. This scaling effect ensures that as $r$ increases, the landscape of the function becomes more complex. This function operates within a range of $[-5, 5]$ for both $x$ and $y$. The function exhibits multiple local minima that depend on the values of $r$.

$$f(\mathbf{x}) = \left(x_0^2 + x_1^2\right) \cos\left(\sqrt{x_0^2 + x_1^2}\right) - 1 \tag{109}$$

## 4.110 Flux

This function incorporates a sinusoidal component alongside a quadratic element and linear modifications. The sinusoidal function introduces periodic variations, which result in complex interactions when combined with the stabilizing influence of the quadratic term and the directional bias of the linear modifications. More specifically, the presence of the sinusoidal term can lead to the formation of valleys and peaks that are periodically spaced, while the quadratic and linear terms can either deepen these valleys or flatten the peaks depending on their coefficients and signs. The function's range spans $[-100, 100]$. A notable point of interest is at $(1, 1, 0)$, where $f(x^*) = 0$.

$$f(\mathbf{x}) = (x_0 - x_1)^2 + \sin(x_0 + x_1) - 1.5x_0 + 2.5x_1 + 1 \tag{110}$$

## 4.111 Forrester [52]

This multimodal function is characterized by sinusoidal terms with the amplitude modulated by a quadratic element. This slightly revised configuration leads to a landscape with periodic peaks and valleys escalating quadratically from specific points. These critical points are highly dependent on the specific values of $x$ and $y$, while the quadratic component amplifies as one moves away from the origin. This function operates within $[0, 1]$ for each variable.

$$f(\mathbf{x}) = (6x - 2)^2 \sin(12x - 4) \tag{111}$$

$$f(\mathbf{x})_{revised} = (6x_0 - 2)^2 \sin(12x_0 - 4) + (6x_1 - 2)^2 \sin(12x_1 - 4)$$

## 4.112 Freudenstein Roth [20,53]

This function combines complex polynomial expressions in $y$, nested within broader quadratic terms of $x$. The interaction between the polynomial terms and linear shifts determines the function's value and corresponding landscape. The function operates within the range of $[-10, 10]$ for both $x$ and $y$. A notable minimum is located at $(5, 4)$ where $f(x^*) = 0$ and another at $(11.41, -0.8968)$ where $f(x^*) = 48.9842$. This resides at a point where these polynomial dynamics align to minimize the function's value effectively. This function is Continuous, Differentiable, Non-separable, Non-scalable, and Multimodal.

$$f(\mathbf{x}) = \left(\left(\left(5 - x_1\right) x_1 - 2\right) x_1 + x_0 - 13\right)^2 + \left(\left(\left(x_1 + 1\right) x_1 - 14\right) x_1 + x_0 - 29\right)^2 \tag{112}$$

## 4.113 Gaussian [54]

This function integrates a simple quadratic term in $x$ with an exponential term influenced by both $x$ and $y$. The exponential component introduces large modulation and creates a landscape with steep gradients. Operating within the range of $[-5, 5]$ for both $x$ and $y$, the function typically exhibits multiple minima with one reported at (0.4, 1, 0) that equal to $1.12793 \times 10^{-8}$. These minima are contingent on where the quadratic term is minimized while simultaneously balancing the influence of the exponential term. This function is Continuous, Differentiable, Non-separable, Non-scalable, and Multimodal.

$$f(\mathbf{x}) = x_1 \exp\left(-\frac{x_2(t_i - x_3)^2}{2}\right) - y_t \tag{113}$$

Where: $t_i = \frac{8-i}{2}$

| $i$ | $y_t$ |
|---|---|
| 1 | 0.0009 |
| 2 | 0.0044 |
| 3 | 0.0175 |
| 4 | 0.0540 |
| 5 | 0.1295 |
| 6 | 0.2420 |
| 7 | 0.3521 |
| 8 | 0.3989 |

## 4.114 Gaussian Perturbation

A variant of the Gaussian function with a range of $[-5, 5]$.

$$f(x, for2D) = e^{-\left(6x^2+0.8y^2\right)} + e^{-\left((3-x)^2+(2-y)^2\right)} \tag{114}$$

## 4.115 Gear [8]

This function is characterized by the reciprocal of the product of the rounded values of $x$ and $y$, which is then squared to penalize deviations from a constant value. The use of rounding introduces sharp discontinuities at integer boundaries, where slight changes in $x$ or $y$ can lead to significant alterations in the function's value. The range of the function is set between $[12, 60]$ and $(16, 19, 43, 49)$ with $f(x^*) = 2.7$.

$$f(\mathbf{x}) = \left(1.0/6.931 - (x_1 x_2/(x_3 x_4))\right)^2 (115)$$

$$f(x, for2D) = \left(1.0/6.931 - \left(x_1 x_2\right)^2\right)$$

## 4.116 Gear N.2

A variant of the Gear function with multiple minima and a range of $[12, 60]$.

$$f(x, for2D) = \left(\frac{1}{6.931} - \left(\frac{x}{12} \cdot \frac{y}{16}\right)^{-1}\right)^2 \tag{116}$$

### 4.117 Giunta [8,46]

This function involves a series of sinusoidal transformations, each adjusted with varying scales and modified by squaring. These transformations produce a wave-like pattern characterized by periodic peaks and troughs. Minima occur at points where the collective contributions of the sinusoidal terms reach their lowest values. This function has a domain of $[-1, 1]$, and a minimum is observed at $(0.458342, 0.458342)$ with $f(x^*) = 0.06447$. This function is Continuous, Differentiable, Separable, Scalable, and Multimodal.

$$f(\mathbf{x}) = 0.6 + \sum_{i=1}^{n} \left( \sin\left(\frac{16}{15}x_i - 1\right) + \sin^2\left(\frac{16}{15}x_i - 1\right) + \frac{1}{50}\sin\left(4\left(\frac{16}{15}x_i - 1\right)\right)\right) \tag{117}$$

### 4.118 Goldstein Price [6,38]

This function features two polynomial expressions, each containing quadratic and higher-order terms, which are multiplied together. The complexity introduced by the multiplication of these polynomial terms results in a highly varied landscape that is shaped by the synergy of the individual polynomials. This function operates within the domain of $[-2, 2]$, with a minimum observed at $(0, -1)$ with $f(x^*) = 3$. This function is Continuous, Differentiable, Non-separable, Non-Scalable, and Multimodal.

$$f(\mathbf{x}) = \left((2x_0 - 3x_1)^2 \cdot \left(12x_0^2 - 36x_0x_1 - 32x_0 + 27x_1^2 + 48x_1 + 18\right) + 30\right) \left((x_0 + x_1 + 1)^2 \cdot \left(3x_0^2 + 6x_0x_1 - 14x_0 + 3x_1^2 - 14x_1 + 19\right) + 1\right) \tag{118}$$

### 4.119 Goldstein Price with Noise

A variant of the Goldstein Price function with multiple minima and a range of $[-2, 2]$ and $(0, -1)$ and $f(x^*) = 3$.

$$f(\mathbf{x}) = \left((2x_0 - 3x_1)^2 \cdot \left(12x_0^2 - 36x_0x_1 - 32x_0 + 27x_1^2 + 48x_1 + 18\right) + 30\right) \left((x_0 + x_1 + 1)^2 \cdot \left(3x_0^2 + 6x_0x_1 - 14x_0 + 3x_1^2 - 14x_1 + 19\right) + 1\right) + \frac{1}{2} \tag{119}$$

### 4.120 Gramacy Lee

This function integrates oscillatory sinusoidal functions with a severe quartic penalty for deviations from 1. The sinusoidal components contribute to periodic behavior and produce a landscape with regular peaks and valleys. This oscillatory nature is altered near $x = 1$ and $y = 1$, where the quartic penalty terms take effect with sharp increases. This function operates within $[-0.5, 2.5]$ and $(0, -1)$, where $f(x^*) = x^2$.

$$f(\mathbf{x}) = (x-1)^4 + \frac{\sin(10\pi x)}{2x} \tag{120}$$

$$f(x, for2D) = (x_0 - 1)^4 + (x_1 - 1)^4 + \frac{\sin(10\pi x_0)}{2x_0} + \frac{\sin(10\pi x_1)}{2x_1}$$

### 4.121 Griewank [55]

This is a multimodal function with many regularly distributed local minima. The suggested search area is the hypercube $[-600, 600]$ with 0 and $f(x^*) = 0$.

$$f(\mathbf{x}) = \sum_{i=1}^{n} \frac{x_i^2}{4000} - \prod_{i=1}^{n} \cos\left(\frac{x_i}{\sqrt{i}}\right) + 1 \tag{121}$$

$$f(x, for2D) = -\cos\left(\frac{\sqrt{2}x_1}{2}\right)\cos(x_0) + \frac{x_0^2}{4000} + \frac{x_1^2}{4000} + 1$$

### 4.122 Griewank with Noise

A variant of the Griewank function with multiple minima and a range of $[-600, 600]$.

$$f(x, for2D) = \frac{x^2 + y^2}{4000} - \cos(x) \cdot \cos\left(\frac{y}{\sqrt{2}}\right) + 1 + \epsilon \tag{122}$$

### 4.123 Gulf Research [4,54]

This function involves an array of values $u$ derived from a logarithmic transformation that is modeled using an exponential decay function. The core of the function's behavior is the fitting process, which aims to minimize the squared residuals between the calculated model and the actual data points. This function operates within a broad range, with domains of $[0, 5]$ and $[0, 60]$, and identifies a global minimizer at $(50, 25, 1.5)$ where $f(x^*) = 0$. This function is Continuous, Differentiable, Non-separable, Non-scalable, and Multimodal.

$$f(\mathbf{x}) = \sum_{i=1}^{99} \left[\exp\left(-\frac{(u_i - x_2)^3}{x_1}\right) - 0.01i\right]^2 \tag{123}$$

where $\mathrm{u}_i = 25 + [-50\ln(0.01i)]^{1/1.5}$,
subject to $0.1 \le x_1 \le 100, \quad 0 \le x_2 \le 25.6$

### 4.124 Gulf Research and Development [56]

This function is composed of several Gaussian functions, each centered at distinct points on the plane, with one of the Gaussians being scaled differently. The overlapping nature of these Gaussian functions results in a complex surface characterized by multiple interaction points (i.e., peaks due to the multiplicative interactions among the Gaussians). The function operates within $[-5, 5]$ where $f(x^*) = 0$. This function is also known as the Weibull function.

$$f(\mathbf{x}) = 2\left(e^{-(x_0-1)^2-(x_1+1)^2} + 0.004\right)\left(e^{-(x_0+1)^2-(x_1-1)^2} + 0.004\right)e^{-(x_0-2)^2-x_1^2} \tag{124}$$

### 4.125 Hansen [8]

This function integrates cosine terms and features increasing frequencies and phases, all of which are combined to form a ratio. The behavior of the function is defined by periodic fluctuations from the cosine terms, and the ratio format amplifies these effects, particularly near points where the denominator approaches zero. This function is prone to dramatic changes near zeros of the denominator. The function operates within a range of $[-10, 10]$ with significant local minima. A notable minimizer is located at $(-7.589, -7.708)$ with a function value of $f(x^*) = -2.345$. This function is Continuous, Differentiable, Separable, Non-scalable, and Multimodal.

$$f(\mathbf{x}) = \sum_{i=1}^{4}(i+1)\cos(ix_0 + i + 1)\sum_{j=1}^{4}(j+1)\cos(j + (j+2)x_1 + 1) \tag{125}$$

### 4.126 Happy Cat

This function integrates absolute value fractional powers with quadratic components. In the numerator, the magnitude of $(x, y)$ is adjusted using a fractional power that interacts with the quadratic terms. The denominator scales the square of the radius and impacts the overall amplitude of the function. This configuration leads to a multimodal landscape with the global minimum located within a narrow, curved valley. This function operates within the domain $[-2, 2]$.

$$f(\mathbf{x}) = \left[\left(||\mathbf{x}||^2 - d\right)^2\right]^{\alpha} + \frac{1}{d}\left(\frac{1}{2}||\mathbf{x}||^2 + \sum_{i=1}^{d} x_i\right) + \frac{1}{2} \tag{126}$$

$$f(x, for2D) = \frac{\left(|x^2+y^2|^{\alpha} - \left(0.5x^2 + 0.5y^2\right) + 0.5\right)}{1 + 0.001(x^2+y^2)^2}$$

### 4.127 Hartman 3 [9,37,57]

This function constructs a complex landscape using arrays of parameters to form a weighted sum of exponentials that are influenced by quadratic differences from specific reference points. Each term within the sum responds to deviations from predefined points listed in $P$, with the magnitude and direction of these responses governed by parameters in $A$ and weighted by $\alpha$. Because of the negative sign in the function's formulation, it effectively seeks the minima of this negative exponential sum. This function operates within $[0, 1]$ and at $(0.1146, 0.5556, 0.8525)$ and $f(x^*) = -3.8627$. This function is Continuous, Differentiable, Non-separable, Non-scalable, and Multimodal.

$$f(\mathbf{x}) = -\sum_{i=1}^{4} \alpha_i \exp\left(-\sum_{j=1}^{3} A_{ij}(x_j - P_{ij})^2\right)$$

$$\text{where } \alpha = \begin{pmatrix} 1.0 \\ 1.2 \\ 3.0 \\ 3.2 \end{pmatrix}$$

$$A = \begin{pmatrix} 3.0 & 10 & 30 \\ 0.1 & 10 & 35 \\ 3.0 & 10 & 30 \\ 0.1 & 10 & 35 \end{pmatrix}$$

$$P = 10^{-4} \times \begin{pmatrix} 3689 & 1170 & 2673 \\ 4699 & 4387 & 7470 \\ 1091 & 8732 & 5547 \\ 381 & 5743 & 8828 \end{pmatrix} \tag{127}$$

### 4.128 Hartmann 6 [6,58]

This function operates within the range of $[0, 1]$ and has multiple minima $(0.2016, 0.1500, 0.4768, 0.275332, 0.3116$ $f(x^*) = -3.3223$. This function is also known as Hartman's Function N.2 or Hartman N.6 Problem. This function is Continuous, Differentiable, Non-separable, Non-scalable, and Multimodal.

$$f(\mathbf{x}) = -\sum_{i=1}^{4} c_i \exp\left(-\sum_{j=1}^{6} a_{ij}(x_j - p_{ij})^2\right)$$
$$\text{subject to } 0 \le x_j \le 1, \quad j \in \{1, \ldots, 6\}, \tag{128}$$

where the constants $a_{ij}, p_{ij}$ and $c_i$ are given as:

$$A = \begin{bmatrix} 10 & 3 & 17 & 3.5 & 1.7 & 8 \\ 0.05 & 10 & 17 & 0.1 & 8 & 14 \\ 3 & 3.5 & 1.7 & 10 & 17 & 8 \\ 17 & 8 & 0.05 & 10 & 0.1 & 14 \end{bmatrix},$$

$$c = \begin{bmatrix} 1 \\ 1.2 \\ 3 \\ 3.2 \end{bmatrix},$$

$$P = \begin{bmatrix} 0.1312 & 0.1696 & 0.5569 & 0.0124 & 0.8283 & 0.5586 \\ 0.2329 & 0.4135 & 0.8307 & 0.3736 & 0.1004 & 0.9991 \\ 0.2348 & 0.1451 & 0.3522 & 0.2883 & 0.3047 & 0.6650 \\ 0.4047 & 0.8828 & 0.8732 & 0.5743 & 0.1091 & 0.0381 \end{bmatrix}.$$

## 4.129 Helical Valley [59]

This function incorporates an angular component $\theta$ that reflects a transformation to polar coordinates and combines it with a radial term. The relation between $\theta$ and the radial distance forms the core of the function's dynamics (with $z$ introducing its own distinct terms into the function). The function operates within the domain $[-10, 10]$, with $(1, 0, 0)$ where $f(x^*) = 0$. This function is Continuous, Differentiable, Non-separable, Scalable, and Multimodal.

$$f(\mathbf{x}) = 100\left((x_3 - 10\theta)^2 + \left(\sqrt{x_1^2 + x_2^2} - 1\right)^2\right) + x_3^2 \tag{129}$$

where $\theta = \begin{cases} \frac{1}{2\pi}\tan^{-1}\left(\frac{x_1}{x_2}\right) & \text{if } x_1 \geq 0 \\ \frac{1}{2\pi}\tan^{-1}\left(\frac{x_1}{x_2} + 0.5\right) & \text{if } x_1 < 0 \end{cases}$

## 4.130 Himmelblau [8,9]

This function integrates squared and linear terms of two variables. The function's dynamic is characterized by the interplay between the squared terms, which create a parabolic response, and the linear terms, which shift the landscape. This function operates in the range of $[-6, 6]$ for one variable and $[-2, 2]$. The function's known solutions at points $(0, 0)$ and $(-1, 1)$ both yield $f(x^*) = 0$. This function is Continuous, Differentiable, Non-separable, Non-scalable, and Multimodal.

$$f(\mathbf{x}) = \left(x_0 + x_1^2 - 7\right)^2 + \left(x_0^2 + x_1 - 11\right)^2 \tag{130}$$

## 4.131 Holder Table 1 [8,46]

This function merges cosine elements with an exponential element influenced by the normalized radial distance from the origin. The behavior of the function is shaped by an exponential decay dependent on the distance from a circle of radius $\pi$. This setup results in a landscape featuring a series of ridges and valleys, where the function values are modulated by the product of cosine functions, creating a complex topographical interplay. These minima occur at points where the cosine values are maximized (i.e., less negative) with the exponential decay minimized. This function has a search space of $[-5, 5]$, with minima at $(\pm 8.055, \pm 9.664)$ and $f(x^*) = -19.2085$ and $(\pm 9.6645, \pm 9.6645)$ and $f(x^*) = -26.9203$. This function is Continuous, Differentiable, Separable, Non-scalable, and Multimodal.

$$f(\mathbf{x}) = -\left|\cos(x)\cos(y)\exp\left(\left|1 - \frac{\sqrt{x^2+y^2}}{\pi}\right|\right)\right| \tag{131}$$

### 4.132 Holder Table 2

A variant of the Holder Table function with multiple minima and a range of $[-5, 5]$. This function is Continuous, Differentiable, Separable, Non-scalable, and Multimodal.

$$f(\mathbf{x}) = -\left|\sin(x)\cos(y)\exp\left(\left|1 - \frac{\sqrt{x^2+y^2}}{\pi}\right|\right)\right| \tag{132}$$

### 4.133 Holder Table N.2

A variant of the Holder Table function with multiple minima and a range of $[-5, 5]$.

$$f(\mathbf{x}) = -\left|\sin(x)\cos(y)\cos(xy)\exp\left(\left|1 - \frac{\sqrt{x^2+y^2}}{\pi}\right|\right)\right| \tag{133}$$

### 4.134 Holzman [8,33]

This function is structured around reciprocal influences of $x$ and $y$, where the impact of each variable on the function value is modulated by the expression of another. This function generally exhibits lower values near the axes where either $x$ or $y$ approaches zero. The function is defined over a range of $[-10, 10]$, with additional specified ranges for other variables in more complex scenarios (e.g., $0.1 \leq x_1 \leq 100$, $0 \leq x_2 \leq 25.6$, $0 \leq x_3 \leq 5$). The minimum of the function occurs at $x = y = 0$, where the function value reaches 0.

$$f(\mathbf{x}) = \sum_{i=0}^{98}\left(-0.1(i+1) + \exp\left(\frac{1}{x_1}(u_i - x_2)^{x_3}\right)\right) \tag{134}$$

where $\mathrm{u}_i = 25 + (-50\log(0.01(i+1)))^{\frac{2}{3}}$

### 4.135 Holzman N.2 [8,33]

A variant of the Holzman function.

$$f(\mathbf{x}) = \sum_{i=0}^{n} \left(i{x_i}^4\right) \tag{135}$$

### 4.136 Hosaki [60]

This function integrates a quartic polynomial in $x$ with an exponentially modulated quadratic term in $y$. The quartic term governs the overall shape and complexity of the function, as $x$ dictates the polynomial's sign and magnitude, creating a varied landscape based on the quartic expression's coefficients. Meanwhile, $y$ impacts the function through an exponential decay applied to the $y^2$ term. Minima are likely to be found near the roots of the polynomial, where its influence is minimized, or in regions where the exponential decay of $y^2$ dominates. This function has a range for $x_1$ in $[0, 5]$ and $x_2$ in $[0, 6]$, and at $(4, 2)$ with $f(x^*) = -2.3458$. This function is Continuous, Differentiable, Non-separable, Non-scalable, and Multimodal.

$$f(\mathbf{x}) = \left(\frac{x_0^4}{4} - \frac{7x_0^3}{3} + 7x_0^2 - 8x_0 + 1\right) e^{-x_1} x_1^2 \tag{136}$$

### 4.137 Hosaki Exponential

This function combines additional complexity added by the modulated polynomial. This function has a range of $[-5, 5]$.

$$f(\mathbf{x}) = 10\left(-x_0^3 + x_1\right)^2 + \left(\frac{x_0^4}{4} - \frac{7x_0^3}{3} + 7x_0^2 - 8x_0 + 1\right) e^{-x_0^2} \tag{137}$$

### 4.138 Hosaki Exponential with Noise

A variant of the Hosaki Exponential function with noise and a range of $[-5, 5]$.

$$f(\mathbf{x}) = \left(1 - 8x + 7x^2 - \frac{7}{3}x^3 + \frac{1}{4}x^4\right) e^{-x^2} + 10(y - x^3)^2 \tag{138}$$

### 4.139 Hqing

This function is composed of quartic terms with opposing signs, potentially leading to a double-well behavior. The interaction between positive and negative powers generates multiple local minima, particularly pronounced where the quartic and quadratic terms are in balance. The function operates within a range of $[-5, 5]$ and at 0 where $f(x^*) = 0$.

$$f(\mathbf{x}) = -x_0^4 + x_0^2 + 16x_1^4 - 16x_1^2 \tag{139}$$

### 4.140 Hump

This function incorporates rich polynomial interactions comprising a sixth-degree polynomial in $x$, a mixed product term, and a fourth-degree polynomial in $y$. This function operates within the domain $[-5, 5]$ with minimizers at $(0.0898, -0.7126)$ and $(-0.0898, 0.7126)$, where $f(x^*) = 0$.

$$f(\mathbf{x}) = \frac{x_0^6}{3} - 2.1x_0^4 + 4x_0^2 + x_0x_1 + 4x_1^4 - 4x_1^2 \tag{140}$$

### 4.141 Infinity

The function is smooth and oscillatory due to the presence of the sin terms. The function can have a range between $[-0.2172, 1]$ with multiple minima.

$$f(\mathbf{x}) = \frac{\sin(x)}{x} \cdot \frac{\sin(y)}{y} \tag{141}$$

### 4.142 Inverted Cosine Wave

The function has a range between $[-5, 5]$.

$$f(\mathbf{x}) = -\sum_{i=1}^{n-1} \left( \exp\left( -\frac{x_i^2 + x_{i+1}^2 + 0.5x_ix_{i+1}}{8} \right) \cos\left( 4\left( x_i^2 + x_{i+1}^2 + 0.5x_ix_{i+1} \right)^{0.5} \right) \right) \tag{142}$$

### 4.143 Jennrich Sampson [4,8]

This function uses exponential terms that are scaled by an index $i$, squared, and further modified by a linear scale of $i$. Due to these terms' exponential nature and squaring, the function can exhibit rapid growth and is designed to penalize deviations from specific exponential behaviors. This function operates within the range of $[-1, 1]$, and the scaling and squaring of the exponential terms greatly influence its behavior. A notable minimum is located at $(0.2578, 0.2578)$ with a function value of $f(x^*) = 124.362$. This function is Continuous, Differentiable, Non-separable, Non-scalable, and Multimodal.

$$f(\mathbf{x}) = \sum_{i=1}^{10} \left( 2i - \left( e^{ix_0} + e^{ix_1} \right)^2 + 2 \right)^2 \tag{143}$$

### 4.144 Judge [8]

This function combines squared differences of sinusoidal terms with a squared distance term between $x$ and $y$. The function's value increases with the square of the linear distance and the square of the trigonometric differences. This design creates potential wells in the function's landscape at points where the sinusoidal terms align or effectively cancel each other out. The function's domain spans $[-10, 10]$ and has a specific minimum at $(0.864, 1.235)$ with a function value of $f(x^*) = 16.0817$.

$$f(\mathbf{x}) = \sum_{i=1}^{20} \left( [x_1 + B_{i2}x_2 + C_{i3}x_3] - A_{i1} \right)^2 \tag{144}$$

where A, B, and C are given as follows:

$$\mathrm{A} = \begin{bmatrix} 4.284 \\ 4.149 \\ 3.877 \\ 0.533 \\ 2.211 \\ 2.389 \\ 2.145 \\ 3.231 \\ 1.998 \\ 1.379 \\ 2.106 \\ 1.428 \\ 1.011 \\ 2.179 \\ 2.858 \\ 1.388 \\ 1.651 \\ 1.593 \\ 1.046 \\ 2.152 \end{bmatrix}, \quad \mathrm{B} = \begin{bmatrix} 0.286 \\ 0.973 \\ 0.384 \\ 0.276 \\ 0.973 \\ 0.543 \\ 0.957 \\ 0.948 \\ 0.543 \\ 0.797 \\ 0.936 \\ 0.889 \\ 0.006 \\ 0.828 \\ 0.399 \\ 0.617 \\ 0.939 \\ 0.784 \\ 0.072 \\ 0.889 \end{bmatrix}, \quad \mathrm{C} = \begin{bmatrix} 0.645 \\ 0.585 \\ 0.310 \\ 0.058 \\ 0.455 \\ 0.779 \\ 0.259 \\ 0.202 \\ 0.028 \\ 0.099 \\ 0.296 \\ 0.296 \\ 0.175 \\ 0.180 \\ 0.842 \\ 0.039 \\ 0.103 \\ 0.620 \\ 0.158 \\ 0.704 \end{bmatrix}.$$

## 4.145 Katsuura [8,22]

This function involves terms that are computed as fractions of $x$ relative to powers of two, with these terms being adjusted by rounding, introducing a periodic component. This setup allows for sophisticated patterns of behavior due to the rounding's impact and the inverse scaling by $x^2$. The function's range spans two domains, $[-1000, 1000]$ and $[0, 100]$, and achieves a minimum at $f(x^*) = 1$.

$$f(\mathbf{x}) = \prod_{i=0}^{n-1} \left( 1 + (i+1) \sum_{k=1}^{d} \mathrm{round}(2^k x_i) 2^{-k} \right) \tag{145}$$

where $\mathrm{n} = 10$ and $\mathrm{d} = 32$ for testing.

## 4.146 Keane [4,8]

This function combines the squares of sinusoidal functions whose arguments depend on both the difference and sum of $x$ and $y$, with an additional complexity added by dividing the product by the radial distance from the origin. This division modifies the amplitude of the sinusoidal output and makes the oscillations more pronounced closer to the origin. This function operates within a domain of $[-10, 10]$ with $(1.3932, 0)$ and $f(x^*) = 0.6736$. This function is Continuous, Differentiable, Non-separable, Non-scalable, and Multimodal.

$$f(\mathbf{x}) = \frac{\sin^2(x_0 - x_1)\sin^2(x_0 + x_1)}{\sqrt{x_0^2 + x_1^2}} \tag{146}$$

## 4.147 Keane N.2

A variant of the Keane function with multiple minima and a range of $[-6.97, 6.97]$.

$$f(\mathbf{x}) = \frac{\sin^2(x_0 - x_1)\sin^2(x_0 + x_1)}{\sqrt{x_0^2 + x_1^2} + 1.0 \cdot 10^{-8}} \tag{147}$$

## 4.148 Kearfott[25]

This function has four global minima at $f(x^*) \approx 0$ with $[\pm\sqrt{1}.5, \pm\sqrt{0}.5]$.

$$f(\mathbf{x}) = \frac{\sin^2(x_0 - x_1)\sin^2(x_0 + x_1)}{\sqrt{x_0^2 + x_1^2} + 1.0 \cdot 10^{-8}} \tag{148}$$

## 4.149 Kowalik [8]

This function is designed as a least squares problem, intended to find the parameter values that minimize its discrepancy. This function has a range within $[0, 42]$ for one set of parameters and $[-5, 5]$ for another; the function identifies a global optimal value with $f(x^*) \approx 3.0748 \times 10^{-4}$. The optimal parameters are approximately at $x \approx (0.192, 0.190, 0.123, 0.135)$.

$$f(\mathbf{x}) = \sum_{i=0}^{10} \left\{a_i - \left(x_0(b_i^2 + b_{i,1}x_1)/(b_i^2 + b_{i,2}x_2 + x_3)\right)\right\}^2 \tag{149}$$

where a = (0.1957, 0.1947, 0.1735, 0.1600, 0.0844, 0.0627, 0.0456, 0.0342, 0.0323, 0.0235, 0.0246),
b = $\left(4, 2, 1, \frac{1}{2}, \frac{1}{4}, \frac{1}{6}, \frac{1}{8}, \frac{1}{10}, \frac{1}{12}, \frac{1}{14}, \frac{1}{16}\right)$.

## 4.150 Langermann [3],[8]

The function is complex and oscillatory, with behavior heavily dependent on the matrix $A$ and vector $x$. This function has a range between $[-10, 10]$, with $(2.0029, 1.006)$ and $f(x^*) = -5.162$. This function is Continuous, Differentiable, Non-separable, Non-scalable, and Multimodal.

$$f(\mathbf{x}) = \sum_{i=1}^{m} c_i exp\left(-\frac{1}{\pi}\sum_{j=1}^{d}(x_j - A_{ij})^2\right) cos\left(\pi\sum_{j=1}^{d}(x_j - A_{ij})^2\right) \tag{150}$$

where $\mathrm{A} = \begin{bmatrix} 3 & 5 \\ 5 & 2 \\ 2 & 1 \\ 1 & 4 \\ 7 & 9 \end{bmatrix}$

## 4.151 Langermann N.2

A variant of the Langermann function with multiple minima and a range of $[-1.52, 10]$.

$$f(\mathbf{x}) = f(x, y) = -\sum_{i=1}^{m} c_i \exp\left(-\frac{1}{\pi}d_i\right)\cos(\pi d_i) \tag{151}$$

where $\mathrm{d}_i = (x - x_i)^2 + (y - y_i)^2$, and

$$m = 5, \quad c = [1, 2, 5, 2, 3], \quad A = \begin{bmatrix} 3 & 5 \\ 5 & 2 \\ 2 & 1 \\ 1 & 4 \\ 7 & 9 \end{bmatrix}.$$

## 4.152 Lennard Jones [8, 62]

The function is modeled after the Lennard-Jones potential, often used in simulations of particle interactions. It features two distinct components that govern its behavior at various distances: the $1/r^{12}$ term and the $1/r^6$ term. The first term is predominant at very close distances, imparting a strong repulsive force to prevent particles from overlapping, simulating the physical impossibility of particles occupying the same space. On the other hand, the second term facilitates attraction at intermediate distances and models the attractive forces that operate at a larger range than the repulsive force. This function is singular at the origin, where it tends to infinity as $r \to 0$. The well in the function's landscape indicates the optimal balance point or the minimum energy state. This characteristic makes the function relevant in fields like molecular dynamics and materials science. The function operates within the domain of $[-5, 5]$ for $r$.

$$f(\mathbf{x}) = \sum_{i=0}^{N-2}\sum_{j=i+1}^{N-1}\left(\frac{1}{r_{ij}^{12}} - \frac{1}{r_{ij}^{6}}\right) \tag{152}$$

where $\mathrm{r}_{ij}$ is the distance between atoms labeled $i$ and $j$ given by:
$r_{ij} = \sqrt{(x_{3i} - x_{3j})^2 + (x_{3i+1} - x_{3j+1})^2 + (x_{3i+2} - x_{3j+2})^2}$

## 4.153 Leon [8, 61]

This function is also known as the Rosenbrock Function (to be fully presented in a subsequent section).

$$(1-x_0)^2 + 100\left(-x_0^2 + x_1\right)^2 \tag{153}$$

## 4.154 Levy Function [7]

This function modulates $x$ and $y$ and applies both sinusoidal and quadratic transformations to create a complex interaction between these elements. Periodic oscillations from the sinusoidal terms characterize the behavior of the function, and an overarching quadratic growth steers the global behavior upward as $x$ and $y$ deviate from $w_1 = 1$ and $w_2 = 1$. This function operates within a domain of $[-10, 10]$ and $f(x^*) = 0$.

$$\begin{aligned} f(\mathbf{x}) = \sin^2(\pi y_1) + \sum_{i=1}^{d-1} &\left[(y_i - 1)^2 \left(1 + 10\sin^2(\pi y_i + 1)\right)\right. \\ &+ (y_d - 1)^2 \left(1 + \sin^2(2\pi y_d)\right) \end{aligned} \tag{154}$$

where $\mathrm{y}_i = 1 + \frac{x_i - 1}{4}$, for all $i = 1, \ldots, d$

## 4.155 Levy N.3 [8]

A variant of the Levy function with a range of $[-10, 10]$.

$$\begin{aligned} f(\mathbf{x}) = \sin^2(\pi y_1) + \sum_{i=1}^{d-1} &\left[(y_i - 1)^2 \left(1 + 10\sin^2(\pi y_i + 1)\right)\right. \\ &+ (y_d - 1)^2 \left(1 + \sin^2(2\pi y_d)\right) \end{aligned} \tag{155}$$

where $\mathrm{y}_i = 1 + \frac{x_i - 1}{4}$, for all $i = 1, \ldots, d$

## 4.156 Levy N.5 [8]

A variant of the Levy function with a range of $[-10, 10]$ and at $(-1.3068, -1.4248)$ and $f(x^*) = -176.1375$.

$$\begin{aligned} f(\mathbf{x}) = \sum_{i=1}^{5} i\cos\left((i-1)x_1 + x_1\right) \times \sum_{j=1}^{5} j\cos\left((j+1)x_2 + x_2\right) + (x_1 + 1.42513)^2 \\ + (x_2 + 0.80032)^2 \end{aligned} \tag{156}$$

## 4.157 Levy N.8

A variant of the Levy function with a range of $[-10, 10]$.

$$f(\mathbf{x}) = \sin^2(\pi y_0) + \sum_{i=0}^{n-2}(y_i - 1)^2 \left[1 + 10\sin^2(\pi \times y_{i+1} + 1)\right] + \\ (y_{n-1} - 1)^2 \left[1 + \sin^2(2\pi \times x_{n-1})\right] \quad (157)$$

where $\mathrm{y}_i = 1 + \frac{x_i - 1}{4}$, for all $i = 1, \ldots, d$

## 4.158 Levy N.13 [8]

A variant of the Levy function with a range of $[-10, 10]$ and at 1, $f(x^*) = 0$.

$$f(\mathbf{x}) = \sin^2(3\pi x_1) + (x_1 - 1)^2 \left[1 + \sin^2(3\pi x_2)\right] + (x_2 - 1)^2 \left[1 + \sin^2(2\pi x_2)\right] \quad (158)$$

## 4.159 Levy and Gomez

This function is non-convex and has a global minimum located at (0.08984, –0.71265) = –1.03162 and at, $f(x^*) = -1.03168$ (For a = 1.5).

$$f(\mathbf{x}) = 4x_1^2 - 2.1x_1^4 + \frac{x_1^6}{3} + x_1x_2 - 4x_2^2 + x_2^4 \\ \text{subjected to: } g = -\sin(4\pi x_1) + 2\sin^2(2\pi x_2) \leq a \quad (159)$$

## 4.160 Matyas [8, 63]

This function incorporates squares of $x$ and $y$ along with a cross-product term acting toward shaping an elliptical paraboloid. The inclusion of the negative cross-product modifies the standard parabolic behavior, potentially introducing saddle points into the function's landscape. This function operates within the domain $[-10, 10]$ and achieves a minimum at $(0, 0)$, $f(x^*) = 0$. This function is Continuous, Differentiable, Non-separable, Non-scalable, and Multimodal.

$$f(\mathbf{x}) = 0.26x_0^2 - 0.48x_0x_1 + 0.26x_1^2 \quad (160)$$

## 4.161 McCormick [8, 63]

This function merges a sinusoidal component with a quadratic form and linear terms to create a dynamic landscape. The sinusoidal part introduces periodic fluctuations, while the quadratic and linear elements contribute to the overall shaping of the terrain to form valleys and ridges. This function operates within the domain of $x_1 \in [-1.5, 4]$ and $x_2 \in [-3, 4]$. A minima can be identified at $(-0.54719, -1.54719)$ with $f(x^*) = -1.9132$. This function is Continuous, Differentiable, Non-separable, Non-scalable, and Multimodal.

$$f(\mathbf{x}) = (x_0 - x_1)^2 + \sin(x_0 + x_1) - 1.5x_0 + 2.5x_1 + 1 \quad (161)$$

## 4.162 McCormick with Noise

A variant of the McCormick function with additional noise. This function has a range of $[-1.5, 4]$ and at $(-0.5471, -1.5471)$ and $f(x^*) = -1.9132$.

$$f(\mathbf{x}) = \sin(x + y) + (x - y)^2 - 1.5x + 2.5y + 1 + (\text{noise between [-1,1]}) \quad (162)$$

## 4.163 Meyer [54]

For n $= 2$, $m = 16$, and$t = 45 + 5i$. This function has a minima of 87.9458 at (0.02, 4000, 250).

$$x_1 \exp\left(\frac{x_2}{t + x_3}\right) - y_t \quad (163)$$

Where the values of $y_t$ corresponding to different $t$ are provided in the table below:

| $t$ | $y_t$ |
|---|---|
| 1 | 34780 |
| 2 | 28610 |
| 3 | 23650 |
| 4 | 19630 |
| 5 | 16370 |
| 6 | 13720 |
| 7 | 11540 |
| 8 | 9744 |
| 9 | 8261 |
| 10 | 7030 |
| 11 | 6005 |
| 12 | 5147 |
| 13 | 4427 |
| 14 | 3820 |
| 15 | 3307 |
| 16 | 2872 |

## 4.164 Michalewicz [7, 8]

This is a multimodal function with a landscape characterized by $d!$ local minima, where $d$ is the dimensionality of the function. This function's difficulty is modulated by the parameter $m$, which adjusts the steepness of valleys and ridges. A higher $m$ value, such as the recommended $m = 10$, significantly intensifies the landscape's complexity. This function operates within the range of $[0, \pi]$. For instance, at two dimensions ($d = 2$), the function reaches a minimum of $-1.8013$. As the dimension increases, the minima deepen, with $-4.6876$ at $d = 5$ and $-9.6601$ at $d = 10$.

$$f(\mathbf{x}) = -\sum_{i=1}^{d} \sin(x_i) \sin^{2m}\left(\frac{ix_i^2}{\pi}\right) \tag{164}$$

## 4.165 Michalewicz N.2

A variant of the Michalewicz function with a range of $[0, \pi]$ and multiple minima. The function values cluster close to zero with negative values, showing the influence of the $-0.5$ scaling factor.

$$f(\mathbf{x}) = -0.5 \sin^{20}\left(\frac{x_0^2}{\pi}\right) \sin(x_0) - 0.5 \sin^{20}\left(\frac{2x_1^2}{\pi}\right) \sin(x_1) \tag{165}$$

## 4.166 Michalewicz with Noise

A variant of the Michalewicz function with a range of $[0, \pi]$ and multiple minima.

$$f(\mathbf{x}) = -\sum_{i=1}^{d} \sin(x_i) \sin^{2m}\left(\frac{ix_i^2}{\pi}\right) + (\text{noise between [-1,1]}) \tag{166}$$

## 4.167 Miele Cantrell [6, 8]

This function is characterized by a combination of exponential, polynomial, and trigonometric terms. The presence of exponential and high-power polynomial components causes the function to exhibit steep variations and then becomes particularly sensitive to minor input adjustments. The function operates within a restricted range of $[-1, 1]$. This function reaches a minimum value of $f(x^*) = 0$ at the point $(0, 1, 1, 1)$. This function is Continuous, Differentiable, Non-separable, Non-scalable, and Multimodal.

$$f(\mathbf{x}) = 100\left(x_1 - x_2\right)^6 + \left(-x_2 + e^{-x_0}\right)^4 + \tan^4\left(x_2 - x_3\right) + x_0^8 \tag{167}$$

## 4.168 Mishra [8]

This function consists of sums of elements of $x$ and their weighted sums and measures a squared error from a target value adjusted by the total count of elements. Its quadratic nature means it quantifies the deviation from a specified linear combination of the variables of zero. This function operates within a domain of $[0, 1]$. This function is Continuous, Differentiable, Non-separable, Scalable, and Multimodal.

$$f(\mathbf{x}) = \left(1 + D - \sum_{i=1}^{N-1} x_i\right)^{N - \sum_{i=1}^{N-1}(x_i)} \tag{168}$$

## 4.169 Mishra N.2 [8]]

A variant of the Mishra function with a range of $[0, 1]$ and at $x = 1$, $f(x^*) = 2$. This function is Continuous, Differentiable, Non-separable, Non-scalable, and Multimodal.

$$f(\mathbf{x}) = \left(1 + D - \sum_{i=1}^{N-1} 0.5(x_i + x_{i+1})\right)^{N - \sum_{i=1}^{N-1} 0.5(x_i + x_i + 1)} \tag{169}$$

## 4.170 Mishra N.3 [8, 64]

A variant of the Mishra function with a range of $[-10, 10]$ and at $-10$, $f(x^*) = -0.1846$. This function is Continuous, Differentiable, Non-separable, Non-scalable, and Multimodal.

$$f(\mathbf{x}) = \sqrt{\left|\cos\left(\sqrt{x_0^2 + x_1^2}\right)\right|} + 0.01x_0 + 0.01x_1 \tag{170}$$

## 4.171 Mishra N.4 [8, 64]

A variant of the Mishra function with a range of $[-10, 10]$ and at $-10$, $f(x^*) = -0.1994$. This function is Continuous, Differentiable, Non-separable, Non-scalable, and Multimodal.

$$f(\mathbf{x}) = \sqrt{\left|\sin\left(\sqrt{x_0^2 + x_1^2}\right)\right|} + 0.01x_0 + 0.01x_1 \tag{171}$$

## 4.172 Mishra N.5 [[8, 64]]

A variant of the Mishra function with a range of $[-10, 10]$ and at $(-1.986, -10)$ and $f(x^*) = -1.01982$. This function is Continuous, Differentiable, Non-separable, Non-scalable, and Multimodal.

$$f(\mathbf{x}) = \left[\sin^2\left(\cos(x_1) + \cos(x_2)\right)^2 + \cos^2\left(\sin(x_1) + \sin(x_2)\right) + x_1\right]^2 + 0.01(x_1 + x_2) \tag{172}$$

### 4.173 Mishra N.6 [8]

A variant of the Mishra function with a range of $[-10, 10]$ and at $(2.886, 1.823)$ and $f(x^*) = -2.283$. This function is Continuous, Differentiable, Non-separable, Non-scalable, and Multimodal.

$$f(\mathbf{x}) = -ln\left[\sin^2\left(\cos(x_1)+\cos(x_2)\right)^2 + \cos^2\left(\sin(x_1)+\sin(x_2)\right) + x_1\right]^2 + 0.01(x_1 + x_2) \quad (173)$$

### 4.174 Mishra N.7 [8]

A variant of the Mishra function with a range of $[-10, 10]$ and $f(x^*) = 0$. This function is Continuous, Differentiable, Non-separable, Non-scalable, and Multimodal.

$$f(\mathbf{x}) = \left[\prod_{i=1}^{D}(x_i - N!)\right]^2 \quad (174)$$

### 4.175 Mishra N.8 [8, 64]

A variant of the Mishra function with a range of $[-10, 10]$ and at $(-3, 2)$ and $f(x^*) = 0$. This function is also known as the Decanomial function. This function is Continuous, Differentiable, Non-separable, Non-scalable, and Multimodal.

$$f(\mathbf{x}) = \left[0.001\left(|x_1^{10} - 20x_1^9 + 180x_1^8 - 960x_1^7 + 3360x_1^6 - 8064x_1^5 + 13344x_1^4 - 15360x_1^3 + 11520x_1^2 - 5120x_1 + 2624 + x_2^4 + 12x_2^3 + 54x_2^2 + 108x_2 + 81|\right]^2 \quad (175)$$

### 4.176 Mishra N.9 [8]

A variant of the Mishra function with a range of $[-10, 10]$ and at $(1, 2, 3)$ and $f(x^*) = 0$. This function is Continuous, Differentiable, Non-separable, Non-scalable, and Multimodal.

$$f(\mathbf{x}) = \left[ab^2c + abc^2 + b^2 + (x_1 + x_2 - x_3)^2\right]^2 \quad (176)$$

where $a = 2x_1^3 + 5x_1x_2 + 4x_3 - 2x_1x_3^3 - 18$,
$b = x_1 + x_3^2 + x_1x_3^2 - 22$,
$c = 8x_2 + 2x_2x_3 + 2x_2^2 + 3x_3^2 - 52$.

## 4.177 Mishra N.10 [8, 64]

A variant of the Mishra function with a range of $[-10, 10]$ and $f(x^*) = 0$. This function is Continuous, Differentiable, Non-separable, Scalable, and Multimodal.

$$f(\mathbf{x}) = [|x_1 + x_2| - |x_1| - |x_2|]^2 \tag{177}$$

## 4.178 Mishra N.11 [8]

A variant of the Mishra function with a range of $[-10, 10]$ and $f(x^*) = 2$. This function is also known as the arithmetic mean-geometric mean (AM-GM) function and is Continuous, Differentiable, Non-separable, Non-scalable, and Multimodal.

$$f(\mathbf{x}) = \left[\frac{1}{D}\sum_{i=1}^{D}|x_i| - \left(\prod_{i=1}^{D}|x_i|\right)^{\frac{1}{N}}\right]^2 \tag{178}$$

## 4.179 Mishra's Bird

A variant of the Mishra function with a range of $[-10, 10]$ and $(-3.1302, -1.5821)$ with $f(x^*) = 0$. This function is Continuous, Differentiable, Non-separable, Scalable, and Multimodal.

$$f(\mathbf{x}) = (x_0 - x_1)^2 + e^{(1-\sin(x_1))^2}\cos(x_0) + e^{(1-\cos(x_0))^2}\sin(x_1) \tag{179}$$

## 4.180 Muller Brown [65]

This function utilizes two exponential decay functions, each centered at distinct points on the $x$-$y$ plane, effectively creating two separate peaks. Each peak resembles a Gaussian bump, with one centered at $(1, 1)$ and the other at $(-2, -2)$. This configuration results in a landscape with clear high points at these centers. The function operates within the ranges of $[-1.5, 1]$ for $x$ and $[-0.5, 2.5]$ for $y$. A notable minimum for this function is found at $(-0.5528, 1.4417)$ with a function value of $f(x^*) = -146.6995$.

$$f(\mathbf{x}) = \sum_{j=1}^{4} A_j \exp\left[a_j(x_1 - x_{1,j}^0)^2 + b_j(x_1 - x_{1,j}^0)(x_2 - x_{2,j}^0) + c_j(x_2 - x_{2,j}^0)^2\right] \tag{180}$$

where the parameters are shown here:

| $j$ | $A_j$ | $a_j$ | $b_j$ | $c_j$ | $x^0_{1,j}$ | $x^0_{2,j}$ |
|---|---|---|---|---|---|---|
| 1 | -200 | -1.0 | 0.0 | -10 | 1.0 | 0.0 |
| 2 | -100 | -1.0 | 0.0 | -10 | 0.0 | 0.5 |
| 3 | -170 | -6.5 | 11 | -6.5 | -0.5 | 1.5 |
| 4 | 15.0 | 0.7 | 0.6 | 0.7 | -1.0 | 1.0 |

## 4.181 Parsopoulos [8, 67]

This function integrates cosine and sine functions applied to the squares of $x$ and $y$, respectively. The periodic nature of cosine and sine ensures that the function's behavior varies periodically, with the squared inputs amplifying the frequency and amplitude. The function operates within the range of $[-2, 2]$, with minima and maxima occurring based on how the values of $x^2$ and $y^2$ align with the natural peaks. The function's domain extends over $[-11, 11]$ for $x$ and $[-5, 5]$ for $y$. A notable minima occurs at values like $(k\pi/2, \pi)$, where $k$ is an integer, with $f(x^*) = 0$. This function is Continuous, Differentiable, Non-separable, Non-scalable, and Multimodal.

$$f(\mathbf{x}) = \sin\left(x_1^2\right) + \cos\left(x_0^2\right) \tag{181}$$

## 4.182 Pathological [7, 8]

This function features a sinusoidal component that is modified by a rational function that includes a quadratic form of $x$ and $y$ in the denominator. The sinusoidal oscillations introduced are then damped by this quadratic term, which reduces the amplitude of these oscillations as $x$ and $y$ values increase. The function operates within a broad range of $[-100, 100]$. A minimum occurs at $f(x^*) = -1.9960$. This function is Continuous, Differentiable, Non-separable, Non-scalable, and Multimodal.

$$f(\mathbf{x}) = \sum_{i=1}^{D-1}(0.5 + \frac{\sin^2\left(\sqrt{100x_0^2 + x_1^2}\right) - 0.5}{0.001\left(x_0^2 - 2x_0x_1 + x_1^2\right)^2 + 1}) \tag{182}$$

## 4.183 Paviani [8, 68]

This function incorporates logarithmic terms, which necessitate that the elements of $x$ are maintained within bounds greater than 2 and less than 10 (to prevent undefined behavior). The function's behavior is targeted at minimizing deviations from the interval $[2, 10]$ and is additionally complicated by the need to optimize the product term based on the geometric mean. This function operates within $[2.001, 9.999]$ and reaches a minimum of 9.3502 with $f(x^*) = -45.778$. This function is Continuous, Differentiable, Non-separable, Scalable, and Multimodal.

$$f(\mathbf{x}) = \sum_{i=1}^{10} \left[ (\ln(x_i - 2))^2 + (\ln(10 - x_i))^2 \right] - \left( \prod_{i=1}^{10} x_i \right)^{0.2} \quad (183)$$

## 4.184 Pen Holder [39, 46]

This function integrates cosine and sine functions under the influence of exponential and absolute terms, with additional modulation by the radial distance from the origin. This combination results in a landscape marked by sharp peaks and rapid decay, which results from the sensitivity of this function to the trigonometric synchronization of $x$ and $y$ as well as their distance from the origin. The range of the function spans $[-11, 11]$. A minimum occurs at $\pm 9.64616$ with $f(x^*) = -0.9635$. This function is Continuous, Differentiable, Non-separable, Non-scalable, and Multimodal.

$$f(\mathbf{x}) = -\exp\left[ |\cos(x_1)\cos(x_2)\exp\left(1 - \frac{(x_1^2 + x_2^2)^{0.5}}{\pi}\right)|^{-1} \right] \quad (184)$$

## 4.185 Penalty N.1 [8]

This function features a primary term that is a sinusoidal function of $x^2$, complemented by a summation term that includes a quadratic penalty influenced by another sinusoidal function. The sinusoidal components induce periodic fluctuations throughout the landscape, while the quadratic penalties target deviations from a value of 1. The parameter $k$ scales the impact of the combined quadratic-sinusoidal penalty term. While the function generally aims to be minimized when $x$ and $y$ are near 1, the influence of $k$ on the sinusoidal penalties means that local minima can appear elsewhere, dictated by the periodic nature of the sinusoidal terms. This function has a search space at $[-50, 50]$ and at 0 with $f(x^*) = -1$. This function is Continuous, Differentiable, and Non-separable.

$$f(\mathbf{x}) = \frac{\pi}{30} \left( 10\sin^2(\pi y_1) + \sum_{i=1}^{n-1} (y_i - 1)^2 \left(1 + 10\sin^2(\pi(y_{i+1}))\right) + (y_n - 1)^2 \right) + \sum_{i=1}^{n} u(x_i, 10, 100, 4) \quad (185)$$

where $y_i = 1 + \frac{1}{4}(x_i + 1)$,

and $u(x_i, a, k, m) = \begin{cases} k(x_i - a)^m & \text{if } x_i > a \\ 0 & \text{if } -a \leq x_i \leq a \\ k(-x_i - a)^m & \text{if } x_i < -a \end{cases}$

## 4.186 Penalty N.2 [80]

A variant of the Penalty function with a range of $[-50, 50]$ and $f(x^*) = 1$. This variant incorporates squared sinusoidal terms along with quadratic displacements that are modified by additional sinusoidal terms. This function is Continuous, Differentiable, and Non-separable.

$$f(\mathbf{x}) = 0.1\left(\sin^3(3\pi x_1) + \sum_{i=1}^{n-1}(x_i - 1)^2\left(1 + \sin^3(3\pi x_{i+1})\right) + (x_n - 1)^2\left(1 + \sin^2(2\pi x_n)\right)\right) + \sum_{i=1}^{n} u(x_i, 5, 100, 4) \tag{186}$$

$$\text{where} \quad \mathrm{u}(\mathrm{x}_i, a, k, m) = \begin{cases} k(x_i - a)^m & \text{if } x_i > a \\ 0 & \text{if } -a \le x_i \le a \\ k(-x_i - a)^m & \text{if } x_i < -a \end{cases}$$

## 4.187 Perez Reche

This function integrates simple sine and cosine functions to create a periodic and symmetrical landscape. Due to the inherent properties of these trigonometric functions, the landscape features smooth transitions between positive and negative values. This function has a space of $[-2, 2]$ with a function's minimum to occur between $(-1, 1)$.

$$f(\mathbf{x}) = \sin(x_0 + x_1) + \sin(x_0)\cos(x_1) \tag{187}$$

## 4.188 Periodic [4]

This function combines sinusoidal terms with a baseline that is modified by an exponential decay term influenced by the squares of $x$ and $y$. This setup results in a landscape where the deepest parts of the function are directly influenced by the strength of the exponential decay, primarily at or near the center. Given the range of $[-10, 10]$, this minimum is likely located at the origin $(0, 0)$, where the exponential decay's impact is maximized. This function is Continuous, Differentiable, Non-separable, Scalable, and Multimodal.

$$f(\mathbf{x}) = -0.1e^{-x_0^2 - x_1^2} + \sin^2(x_0) + \sin^2(x_1) + 1 \tag{188}$$

## 4.189 Perm [7, 8]

This unimodal function features a polynomial in $x$ and $y$ that includes a range of coefficients from linear to constant terms, complemented by nested sums and alternating signs to add complexity. The function operates within a domain of $[-d, d]$, where $d$ represents the degree or the number of dimensions. This function achieves a minimum value of $f(x^*) = 0$ at the points $(1, 2, \ldots, d)$.

$$f(\mathbf{x}) = \sum_{k=1}^{n} \sum_{i=1}^{n} \left[ \left( i^k + 0.5 \right) \left( x_i^k - 1 \right) \right]^2 \tag{189}$$

## 4.190 Perm N.1 [7, 8]

A variant of the Perm function.

$$f(\mathbf{x}) = 6.25 \left( x_0^3 x_1 + 0.6 x_0^2 x_1 + 0.2 x_0 x_1 + 0.6 \right)^2 + \\ 6.25 \left( x_0^3 x_1^2 + 0.6 x_0^2 x_1^2 + 0.2 x_0 x_1^2 + 0.6 \right)^2 + \\ 6.25 \left( x_0^3 + 0.6 x_0^2 + 0.2 x_0 + 0.6 \right)^2 \tag{190}$$

## 4.191 Perm Function 0,d beta

This is a unimodal function with search within a hypercube $[-d, d]^d$. The function is designed such that the global minimum, $f(x^*) = 0$, is located at $(1, 2, \ldots, d)$. The correct representation of the minimum in terms of the function's formula for a continuous domain is $(1, 12, \ldots, d)$.

$$f(\mathbf{x}) = \sum_{i=1}^{d} \left( \sum_{j=1}^{d} (j + \beta) \left( x_j^i - \frac{1}{j^i} \right) \right)^2 \tag{191}$$

## 4.192 Pinter N.1 [8]

This function has a search space of $[-10, 10]$, with 0 and $f(x^*) = 0$. This function is Continuous, Differentiable, Non-separable, Scalable, and Multimodal.

$$f(\mathbf{x}) = f_1(\mathbf{x}) + f_2(\mathbf{x}) + f_3(\mathbf{x}) \tag{192}$$

where:
$\mathrm{f}_1(\mathbf{X}) = \sum_{i=1}^{n} i x_i^2$
$f_2(\mathbf{X}) = \sum_{i=1}^{n-1} (x_{i-1} + 5\sin(x_i) + x_{i+1}^2)^2$
$f_3(\mathbf{X}) = \sum_{i=2}^{n} \ln \left[ 1 + i \left( \sin^2(x_{i-1}) + 2x_i + 3x_{i+1} \right)^2 \right]$
subject to:
$\mathrm{a}_i \le x_i \le b_i, \quad i = 1, 2, \ldots, n.$

Parameters as defined in the following table (same for Pinter N.2):

### 4.193 Pinter N.2 [8]

This function has a search space of $[-10, 10]$, with 0 and $f(x^*) = 0$.

$$f(\mathbf{x}) = f_1(\mathbf{x}) + f_2(\mathbf{x}) + f_3(\mathbf{x}) \tag{193}$$

where:
$\mathrm{f}_1(\mathbf{X}) = \sum_{i=1}^{n} i x_i^2$
$f_2(\mathbf{X}) = i \sin^2(x_i - \sin(x_i) - x_i + \sin(x_{i+1}))$
$f_3(\mathbf{X}) = i \ln \left[1 + i \left(x_{i-1}^2 - 2x_i + 3x_{i+1} - \cos(x_i) + 1\right)^2\right]$

Table 1: Parameters for Function f1

| $i$ | $a_i$ | $b_i$ | $i$ | $a_i$ | $b_i$ | $i$ | $a_i$ | $b_i$ | $i$ | $a_i$ | $b_i$ |
|---|---|---|---|---|---|---|---|---|---|---|---|
| 1 | -8.8 | 1.4 | 11 | -7.8 | 2.1 | 21 | -8.5 | 7.2 | 31 | -3.5 | 7.7 |
| 2 | -6.2 | 0.9 | 12 | -5.2 | 4.9 | 22 | -1.2 | 3.9 | 32 | -6.2 | 4.2 |
| 3 | 8.7 | 1.7 | 13 | -6.1 | 3.5 | 23 | 5.7 | 3.5 | 33 | 7.3 | 4.3 |
| 4 | -7.7 | 0.8 | 14 | -2.7 | 1.5 | 24 | -7.7 | 0.7 | 34 | -4.7 | 2.5 |
| 5 | 3.2 | 5.2 | 15 | 2.6 | 5.6 | 25 | -8.6 | 3.6 | 35 | -8.6 | 3.6 |
| 6 | -3.5 | 7.9 | 16 | -7.1 | 2.9 | 26 | -9.5 | 2.8 | 36 | -9.5 | 2.8 |
| 7 | 5.1 | 8.7 | 17 | -2.7 | 6.8 | 27 | -3.7 | 8.1 | 37 | -3.7 | 8.1 |
| 8 | -2.2 | 4.7 | 18 | -5.2 | 2.8 | 28 | -6.7 | 3.8 | 38 | -6.7 | 3.8 |
| 9 | 3.1 | 9.8 | 19 | -4.1 | 1.7 | 29 | -5.1 | 5.8 | 39 | -5.1 | 5.8 |
| 10 | -6.3 | 1.7 | 20 | -7.3 | 4.0 | 30 | -4.3 | 7.2 | 40 | -4.3 | 7.2 |
| 41 | -5.5 | 2.2 | 42 | -3.2 | 4.9 | 43 | -7.8 | 3.6 | 44 | -4.7 | 2.5 |
| 45 | 3.6 | -3.6 | 46 | -4.5 | 1.9 | 47 | 4.1 | -4.1 | 48 | -7.2 | 3.2 |
| 49 | -4.1 | 1.8 | 50 | -5.3 | 1.3 | | | | | | |

### 4.194 Plateau [8]

This function has a search space of $[-5.12, 5.12]$, with 0 and $f(x^*) = 30$.

$$f(\mathbf{x}) = 30 + \sum_{i=1}^{n} |x_i| \tag{194}$$

### 4.195 Powell [8]

This function has a search space of $[-4, 5]$, with $(3, -1, 0, 1)$, and $f(x^*) = 0$. This function is Continuous, Differentiable, Non-separable, Scalable, and Unimodal.

$$f(\mathbf{x}) = \sum_{i=1}^{d/4} \left[(x_{4i-3} + 10x_{4i-2})^2 + 5(x_{4i-1} - x_{4i})^2 + (x_{4i-2} - 2x_{4i-1})^4 + 10(x_{4i-3} - x_{4i})^4\right] \tag{195}$$

## 4.196 Powell Singular [69]

This function has a search space of $[-10, 10]$, with $(3, -1, 0, 1)$ and $f(x^*) = 0$. This function is also known as Powell's Quartic function. This function is Continuous, Differentiable, Non-separable, Scalable, and Unimodal.

$$f(\mathbf{x}) = (x_{i-1} + 10x_i)^2 + 5(x_{i+1} - x_{i+2})^2 + (x_i - 2x_{i+1})^4 + 10(x_{i-1} - x_{i+2})^4] \quad (196)$$

## 4.197 Powell's Badly Scaled Function

A variant of the Powell function with a range of $[-10, 10]$. This variant has a minima of zero at $(1.098 \times 10^{-5}, 9.106)$.

$$f(\mathbf{x}) = \left(10^4 x_1 x_2 - 1\right)^2 + \left(e^{-x_1} + e^{-x_2} - 1.0001\right)^2 \quad (197)$$

## 4.198 Fletcher-Powell's Helical Valley Function

A variant of the Powell function with a range of $[-100, 100]$. This variant has a minima of zero at (1, 0, 0).

$$f(\mathbf{x}) = 100\left[(x_3 - 100\theta(x_1, x_2))^2 + \left(\sqrt{x_1^2 + x_2^2} - 1\right)^2\right] + x_3^2 \quad (198)$$

where:

$$2\pi\theta(x_1, x_2) = \begin{cases} \arctan\left(\frac{x_2}{x_1}\right) & \text{if } x_1 \geq 0 \\ \pi + \arctan\left(\frac{x_2}{x_1}\right) & \text{otherwise} \end{cases}$$

## 4.199 Powell Sum [4]

A variant of the Powell Singular function with a range of $[-1, 1]$ with 0 and $f(x^*) = 0$. This function is Continuous, Differentiable, Separable, Scalable, and Multimodal.

$$f(\mathbf{x}) = \sum_{i=1}^{D} |x_i|^{i+1} \quad (199)$$

## 4.200 Power Sum [3, 8]

This is a unimodal function with a search space of $[0, d]$, with $(1, 2, 2, 3)$ and $f(x^*) = 0$.

$$f(\mathbf{x}) = \sum_{i=1}^{d} \left[ \left( \sum_{j=1}^{d} x_j^i \right) - b_i \right]^2 \tag{200}$$

The recommended value of the b -vector, for d = 4, is: b = (8, 18, 44, 114).

### 4.201 Price N.1 [8, 70]

This function has a search space of $[-10, 10]$, with $\pm 5$ and $f(x^*) = 0$. This function is also known as Becker-Lago's function. This function is Continuous, Non-differentiable, Separable, Non-scalable, and Multimodal.

$$f(\mathbf{x}) = (|x_0| - 5)^2 + (|x_1| - 5)^2 \tag{201}$$

### 4.202 Price N.2 [8, 70]

A variant of the Price function with a range of $[-10, 0]$ and at $(1, 2, 2, 3)$, $f(x^*) = 0$. This function is also known as the Periodic function. This function is Continuous, Differentiable, Non-separable, Non-scalable, and Multimodal.

$$f(\mathbf{x}) = -0.1e^{-x_0^2 - x_1^2} + \sin^2(x_0) + \sin^2(x_1) + 1 \tag{202}$$

### 4.203 Price N.3 [8, 70]

A variant of the Price function with a range of $[-5, 5]$ and at $((0.3413, 0.1164), (1, 1))$, $f(x^*) = 0$. This function is also known as the Modified Rosenbrock's or Price-Rosenbrock's function. This function is Continuous, Differentiable, Non-separable, Non-scalable, and Multimodal.

$$f(\mathbf{x}) = \left(100(x_2 - x_1^2)^2 + 6.4((x_2 - 0.5)^2 - x_1 - 0.6)^2\right) \tag{203}$$

### 4.204 Price N.4 [7, 8]

A variant of the Price function with a range of $[-500, 500]$ and at $(0, 0)$, $(2, 4)$, $(1.4643, 2.5060)$, $f(x^*) = 0$. This function is Continuous, Differentiable, Separable, Scalable, and Multimodal.

$$f(\mathbf{x}) = \left(2x_0^3 x_1 - x_1^3\right)^2 + (6x_0 - x_1^2 + x_1)^2 \tag{204}$$

### 4.205 Qing [71]

This function is simple but effective for testing optimization algorithms that can efficiently find minimum values in a straightforward but potentially large-dimensional space of $[-500, 500]$. This function is 0 with $f(x^*) = 0$. This function is Continuous, Differentiable, Separable, Scalable, and Multimodal.

$$f(\mathbf{x}) = \sum_{i=1}^{D} \left(x_i^2 - i\right)^2 \tag{205}$$

## 4.206 Qing Variant

A variant of the Qing function for lower dimensions with a range of $[-500, 500]$ and at 0, $f(x^*) = 0$.

$$f(x, for2D) = \left(x_0^2 - 1\right)^2 + \left(x_1^2 - 2\right)^2 \tag{206}$$

## 4.207 Qing N.2

A variant of the Qing function with a range of $[-500, 500]$ and at 0, $f(x^*) = 0$. This variant combines a slowly increasing quadratic term scaled down by 4000, subtractive cosine terms that introduce oscillations and periodicity, and additive sine squared terms that penalize larger values of $x$ and $y$.

$$f(\mathbf{x}) = \sin^2(x_0) + \sin^2(x_1) - \cos\left(\frac{\sqrt{2}x_1}{2}\right)\cos(x_0) + \frac{x_0^2}{4000} + \frac{x_1^2}{4000} + 1 \tag{207}$$

## 4.208 Quadratic [7, 8]

This function combines quadratic terms with a bilinear term $xy$. The quadratic elements typically promote a parabolic shape to influence the function to exhibit curvatures along both axes, while the bilinear term ties $x$ and $y$ to yield complex features (e.g., saddle points or modified curvatures). The function is defined within a range of $[-1.28, 1.28]$ and $[-10, 10]$, with an observed minimum at $(0.1938, 0.4851)$ and $f(x^*) = -3873.7241$. This function is Continuous, Differentiable, Non-separable, Non-scalable, and Multimodal.

$$f(\mathbf{x}) = 128.018x_0^2 + 182.25x_0x_1 + 138.082x_0 + 203.64x_1^2 - 232.92x_1 - 3803.84 \tag{208}$$

## 4.209 Quartic [72]

This function is defined by dominant quartic terms in $x^4$ and $y^4$, which ensure that the function values escalate rapidly as $x$ and $y$ increase. This function further complicates its landscape by using random coefficients applied to $x$ and $y$. This randomization renders the function non-deterministic, making consistent optimization more challenging as the landscape can shift unpredictably with every computation. This function operates within a narrow range of $[-1.28, 1.28]$ and ideally achieves a minimum at the origin with $f(x^*) = 0$. This function is Continuous, Differentiable, Separable, and Scalable.

$$f(\mathbf{x}) = \sum_{i=1}^{D} \left(ix_i^4\right) \tag{209}$$

$$f(x, for2D) = x_0^4 + \frac{x_0}{2} + x_1^4 + \frac{x_1}{2}$$

## 4.210 Quartic Variant [72]

A variant of the Quartic function with noise and a range of $[-1.28, 1.28]$ and at $0$, $f(x^*) = 0$.

$$f(\mathbf{x}) = \sum_{i=1}^{D} \left(ix_i^4\right) + random[0, 1] \tag{210}$$

## 4.211 Quintic [49]

This function has quintic (i.e., fifth degree) polynomials in both $x$ and $y$ and has highly nonlinear behavior. This results in a landscape rich with peaks and valleys. This function has a search space of $[-10, 10]$ and achieves a critical minimum at $[-1,\ 2]$. This function is Continuous, Differentiable, Separable, Non-scalable, and Multimodal.

$$f(\mathbf{x}) = \sum_{i=1}^{D} \left|x_0^5 - 3x_0^4 + 4x_0^3 + 2x_0^2 - 10x_0 - 4\right| \tag{211}$$

$$f(\mathbf{x}) = x_0^5 - 4x_0^3 + 2x_0^2 - 10x_0 + x_1^5 - 4x_1^3 + 2x_1^2 - 10x_1 - 8$$

## 4.212 Rana [8]

This function incorporates a complex and iterative trigonometric mechanism that employs trigonometric functions on the square roots of absolute differences and sums of these pairs. Due to the trigonometric functions' sensitivity to input variations, minor changes in the elements of $x$ can lead to significant alterations in the function's output. This function operates within a broad range of $[-500, 500]$, and numerous local minima and maxima characterize its landscape. A notable minimum is observed at $x = -500$ with $f(x^*) = -928.5478$. This function is Continuous, Differentiable, Non-separable, Scalable, and Multimodal.

$$f(\mathbf{x}) = (x_{i+1} + 1) \sin\left(\sqrt{|x_{i+1} + x_i + 1|}\right) \cos\left(\sqrt{|x_{i+1} - x_i + 1|}\right) + (x_i) \sin\left(\sqrt{|x_{i+1} - x_i + 1|}\right) \cos\left(\sqrt{|x_{i+1} + x_i + 1|}\right) \tag{212}$$

## 4.213 Rastrigin [73]

This is a non-separable and multimodal function with a large number of local minima. This function has a typical search domain defined within the range $[-5.12, 5.12]$, and a global minimum of the function is at $x = 0$, where $f(x^*) = 0$. There are a number of variants to this function.

$$f(\mathbf{x}) = 10d + \sum_{i=1}^{d} \left[x_i^2 - 10\cos(2\pi x_i)\right] \tag{213}$$

$$f(x, for2D) = -10\cos(2\pi x_0) - 10\cos(2\pi x_1) + x_0^2 + x_1^2 + 20$$

## 4.214 Rastrigin Modified

This function combines quadratic components with a cosine function to create a dual-layered landscape. The quadratic terms provide a globally convex shape, and the cosine terms superimpose a series of local minima and maxima. The function operates within a range of $[-5.12, 5.12]$, with a global minimum at $x = 0$, where $f(x^*) = 0$. This function is also known as the Extended Rastrigin function.

$$f(x, for2D) = -10\cos(2\pi x_0) - 10\cos(2\pi x_1) + x_0^2 + x_1^2 + 20 \tag{214}$$

## 4.215 Rayleigh

This function combines growth and decay in such a way that it initially decreases from the origin, potentially reaching a minimum where the linear growth and exponential decay balance each other before increasing again at larger distances. This is due to the combination of the radial increase and the exponential decay, which compete against each other as the distance from the origin increases. The function operates within a range of $[-5, 5]$.

$$f(x, for2D) = \left(x_0^2 + x_1^2\right)^{0.5} - e^{-x_0^2 - x_1^2} \tag{215}$$

## 4.216 Ridge

This is a straightforward quadratic function where $x$ has a full influence, and $y$ has a reduced influence due to its coefficient of 0.1. This makes the function grow faster with changes in $x$ than with changes in $y$. The function is convex and increases as $x$ and $y$ move away from the origin. The function operates within a range of $[-5, 5]$.

$$f(x, for2D) = x_0^2 + 0.1x_1^2 \tag{216}$$

## 4.217 Ripple [8]

The function can be quite volatile due to the high-frequency cosine and the strong power of the sine term. Its range is complex due to these oscillations and depends significantly on the distribution and values of $x$ and can be represented in $[0, 1]$. with a global minimum at $x = 0.1$, where $f(x^*) = -2.2$. This function is Non-separable.

$$f(\mathbf{x}) = \sum_{i=1}^{2} \left[ -e^{-2\ln 2\left(\frac{x_i - 0.1}{0.8}\right)^2} \left(\sin^6(5\pi x_i) + 0.1\cos^2(500\pi x_i)\right) \right] \tag{217}$$

## 4.218 Ripple N.25 [8]

A variant of the Ripple function with a range of $[0, 1]$ and at 0.1, $f(x^*) = -2$. This function has one global and 252,004 local minima. The global form of the function consists of 25 holes, which form a 5 by 5 regular grid [42]. This function is Non-separable.

$$f(\mathbf{x}) = \sum_{i=1}^{2} \left[ -e^{-2\ln 2\left(\frac{x_i - 0.1}{0.8}\right)^2} \sin^6(5\pi x_i) \right] \tag{218}$$

$$f(x, for2D) = -2^{-3.125(x_1 - 0.1)^2} \sin^6(5\pi x_1) - 2^{-3.125(x_0 - 0.1)^2} \sin^6(5\pi x_0)$$

## 4.219 Rosenbrock [25, 73, 74]

This function is characterized by a long, narrow, parabolic-shaped valley, which makes this function non-convex with steep ridges. While identifying the valley where the minimum lies is relatively straightforward, converging to the global optimum can be difficult due to the function's intricate topology. This function's landscape is especially challenging for gradient descent methods, which may struggle with the steep gradients without precise initial guesses or sophisticated adaptive steps. The recommended search areas for this function vary, with typical bounds set at $[-5, 5]$, $[-5, 10]$, $[-2.048, 2.048]$, to name a few. The global minimum of the function is achieved at the point $(1, 1, \ldots, 1)$ with $f(x^*) = 0$. This function is also known as the Fletcher function, the Leon function, the Rosenbrock's Valley, or the Banana function. This function is Continuous, Differentiable, Non-separable, Scalable, and Unimodal.

$$f(\mathbf{x}) = \sum_{i=1}^{D-1} \left[ 100(x_{i+1} - x_i^2)^2 + (x_i - 1)^2 \right] \tag{219}$$

$$f(x, for2D) = 100\left(x_1 - x_0^2\right)^2 + (x_0 - 1)^2$$

### 4.220 Rosenbrock Modified [70]

A variant of the Rosenbrock function that has been modified to contain more than one global minimum. It has four regularly spaced global minima and 96 local minima [42]. This function has a range of $[-5, 2.5]$ and at $(-0.9, -0.95)$ and $f(x^*) = 34.37$. This function is Continuous, Differentiable, Non-separable, Non-scalable, and Multimodal.

$$f(\mathbf{x}) = 74 + 100(x_2 - x_1^2)^2 + (1 - x_1)^2 - 400e^{-\frac{(x_1+1)^2+(x_2+1)^2}{0.1}} \tag{220}$$

### 4.221 Rosenbrock with additional sine terms [4, 66]

This function integrates the classic Rosenbrock function with additional sinusoidal terms that introduce periodic variations to the landscape. The primary behavior of this enhanced Rosenbrock function retains the characteristic valley leading to a global minimum, typically near $(1, 1)$. However, the added sinusoidal terms can alter the exact location and the minimum value by introducing fluctuations in the function's gradient. This function operates within a domain of $[-10, 10]$ with a global minimizer at $(1, \ldots, 1)$.

$$f(\mathbf{x}) = 100 \cdot (y - x^2)^2 + (1 - x)^2 + \sin(2\pi x) + \sin(2\pi y) \tag{221}$$

### 4.222 Rotated Ellipse [4, 8]

This function has a range between $[-500, 500]$ and at 0 and $f(x^*) = 0$. This function is Continuous, Differentiable, Non-separable, Non-scalable, and Unimodal.

$$f(\mathbf{x}) = 7x_0^2 - 6\sqrt{3}x_0x_1 + 13x_1^2 \tag{222}$$

### 4.223 Rotated Ellipse N.2 [4, 8]

A variant of the Rotated Ellipse function with a range of $[-500, 500]$ and at 0, $f(x^*) = 0$. This function is Continuous, Differentiable, Non-separable, Non-scalable, and Unimodal.

$$f(\mathbf{x}) = x_0^2 - x_0x_1 + x_1^2 \tag{223}$$

### 4.224 Rotated Hyper Ellipsoid [3, 73]

This function is an extension of the Axis Parallel Hyper-Ellipsoid function, also referred to as the Sum Squares function. It has a range between $[-65.536, 65.536]$.

$$f(\mathbf{x}) = \sum_{i=1}^{d-1}\sum_{j=1}^{i}(x_j)^2 \tag{224}$$

### 4.225 Rump [4]

This function is defined by high-degree polynomial terms. The presence of terms like $y^6$ and $y^8$ heavily influences the function's behavior as $y$ increases, causing these terms to dominate and drastically shape the landscape. Additionally, the term $\frac{x}{2y}$ introduces a singularity at $y = 0$, adding complexity to the landscape. This function has a search space of $[-500, 500]$, and at 0 and $f(x^*) = 0$. This function is Continuous, Differentiable, Non-separable, Non-scalable, and Unimodal.

$$f(\mathbf{x}) = \left(333.75 - x_0^2\right) x_1^6 + \left(11x_0^2x_1^2 - 121x_1^4 - 2\right) x_0^2 + \frac{x_0}{2x_1} + 5.5x_1^8 \tag{225}$$

### 4.226 Salomon [4, 73]

This function exhibits radial symmetry that is determined by the distance $r$ from the origin. The incorporation of a cosine term adds a periodic oscillation. Additionally, a linear term, $0.1r$, is integrated to ensure that the overall function value gradually increases as the distance from the origin grows and prevents the function value from becoming overly negative at larger radii. The specified range for analysis extends from $[-20, 20]$ to $[-100, 100]$, accommodating a wide spread of possible $r$ values, and theoretically, the function reaches its minimum at $f(x^*) = 0$. This function is Continuous, Differentiable, Non-separable, Scalable, and Multimodal.

$$f(\mathbf{x}) = 1 - \cos\left(2\pi\sqrt{\sum_{i=1}^{D} x_i^2}\right) + 0.1\sqrt{\sum_{i=1}^{D} x_i^2} \tag{226}$$

### 4.227 Sargan [8]

The function is non-negative and can grow depending on both the individual values and their interactions. The function has a range of $[-100, 100]$ and at 0 and $f(x^*) = 0$. This function is Continuous, Differentiable, Non-separable, Scalable, and Multimodal.

$$f(\mathbf{x}) = -\sum_{i=1}^{D}\left(Dx_i^2 + 0.4\sum_{j\neq i} x_i x_j\right) \tag{227}$$

### 4.228 Sawtooth [4]

This function stems from a piecewise linear (PWL) signal where each segment has the same negative slope. This translates to the visual appearance of a "sawtooth" shape when plotted in time [75]. The function has a range of $[-20, 20]$ and at 0 and $f(x^*) = 0$.

$$f(\mathbf{x}) = g(r) \cdot h(t) \tag{228}$$

where:
g(r) $= \sin(r) - \frac{\sin(2r)}{2} + \frac{\sin(3r)}{3} - \frac{\sin(4r)}{4} + \left(\frac{r^2}{r+1}\right)$
$h(t) = \frac{1}{2}\cos(2t-1) + \cos(t) + 2$
$r = \sqrt{x_1^2 + x_2^2}$
$t = 2(x_2, x_1)$

### 4.229 Schaffer N.1 [46, 51]

This function has a pronounced peak at the origin, where it attains its global minimum. This peak is surrounded by multiple local minima formed from the function's sinusoidal component, which creates oscillations in the landscape. Further, the function features sharp ridges and deep valleys, which result from the square root term and the sinusoidal squaring in the formula. As $x$ and $y$ increase in magnitude, the value of the function tends to stabilize due to the division by the squared term in the denominator, which grows faster than the numerator. The function has a range of $[-100, 100]$ and at 0 and $f(x^*) = 0$. This function is Continuous, Differentiable, Non-separable, Non-scalable, and Unimodal.

$$f(\mathbf{x}) = \frac{\sin^2\left(x_0^2 + x_1^2\right) - 0.5}{\left(0.001x_0^2 + 0.001x_1^2 + 1\right)^2} + 0.5 \tag{229}$$

### 4.230 Schaffer N.2 [46, 76]

A variant of the Schaffer function with a range of $[-100, 100]$ and at 0, $f(x^*) = 0$. This function is Continuous, Differentiable, Non-separable, Non-scalable, and Unimodal.

$$f(\mathbf{x}) = \frac{\sin^2\left(x_0^2 - x_1^2\right) - 0.5}{\left(0.001x_0^2 + 0.001x_1^2 + 1\right)^2} + 0.5 \tag{230}$$

### 4.231 Schaffer N.3 [8, 46]

A variant of the Schaffer function with a range of $[-100, 100]$ and at $(0, 1.2531)$, $f(x^*) = 0.00156$. This function demonstrates how the trigonometric transformations combined with a decay factor create a landscape with multiple peaks and valleys that are dependent on the squared differences $x^2 - y^2$. The function value is designed to fluctuate around 0.5, with deviations modulated both by the trigonometric transformation and the decaying quadratic term in the denominator. This function is Continuous, Differentiable, Non-separable, Non-scalable, and Unimodal.

$$f(\mathbf{x}) = \frac{\sin^2\left(\cos\left(x_0^2 - x_1^2\right)\right) - 0.5}{\left(0.001x_0^2 + 0.001x_1^2 + 1\right)^2} + 0.5 \tag{231}$$

### 4.232 Schaffer N.4 [[8, 10, 46]

A variant of the Schaffer function with a range of $[-100, 100]$ and at $(0, 1.2531)$, $f(x^*) = 0.2925$. This function is Continuous, Differentiable, Non-separable, Non-scalable, and Unimodal.

$$f(\mathbf{x}) = \frac{\cos^2\left(\sin\left(x_0^2 - x_1^2\right)\right) - 0.5}{\left(0.001x_0^2 + 0.001x_1^2 + 1\right)^2} + 0.5 \tag{232}$$

### 4.233 Schaffer N.7 [25]

A variant of the Schaffer function with a range of $[-100, 100]$ and a minima at 0.

$$f(\mathbf{x}) = \left(x_1^2 + x_2^2\right)^{0.25}\left[50\left(x_1^2 + x_2^2\right)^{0.1} + 1\right] \tag{233}$$

### 4.234 Schaffer F6 [7, 46]

A variant of the Schaffer function with a range of $[-100, 100]$ and at 0, $f(x^*) = 0$. This variant is Continuous, Differentiable, Non-separable, Scalable, and Multimodal and is also known as the Expanded Schaffer's F6.

$$f(\mathbf{x}) = \frac{\sin^2\left(\sqrt{x_0^2 + x_1^2}\right) - 0.5}{1 + 0.001\left(x_0^2 + x_1^2\right)^2} + 0.5 \tag{234}$$

### 4.235 Schaffer F7 [29]

A variant of the Schaffer function that is asymmetric. This variant involves a more complex mathematical structure with non-linear transformations and potential optimization parameters.

$$f(\mathbf{x}) = \left(\frac{1}{D-1}\sum_{i=1}^{D-1}\sqrt{s_i} + \sqrt{s_i}\sin^2\left(50s_i^{1/5}\right)\right)^2 + 10f_{\text{pen}}(x) + f_{\text{opt}} \tag{235}$$

where:

$\mathrm{z} = \Lambda^{10}Q^{T_{\text{asy}}}(R(x - x^{\text{opt}}))$

$s_i = \sqrt{x_i^2 + x_{i+1}^2}$ for $i = 1, \ldots, D$

### 4.236 Schaffer F7 N.1 [29]

A variant of the Schaffer function that is moderately ill-conditioned. This variant features a non-linear transformation of the variables combined with sinusoidal modulations. The sin term adds a complex, oscillatory behavior based on the scaled variables, which are the Euclidean distances between successive pairs of decision variables.

$$f(\mathbf{x}) = \left( \frac{1}{D-1} \sum_{i=1}^{D-1} \sqrt{s_i} + \sqrt{s_i} \sin^2 \left( 50 s_i^{1/5} \right) \right)^2 + 10 f_{\text{pen}}(x) + f_{\text{opt}} \tag{236}$$

where:
$\text{z} = \Lambda^{1000} Q^{T_{\text{asy}}} (R(x - x^{\text{opt}}))$
$s_i = \sqrt{x_i^2 + x_{i+1}^2} \quad \text{for } i = 1, \ldots, D$

### 4.237 Schmidt Vetters [4, 8]

The function displays intricate patterns due to the interaction between the sine, reciprocal, and exponential terms affecting the output. This function has a range of $[0, 10]$, and at 0.7854, $f(x^*) = 3$. This function is Continuous, Differentiable, Non-separable, Non-scalable, and Multimodal.

$$f(\mathbf{x}) = \sin\left( \frac{\pi x_1}{2} + \frac{x_2}{2} \right) + e^{-\frac{(x_0 + x_1 - x_2)^2}{2}} + \frac{1}{(x_0 - x_1)^2 + 1} \tag{237}$$

### 4.238 Schumer Steiglitz [58]

The function is non-negative and grows as any component of $x$ increases. It is minimized (function value is zero) when all components of $x$ are zero. This function has a range between $[-100, 100]$, with 0 and $f(x^*) = 0$.

$$f(\mathbf{x}) = \sum_{i=1}^{D} (x_i)^4 \tag{238}$$

### 4.239 Schwefel [73, 80]

The structure of this function is particularly deceptive because the global minimum is located far from other local minima across the parameter space. This function has multiple ranges and variants varying between $[-500, 500]$ to smaller intervals like $[-100, 100]$. This function's global minimum is located at 420.9687 with $f(x^*) = 0$. This function is Continuous, Differentiable, Partially separable, Scalable, and Unimodal. It is worth noting that the number of the Schwefel variants is significant and hence only a few are presented herein.

$$f(\mathbf{x}) = -\sin\left( \sqrt{|x_0|} \right) x_i \tag{239}$$

$$f(\mathbf{x}, \text{with dimensional shift}) = 418.9829d - \sin\left(\sqrt{|x_0|}\right)x_i$$

$$f(x, for2D) = -\sin\left(\sqrt{|x_0|}\right)x_0 - \sin\left(\sqrt{|x_1|}\right)x_1 + 837.9658$$

### 4.240 Schwefel 1.2 [7, 80]

This function computes the cumulative sum up to each element in the vector $x$, squares this sum, and then sums these squared values. This procedure effectively magnifies any deviations early in the vector, as earlier values influence more terms in the sum. This is a variant of the Schwefel function with a range of $[-100, 100]$ and at 0, $f(x^*) = 0$. This variant is Continuous, Differentiable, Non-separable, Scalable, and Multimodal, and is also called the Expanded Schaffer's F6. This function is also known as the Shifted Schwefel's function, Double-Sum or Rotated Hyper-Ellipsoid function.

$$f(\mathbf{x}) = \sum_{i=1}^{n}(\sum_{j=1}^{i} x_j)^2 \tag{240}$$

### 4.241 Schwefel 1.2 N.1

This is a variant of the Schwefel 1.2 function with the addition of noise.

$$f(\mathbf{x}) = \sum_{i=1}^{n}(\sum_{j=1}^{i} x_j)^2(1 + 0.4|N(0,1)|) \tag{241}$$

### 4.242 Schwefel 2.20 [80]

This is a variant of the Schwefel function with a range of $[-100, 100]$ and at 0, $f(x^*) = 0$. This variant comprises a simple summation of the absolute values of the elements in the vector $x$. It measures the Manhattan norm of the vector, or the $L_1$ norm, which is often used in optimization to enforce sparsity. The function is always non-negative, linearly increasing with the absolute values of the elements in $x$. This variant is Continuous, Non-differentiable, Separable, scalable, and Unimodal.

$$(\mathbf{x}) = \sum_{i=1}^{d} |x_i| \tag{242}$$

### 4.243 Schwefel 2.21 [7, 80]

This is a variant of the Schwefel function that is also Convex, Continuous, and Unimodal.

$$f(\mathbf{x}) = \max_{1 \le i \le n} |x_i| \tag{243}$$

### 4.244 Schwefel 2.22 [7, 80]

This is a variant of the Schwefel function with a range of $[-10, 10]$. This variant combines a linear summation and a multiplicative product of the absolute values of elements in $x$. As such, this variant is always non-negative and is influenced heavily by both the sum and the product of absolute values. If any component of $x$ is zero, the product term becomes zero, but the sum term ensures that the function value remains non-negative. This variant is Convex, Continuous, Separable, and Unimodal.

$$f(\mathbf{x}) = \sum_{i=1}^{n} |x_i| + \prod_{i=1}^{n} |x_i| \tag{244}$$

### 4.245 Schwefel 2.23 [80]

This is a variant of the Schwefel function with a range of $[-10, 10]$. The variant grows rapidly with the increase in any component's absolute value. It is non-negative and symmetric around zero for each variable. Due to the high power of 10, this variant can be extremely sensitive to large values and outliers. This variant is Continuous, Differentiable, Non-separable, Non-scalable, and Multimodal.

$$f(\mathbf{x}) = \sum_{i=1}^{d} {x_i}^{10} \tag{245}$$

### 4.246 Schwefel 2.25 [80]

This is a variant of the Schwefel function with a range of $[-100, 100]$. This variant includes a sum of squared deviations from 1 for each component and a circular term linking the first and last elements. This promotes uniformity towards one across all components and continuity between the start and end of the vector $x$ (which makes this variant suitable for scenarios where cyclic or continuity conditions are required). This variant is Continuous,

$$f(\mathbf{x}) = \sum_{i=1}^{n} \left((x_i - 1)^2 + (x_1 - x_i^2)^2\right) \tag{246}$$

### 4.247 Schwefel 2.26 [80]

This is a variant of the Schwefel function with a range of $[-500, 500]$. This function applies a sinusoidal function modulated by the square root of the absolute values of each element in $x$, multiplied by the elements themselves. The negative sign inverts the typical sinusoidal output, potentially creating a landscape with sharp peaks and valleys depending on the values of $x$. The function can exhibit both positive and negative values due to this inversion. This variant is Continuous, Differentiable, Separable, Non-scalable, and Multimodal.

$$f(\mathbf{x}) = -1/d \sum_{i=1}^{d} x_i \sin(\sqrt{|x_i|}) \tag{247}$$

## 4.248 Schwefel 2.36 [80]

This is a variant of the Schwefel function with a range of $[0, 500]$. This variant is non-linear and its output significantly depends on the relationships and magnitudes of the first three vector elements. This variant is Continuous, Differentiable, Separable, Scalable, and Multimodal.

$$f(\mathbf{x}) = -x_1(2x_2 - 1 - 2x_2^2) \tag{248}$$

## 4.249 Schwefel 2.40 [80]

This is a variant of the Schwefel function with a minima of -5000.

$$f(\mathbf{x}) = -\sum_{i=1}^{5} x_i \tag{249}$$

## 4.250 Schwefel 2.60 [80]

This is a variant of the Schwefel function with a range of $[-100, 100]$. This variant finds the maximum of two absolute linear combinations. This variant is Continuous, Non-differentiable, Separable, Scalable, and Unimodal.

$$f(\mathbf{x}) = \max\left(|x_0 + 2x_1 - 7|, |2x_0 + x_1 - 5|\right) \tag{250}$$

## 4.251 Schwefel N.7 [80]

This is a variant of the Schwefel function with a range of $[-500, 500]$.

$$f(\mathbf{x}) = \sum_{i=1}^{n} -x_i \sin\left(\sqrt{|x_i|}\right) \tag{251}$$

## 4.252 Shekel [37]

This is a single objective function with a range of [0, 10] and multiple minima located at different m-values.

$$-\sum_{i=1}^{m}\left(\sum_{j=1}^{n}(x_{ij}-a_{ij})^2+c_j\right)^{-1} \tag{252}$$

where $c_i$ (for $i = 1, \ldots, m$), $a_{ij}$ (for $j = 1, \ldots, n$ and $i = 1, \ldots, m$) are constant numbers fixed in advance. It is recommended to set $m = 30$.

## 4.253 Shekel N.5 [37]

This is a variant of the Shekel function with a range of [0, 10] and at 4, $f(x^*) = -10.1499$. This variant has multiple minima and is Continuous, Differentiable, Non-separable, Scalable, and Multimodal.

$$f(\mathbf{x}) = -\sum_{i=1}^{5}\frac{1}{\left(\sum_{j=1}^{4}(x_j-a_{ij})^2\right)+c_i} \tag{253}$$

$$\text{where } A = [a_{ij}] = \begin{bmatrix} 4 & 4 & 4 & 4 \\ 1 & 1 & 1 & 1 \\ 8 & 8 & 8 & 8 \\ 6 & 6 & 6 & 6 \\ 3 & 7 & 3 & 7 \end{bmatrix} \text{ and } c = [c_i] = \begin{bmatrix} 0.1 \\ 0.2 \\ 0.2 \\ 0.4 \\ 0.4 \end{bmatrix}.$$

## 4.254 Shekel N.7 [37]

This is a variant of the Shekel function with a range of [0, 10] and at 4, $f(x^*) = -10.3999$. This variant has multiple minima and is Continuous, Differentiable, Non-separable, Scalable, and Multimodal.

$$f(\mathbf{x}) = -\sum_{i=1}^{7}\frac{1}{\left(\sum_{j=1}^{4}(x_j-a_{ij})^2\right)+c_i} \tag{254}$$

$$\text{where } A = [a_{ij}] = \begin{bmatrix} 4 & 4 & 4 & 4 \\ 1 & 1 & 1 & 1 \\ 8 & 8 & 8 & 8 \\ 6 & 6 & 6 & 6 \\ 3 & 7 & 3 & 7 \\ 2 & 9 & 2 & 9 \\ 5 & 5 & 3 & 3 \end{bmatrix} \text{ and } c = [c_i] = \begin{bmatrix} 0.1 \\ 0.2 \\ 0.2 \\ 0.4 \\ 0.4 \\ 0.6 \\ 0.3 \end{bmatrix}.$$

## 4.255 Shekel N.10 [37]

This is a variant of the Shekel function with a range of $[0, 10]$ and at 4, $f(x^*) = -10.5319$. This variant is predominantly negative and decreases sharply in the negative direction as $x$ approaches any of its vectors, creating potential wells around these points. This variant is Continuous, Differentiable, Non-separable, Scalable, and Multimodal.

$$f(\mathbf{x}) = -\sum_{i=1}^{10} \left( \sum_{j=1}^{4} \frac{1}{(x_j - a_{ij})^2 + c_i} \right) \tag{255}$$

where

$$[a_{ij}] = \begin{bmatrix} 4 & 4 & 4 & 4 \\ 1 & 1 & 1 & 1 \\ 8 & 8 & 8 & 8 \\ 6 & 6 & 6 & 6 \\ 3 & 7 & 3 & 7 \\ 2 & 9 & 2 & 9 \\ 5 & 5 & 3 & 3 \\ 8 & 1 & 8 & 1 \\ 6 & 2 & 6 & 2 \\ 7 & 3.6 & 7 & 3.6 \end{bmatrix}, \quad c_i = \begin{bmatrix} 0.1 \\ 0.2 \\ 0.2 \\ 0.4 \\ 0.4 \\ 0.6 \\ 0.3 \\ 0.7 \\ 0.5 \\ 0.5 \end{bmatrix}$$

## 4.256 Shubert [10, 76]

This function is constructed from a series of cosine terms, each multiplied by an increasing integer for each component of $x$, and then aggregates these results through the product of sums across all components. The inclusion of multiple frequencies and phases within the cosine terms generates complex oscillations within each dimension. Due to the product of the sums mechanism, small variations in $x$ can lead to significant changes in the function's landscape. The function exhibits a particularly challenging optimization landscape, featuring 18 global minima grouped into nine pairs, forming a 3 by 3 square grid pattern, and 742 local minima. The function operates within the domain of $[-10, 10]$ and achieves a minimum value of $f(x^*) = -186.7309$ [42]. This function is Continuous, Differentiable, Non-scalable, and Multimodal.

$$f(\mathbf{x}) = \prod_{i=1}^{2} \left( \sum_{j=1}^{5} \cos\left((j+1)x_i + j\right) \right) \tag{256}$$

## 4.257 Shubert N.3

This is a variant of the Shubert function with a range of $[-10, 10]$. This variant is also known as the Trigonometric Polynomial or Suharev-Zilinskas' Function. This variant has multiple minima and is Continuous, Differentiable, Separable, Non-scalable, and Multimodal.

$$f(\mathbf{x}) = \left( \sum_{i=1}^{D} \sum_{j=1}^{5} j \sin\left((j+1)x_i + j\right) \right) \tag{257}$$

## 4.258 Shubert N.4

This is a variant of the Shubert function with a range of $[-10, 10]$. This variant has multiple minima and is Continuous, Differentiable, Separable, Non-scalable, and Multimodal.

$$f(\mathbf{x}) = \left( \sum_{i=1}^{D} \sum_{j=1}^{5} j \cos\left((j+1)x_i + j\right) \right) \tag{258}$$

## 4.259 Sine Cosine and half

In this function, the combination of sine and cosine terms leads to oscillations with the distance from the origin. These are modulated by a growing denominator to create a damping effect as $x$ and $y$ increase. This results in a diminishing amplitude of oscillation with increasing distance from the origin. The function operates within the domain of $[-5, 5]$ and achieves a minimum value of $f(x^*) = 0$.

$$f(\mathbf{x}) = \frac{\sin\left(\sqrt{x_0^2 + x_1^2}\right) - \cos\left(\sqrt{x_0^2 + x_1^2}\right)}{0.001x_0^2 + 0.001x_1^2 + 1} + 0.5 \tag{259}$$

## 4.260 Sine Envelope Sine Wave [46]

In this function, the squaring of the sine function ensures that the output is always non-negative (post subtraction of 0.5), and the rapidly growing denominator reduces the impact of the numerator as $x$ and $y$ increase. This leads to a smooth decay towards 0.5. The function operates within the domain of $[-100, 100]$ and achieves a minimum value of $f(x^*) = 0$. This function is also known as the Schaffer function (m=2).

$$f(\mathbf{x}) = \sum_{i=1}^{n-1} \left( \frac{\sin^2\left(x_0^2 + x_1^2\right)^{0.5} - 0.5}{\left(0.001x_0^2 + 0.001x_1^2 + 1\right)^2} \right) + 0.5 \tag{260}$$

## 4.261 SODP [8]

The function is always non-negative and grows as $x$ and $y$ move away from zero (due to the cubic and quartic power terms, with $y^4$ contributing more significantly to the rapid increase in function values. The function operates within the domain of $[-1, 1]$ and achieves a minimum value of $f(x^*) = 0$.

$$f(\mathbf{x}) = \sum_{i-1}^{n} |x_i|^{i+1} \tag{261}$$

$$f(x, for2D) = |x_0|^3 + |x_1|^4$$

## 4.262 Sphere [77]

The Sphere function is one of the simplest optimization test functions. It is continuous, convex, unimodal, differentiable, separable, highly symmetric, and rotationally invariant. The function has a space search area within $[-100, 100]$ and/or $[-5.12, 5.12]$ and achieves a minimum value of $f(x^*) = 0$. This function is also known as De Jong's function. This function is Continuous, Differentiable, Separable, Scalable, and Multimodal.

$$f(\mathbf{x}) = \sum_{i-1}^{n} x_i^2 \tag{262}$$

## 4.263 Step [3, 7]

This function computes the sum of the floor values of each element in the vector $x$, effectively summing the greatest integers less than or equal to each element. This function presents a piecewise constant function and is linear in terms of integer steps. The function has a range of $[-100, 100]$, and at $(-0.5, 0.5)$, $f(x^*) = 0$. This function is Non-continuous, Differentiable, Separable, Scalable, and Multimodal.

$$f(\mathbf{x}) = \sum_{i=1}^{D} \lfloor |x_i| \rfloor \tag{263}$$

$$f(x, for2D) = \lfloor x_0 \rfloor + \lfloor x_1 \rfloor$$

## 4.264 Step N.2 [4]

This is a variant of the Step function with a range of $[-100, 100]$ and at $(-0.5, 0.5)$, $f(x^*) = 0$. This variant is Non-continuous, Non-differentiable, Separable, Scalable, and Unimodal.

$$f(\mathbf{x}) = \sum_{i=1}^{D} (\lfloor x_i + 0.5 \rfloor)^2 \tag{264}$$

## 4.265 Step N.3 [4]

This is a variant of the Step function with a range of $[-100, 100]$ and at $(-1, 1)$, $f(x^*) = 0$. This variant is Non-continuous, Non-differentiable, Separable, Scalable, and Unimodal.

$$f(\mathbf{x}) = \sum_{i=1}^{D} \lfloor (x_i)^2 \rfloor \tag{265}$$

## 4.266 Stepint [4]

Similar to the Step function but modified by absolute values. This ensures that all negative values contribute positively, and scaling increases the overall impact of each component. The function has a range of $[-5.12, 5.12]$, and $f(x^*) = 0$. This function is Non-continuous, Non-differentiable, Separable, Scalable, and Unimodal.

$$f(\mathbf{x}) = 25 + \sum_{i=1}^{D} \lfloor x_i \rfloor \tag{266}$$

## 4.267 Stochastic [8]

The output of this function depends on the interval $[x, y]$, and its square ensures a non-negative result, with a spread and average value determined by the range between $x$ and $y$. A common range is at $[-5, 5]$, with $1/n$ and $f(x^*) = 0$.

$$f(\mathbf{x}) = \sum_{i=1}^{n} \epsilon_i \left| x_i - \frac{1}{i} \right| \tag{267}$$

## 4.268 Stretched Cosine Wave

This function combines the effects of position and angle to create a landscape that varies both with the distance from the origin and the angle in a polar coordinate system. The addition of 0.5 and normalization ensure well-defined behavior near the origin. This function has a range of $[-10, 10]$.

$$f(\mathbf{x}) = 0.5 + \frac{(x_0^2 + x_1^2)\cos\left(\frac{x_0}{3\pi}\right)\cos\left(\frac{x_1}{3\pi}\right)}{\sqrt{x_0^2 + x_1^2 + 1.0 \cdot 10^{-8}}} \tag{268}$$

## 4.269 Stretched V Sine Wave [33]

This function's behavior is heavily modulated by both the amplitude, which increases with distance due to the power of 0.25, and the frequency of oscillations, which also increases with distance due to the power transformation inside the sinusoid. This function has a range of $[-10, 10]$ and is Continuous, Differentiable, Non-separable, Scalable, and Unimodal.

$$f(\mathbf{x}) = \sum_{i=1}^{D-1} \left((x_{i+1}^2 + x_i^2)^{0.25} \cdot \sin^2\left(50 \cdot (x_{i+1}^2 + x_i^2)^{0.1}\right) + 0.1\right) \tag{269}$$

## 4.270 Styblinski Tang [8]

This function exhibits multiple local minima and maxima due to the structure of the polynomial. The coefficients significantly shape the landscape, providing steep valleys and hills as characteristic of quartic polynomials. This function has a range between $[-5, 5]$, with a global minimum of the function at $x_i = -2.9035$ for all $i$, with the function value at the global minimum being $-39.16599 \cdot n$ where $n$ is the number of dimensions. This function is Continuous, Differentiable, Non-separable, Non-scalable, and Multimodal.

$$f(\mathbf{x}) = \frac{1}{2} \sum_{i=1}^{n} \left(x_i^4 - 16x_i^2 + 5x_i\right) \tag{270}$$

## 4.271 Sum of Different Powers

This unimodal function represents the sum of absolute power terms. This function is always non-negative, increasing quickly with the magnitude of $y$ compared to $x$ due to the cubic term applied to $y$. This function has a range between $[-5, 5]$ and $f(x^*) = 0$.

$$f(\mathbf{x}) = |x_0|^2 + |x_1|^3 \tag{271}$$

## 4.272 Sum Squares

The Sum Squares function, also referred to as the Axis Parallel Hyper-Ellipsoid function, has no local minimum except the global one. It is continuous, convex, and unimodal with a range of $[-5, 5]$, with 0 and $f(x^*) = 0$.

$$f(\mathbf{x}) = \sum_{i=1}^{2} i{x_i}^2 \tag{272}$$

## 4.273 Tablet

The function features a quadratic term for $x$ with a significantly larger coefficient compared to $y$, leading to a much steeper growth in the $x$-direction. This function has a range of $[-5, 5]$.

$$f(\mathbf{x}) = \exp(6)x_0^2 + x_i^2 \tag{273}$$

### 4.274 Testtube Holder [8, 46]

This function combines sinusoidal functions with an exponential factor that depends on the radial distance from the origin normalized by $\pi$. Thus, this function features oscillations with an amplitude modulated by an exponential decay or growth factor that is centered on the unit circle in the $x$-$y$ plane. This function has a range between $[-1, 4]$, with $(-\pi/2, 0)$ and $f(x^*) = -10.8722$. This function is Continuous, Differentiable, Separable, Scalable, and Multimodal.

$$f(\mathbf{x}) = -4\left(\sin(x_1)\cos(x_2) \cdot e^{|\cos(x_1^2+x_2^2)/200|}\right) \tag{274}$$

### 4.275 Thevenot [83]

This function has a minima at f(0, ..., 0)=-1, for, $m$=5 and $\beta$=15. This function is Continuous, Differentiable, Separable, Scalable, and Multimodal.

$$f(\mathbf{x}) = exp(-\sum_{i=1}^{d}(x_i/\beta)^{2m}) - 2exp(-\prod_{i=1}^{d} x_i^2)\prod_{i=1}^{d} cos^2(x_i) \tag{275}$$

### 4.276 Thurber

In this function, the logarithmic transformation dampens the extreme values and reflects the underlying quadratic behavior. This generates a complex landscape with multiple local minima or maxima depending on the relationship between $x$ and $y$. This function has a range between $[-5, 5]$.

$$f(\mathbf{x}) = \log\left(\left(-x_0^2 - 100000x_0 + x_1 - 1000\right)^2 + \left(-x_0^2 - 100000x_0 + x_1 + 1000\right)^2\right) \tag{276}$$

### 4.277 Trec

The function has a quartic term in $x$ that is modified by a subtractive quadratic term in $x$ and a smaller quadratic term in $y$. This setting creates a double-well potential along the $x$-axis, with two local minima and one local maximum at $x = 0$ and a gentle increase in function value with $y$. This function has a range between $[-5, 5]$.

$$f(\mathbf{x}) = x_0^4 - x_0^2 + 0.1x_1^2 \tag{277}$$

### 4.278 Trecanni [8, ?]

This function features a quartic polynomial in $x$ with distinct coefficients that create a landscape with potential local minima or maxima (due to the presence of cubic and linear terms). The function includes a simple quadratic term in $y$, which influences the function's value based on the square of $y$. Hence, this landscape can be more dynamic in the $x$-direction while maintaining a parabolic shape in the $y$-direction. This function has a range between $[-5, 5]$, with $(-2, 0)$ and $f(x^*) = 0$. This function is Continuous, Differentiable, Separable, Scalable, and Unimodal.

$$f(\mathbf{x}) = x_0^4 - 4x_0^3 + 4x_0 + x_1^2 \tag{278}$$

## 4.279 Trefethen [8],reference 46

This function is expected to exhibit a chaotic behavior with numerous local minima and maxima that are influenced by the specific values and combinations of $x$ and $y$. This function has a range between $[-10, 10]$, $(-0.0244, 0.2106)$, and $f(x^*) = -3.3068$.

$$f(\mathbf{x}) = e^{\sin(50x_0)} - \sin(10x_0 + 10x_1) + \sin(60e^{x_1}) + \sin(70\sin(x_0)) + \sin(\sin(80x_1)) + \frac{x_0^2}{4} + \frac{x_1^2}{4} \tag{279}$$

## 4.280 Trid [8]

This function integrates simple quadratic terms (in terms of $x$ and $y$) together with a negative interaction term $-x \cdot y$. The presence of coupling significantly influences the shapes of the function's landscape. This function is noted for its distinct behavior in higher dimensions as it is characterized by having no local minima other than the global one. The function operates within a range of $[-d^2, d^2]$. The function's global minimum with $f(x^*) = -50$ for $d = 6$ and $f(x^*) = -200$ for $d = 10$.

$$f(\mathbf{x}) = \sum_{i=2}^{d} (x_0 - 1)^2 - \sum_{i=2}^{d} x_0 x_1 \tag{280}$$

## 4.281 Trid N.6

This is a variant of the Trid function with a range of $[-6, 6]$ with a minima at -50. This variant is Continuous, Differentiable, Non-separable, Non-scalable, and Multimodal.

$$f(\mathbf{x}) = \sum_{i=2}^{d} (x_0 - 1)^2 - \sum_{i=2}^{d} x_0 x_1 \tag{281}$$

## 4.282 Trid N.10

This is a variant of the Trid function with a range of $[-100, 100]$ with a minima at -200. This variant is Continuous, Differentiable, Non-separable, Non-scalable, and Multimodal.

$$f(\mathbf{x}) = \sum_{i=2}^{d} (x_0 - 1)^2 - \sum_{i=2}^{d} x_0 x_1 \tag{282}$$

## 4.283 Trigonometric [8]

The function combines a simple trigonometric reduction with a more complex polynomial increase. Thus, this function is sensitive to the values and order of elements in $x$. The function's range is $[0, \pi]$, with $f(x^*) = 0$. This function is Continuous, Differentiable, Non-separable, Scalable, and Multimodal.

$$f(\mathbf{x}) = \sum_{i=1}^{D} \left( D - \sum_{j=1}^{D} \cos x_j \right) + i(1 - \cos(x_i) - \sin(x_i))^2 \tag{283}$$

## 4.284 Trigonometric N.2 [8]

This is a variant of the Trigonometric function with a range of $[-500, 500]$ and at 0.9, $f(x^*) = 1$. This variant exhibits highly oscillatory behavior due to the inclusion of the sine terms that are compounded by the scaling factor that increases with $i$, leading to significant contributions from later elements. This variant is Continuous, Differentiable, Non-separable, Scalable, and Multimodal.

$$f(\mathbf{x}) = 1 + \sum_{i=1}^{D} 8\sin^2\left(7(x_i - 0.9)^2\right) + 6\sin^2\left(14(x_i - 0.9)^2\right) + (x_i - 0.9)^2 \tag{284}$$

## 4.285 Trimmed Sphere

This function calculates the radial distance squared and clips it at 0.5. The function is effectively a plateaued disk with a continuous but non-differentiable boundary at the clipping radius. This function has a range between $[-10, 10]$.

$$f(\mathbf{x}) = \begin{cases} x_0^2 + x_1^2 & \text{for } x_0^2 + x_1^2 < 0.5 \\ 0.5 & \text{otherwise} \end{cases} \tag{285}$$

## 4.286 Tripod [7, ?]

The behavior of this function changes as a result of the absolute values and conditions that change the offsets and scaling of $x$ and $y$. The function's range is $[-100, 100]$, with $(-50, 0)$ and $f(x^*) = 0$. This function is Non-separable and Multimodal.

$$f(\mathbf{x}) = p(y) \cdot (1 + p(x)) + |x + 50 \cdot p(y) \cdot (1 - 2 \cdot p(x))| + |y + 50 \cdot (1 - 2 \cdot p(y))| \tag{286}$$

## 4.287 Two Axis

This function multiplies the squares of $x$ and $y$, producing a surface that increases quadratically from the origin in all directions. The function's range is $[-5, 5]$, and $f(x^*) = 0$.

$$f(\mathbf{x}) = x_0^2 x_1^2 \tag{287}$$

## 4.288 Urfun N.2

This function incorporates a series of sinusoidal functions in both $x$ and $y$, each term modulated by increasing phase shifts and frequencies and scaled by a decreasing function of $i$. As such, this function exhibits complex oscillatory patterns due to the sinusoidal terms. The function's range is $[-5, 5]$.

$$f(\mathbf{x}) = \sum_{i=1}^{5} \frac{\sin(2ix_0 + i)\sin(2ix_1 + i)}{i^3(i+1)} \tag{288}$$

## 4.289 Ursem [8, ?]

This function combines a shifted sine wave in $x$, a cosine wave in $y$, and a linear term in $x$. This function also varies periodically in both $x$ and $y$ with a consistent decrease in value with increasing $x$ due to the linear term. The function has one global and one local minimum [42] at $(1.6971, 0)$ and $f(x^*) = -4.816$. The function has a range across $[-2.3, 3]$ and is Non-scalable.

$$f(\mathbf{x}) = -\sin(2x_0 - 0.5\pi) - 3\cos(x_1) - 0.5x_0 \tag{289}$$

## 4.290 Ursem N.3 [8, ?]

This is a variant of the Ursem function with a range of $x \in [-2, 2]$ and $y \in [-1.5, 1.5]$ and at 0, $f(x^*) = -2.5$. This function is Non-scalable.

$$f(\mathbf{x}) = -\sin(2.2\pi x_1 + 0.5\pi)\left(\frac{2-|x_1|}{2}\right)\left(\frac{3-|x_1|}{2}\right) - \sin(0.5\pi x_2^2 + 0.5\pi)\left(\frac{2-|x_2|}{2}\right)\left(\frac{3-|x_2|}{2}\right) \tag{290}$$

## 4.291 Ursem N.4 [8, ?]

This is a variant of the Ursem function with a range of $x \in [-2, 2]$ and at 0, $f(x^*) = -1.5$. This variant has one global minimum positioned at the middle and four local minima at the corners of the search space and is Non-scalable.

$$f(\mathbf{x}) = -3\sin(0.5\pi x_1 + 0.5\pi)\left(\frac{2-\sqrt{x_1^2+x_2^2}}{4}\right) \tag{291}$$

### 4.292 Ursem Waves [8, ?]

This function has a range of $x \in [-0.9, 1.2]$ and $y \in [-1.2, 1.2]$
and at $(-0.6056, -1.1775)$, $f(x^*) = -8.553$.

$$f(\mathbf{x}) = -0.9x_1^2 + (x_2^2 - 4.5x_2)x_1x_2 + 4.7\cos(3x_1 - x_2^2(2 + x_1))\sin(2.5\pi x_1) \tag{292}$$

### 4.293 Ursem Wavesfun N.2 [42]

This is a variant of the Ursem Wavesfun function with a range of $x \in [-0.9, 1.2]$ and $y \in [-1.2, 1.2]$ and at $(-0.6056, -1.1775)$, $f(x^*) = -7.3069$. The variant involves quadratic and higher-order polynomial terms in $x$ and $y$, coupled with trigonometric terms that are complexly related to both $x$ and $y$. As the $\sin(2.5\pi) = 0$, the last term effectively cancels out. This simplifies the function to primarily polynomial terms. This variant has one global minimum positioned at the middle and four local minima at the corners of the search space and is Non-scalable.

$$f(\mathbf{x}) = 2.5\sin\left(\frac{3\pi x_0}{2}\right)x_0 + 1.5\sin\left(\frac{3\pi x_1}{2}\right)x_1 \tag{293}$$

### 4.294 Venter Sobiezcczanski Sobieski [8, ?]

This function includes a straightforward quadratic term for both $x$ and $y$ and introduces significant modulations with cosine terms that are both fast and slow oscillating due to the frequency adjustments. The function's landscape has multiple local minima and maxima superimposed on a generally increasing quadratic function. The function's range is $[-50, 10]$, and $f(x^*) = 1000$. This function is Continuous, Differentiable, Separable, and Non-scalable.

$$f(\mathbf{x}) = x_1^2 - 100\cos(x_1^2) - 100\cos\left(\frac{x_1^2}{30}\right) + x_2^2 - 100\cos(x_2^2) - 100\cos\left(\frac{x_2^2}{30}\right) \tag{294}$$

### 4.295 Vincent [8]

This function applies a sinusoidal term to the logarithm of $x$ and $y$. Since this application is sensitive to the sign and magnitude of $x$ and $y$, this function exhibits complex behavior arising from the logarithm inside a sine function. The function's range is $[0.25, 10]$.

$$f(\mathbf{x}) = -\sum_{i=1}^{n} \sin(10\log x_i) \tag{295}$$

### 4.296 Watson [8]

This function involves a double summation where each element $x[i]$ contributes to polynomial terms of increasing degree scaled by both the term order $j$ and a factor derived from $j$. The function's behavior is highly influenced by the values and the size of $x$. The function's range is $[-5, 5]$, with $(-0.0158, 1.012, -0.2329, 1.60, -1.513, 0.9928)$ and $f(x^*) = 0.0228$. This function is Continuous, Differentiable, Non-separable, Scalable, and Unimodal.

$$f(\mathbf{x}) = \sum_{i=0}^{29} \left( \sum_{j=0}^{4} ((j-1)d_{i,j}x_{j+1}) - \left( \sum_{j=0}^{5} d_j x_{j+1} \right)^2 - 1 \right)^2 + x_1^2 \tag{296}$$

### 4.297 Wavy [8]

This function multiplies the sine of $x$ with the cosine of $y$, resulting in a function that exhibits standard trigonometric product properties. Hence, it is periodic in both $x$ and $y$ with a pattern that repeats every $2\pi$. The function's output oscillates between $-1$ and 1. The function has a range between $[-\pi, \pi]$ with $(0.2, 1)$ and $f(x^*) = 0$. This function is Continuous, Differentiable, Separable, Scalable, and Multimodal.

$$f(\mathbf{x}) = 1 - \frac{1}{n} \sum_{i=1}^{n} \cos(kx_i) e^{-\frac{x_i^2}{2j^2}} \tag{297}$$

### 4.298 Wayburn Seader [8, ?]

This function combines a sixth-degree polynomial in $x$ and a fourth-degree polynomial in $y$ with a quadratic expression. The first term creates a complex landscape with potentially multiple local minima due to the high power, and the second term provides fine-tuning adjustment. The function has a range between $[-500, 500]$ with $(1, 2)$, $(1.5968, 0.8063)$, and $f(x^*) = 0$.

$$f(\mathbf{x}) = (2x_0 + x_1 - 4)^2 + \left(x_0^6 + x_1^4 - 17\right)^2 \tag{298}$$

### 4.299 Wayburn Seader N.2 [8, ?]

This is a variant of the Wayburn Seader function with a range of $[-500, 500]$ and at $(0.20013, 1)$, $(0.42486, 1)$, $f(x^*) = 0$. This variant features quadratic terms that are shifted by constants that affect the curvature and positioning of the minima. Further, introducing the $(x - y)^2$ term introduces a continued interaction effect that ties $x$ and $y$ to further complicate the variant's landscape. This variant is Continuous, Differentiable, Non-separable, Scalable, and Unimodal.

$$f(\mathbf{x}) = \left[1.613 - 4(x_1 - 0.3125)^2 - 4(x_2 - 1.625)^2\right]^2 + (x_2 - 1)^2 \tag{299}$$

### 4.300 Wayburn Seader N.3 [8, ?]

This is a variant of the Wayburn Seader function with a range of $[-0.5, 0.5]$ and at $(0, .., 0, f(x^*) = 0$. This variant is Continuous, Differentiable, Non-separable, Scalable, and Unimodal.

$$f(\mathbf{x}) = \frac{2}{3}x_1^3 - 8x_2^2 + 33x_1x_2 - x_1x_2 + 5 + \left[(x_1 - 4)^2 + (x_2 - 5)^2 - 4\right]^2 \quad (300)$$

### 4.301 Weierstrass [8]

This function is multimodal, and it is continuous everywhere but only differentiable on a set of points. The suggested search area is the hypercube $[-0.5, 0.5]^D$. Note that if a larger search area is considered, then there might be multiple global optima as the function is periodic within $[-10, 10]$. This function is Continuous, Differentiable, Separable, Scalable, and Multimodal.

$$f(\mathbf{x}) = \sum_{i=1}^{n} \sum_{k=0}^{k_{\max}} \left[a^k \cos(2\pi b^k (x_i + 0.5))\right] - n \sum_{k=0}^{k_{\max}} \left[a^k \cos(\pi b^k j)\right] \quad (301)$$

### 4.302 Whitley [8]

This function combines a complex rational expression with a cosine function. The rational part can grow very large, mainly when the values of $x$ and $y$ do not satisfy the condition $x^2 \approx y$. On the other hand, the cosine term introduces periodic fluctuations, which can significantly impact the function's smoothness and predictability. This function has a range of $[-10.24, 10.24]$ and at 0, $f(x^*) = 0$. This function is Continuous, Differentiable, Non-separable, Scalable, and Multimodal.

$$f(\mathbf{x}) = \frac{\left((1 - x_j)^2 + 100\left(x_i^2 - x_j\right)^2\right)^2}{4000} - \cos\left(100(x_i^2 - x_j)^2 + (1 - x_j)^2 + 1\right) \quad (302)$$

### 4.303 Wolfe [8, 80]

This function applies a power of 0.75 to a quadratic form involving both $x$ and $y$. The subtraction of $x \cdot y$ introduces an interaction term that affects the geometry of the function's graph. The function has a softened landscape compared to typical quadratic or higher-degree polynomials (with gentle slopes). This function ranges from $[-10, 10]$, and at 0, $f(x^*) = 0$. This function is Continuous, Differentiable, Separable, Scalable, and Multimodal.

$$f(\mathbf{x}) = 4/3\left(x_0^2 - x_0x_1 + x_1^2\right)^{0.75} + x_3 \quad (303)$$

### 4.304 Xin She Yang N.1 [8]

This function is nonconvex and nonseparable. The function is not smooth, and its derivatives are not well-defined at the optimum. The suggested search area is the hypercube $[-2\pi, 2\pi]^D$.

$$f(\mathbf{x}) = \sum_{i=1}^{D} \epsilon_i |x_i|^i \tag{304}$$

### 4.305 Xin She Yang N.2 [8]

This is a variant of the Xin She Yang function with a range of $[-2\pi, 2\pi]$ and at 0, $f(x^*) = 0$. The variant combines a linear sum of absolute values with an exponential term that depends negatively on the sum of the sine of squares of $x$. This exponential term smoothness and dampness are the contributions of the linear term as the values of $x$ lead to higher values of $\sin(x^2)$. This variant is Separable.

$$f(\mathbf{x}) = \left(\sum_{i=1}^{D} |x_i|\right) \exp\left(-\sum_{i=1}^{D} \sin(x_i^2)\right) \tag{305}$$

### 4.306 Xin She Yang N.3 [8, 81]

This is a variant of the Xin She Yang function with a range of $[-20, 20]$ and at 0, $f(x^*) = -1$. The variant combines a linear sum with an exponential term that depends negatively on the sum of the cosine of squares of $x$. This variant is Non-convex and Non-separable, with multiple local minima and a unique global minimum.

$$f(\mathbf{x}) = e^{-\sum_{i=1}^{D}(x_i^2/\beta)^{2m}} - 2e^{-\sum_{i=1}^{D} x_i^2} \prod_{i=1}^{D} \cos^2(x_i) \tag{306}$$

### 4.307 Xin She Yang N.4 [8, 81]

This is a variant of the Xin She Yang function with a range of $[-10, 10]$. The variant combines sinusoidal functions, both directly and within an exponential, with a double-layer of exponential functions modifying the impact. This variant also contains a subtractive interaction between a straightforward sinusoidal sum and a complex exponential term that combines Gaussian decay and modified sinusoidal terms. This variant is Non-scalable.

$$f(\mathbf{x}) = (\sum_{i=1}^{D} \sin^2(x_i) - e^{-\sum_{i=1}^{D} x_i^2}) - e^{-\sum_{i=1}^{D} \sin^2(\sqrt{|x_i|})} \tag{307}$$

### 4.308 Xin She Yang N.7 [82]

This is a variant of the Xin She Yang function with a range of $[-5, 5]$ and at $1/i$, $f(x^*) = -1$. The variant combines two modified Rastrigin functions, one for each variable. This variant is non-negative due to the square and positive constant terms.

$$f(\mathbf{x}) = \sum_{i=1}^{n} \epsilon_i \left| x_i - \frac{1}{i} \right| \tag{308}$$

### 4.309 Yao Liu N.4 [8]

This is a variant of the Yao Liu function with a range of $[-5, 5]$ and at $1/i$, $f(x^*) = -1$. The variant combines two modified Rastrigin functions, one for each variable. This variant is non-negative due to the square and positive constant terms, and it combines a radial distance measurement with sinusoidal functions applied to the squares of $x$ and $y$.

$$f(\mathbf{x}) = \max_i |x_i| \quad \text{for } i = 1, 2, \ldots, n \tag{309}$$

### 4.310 Yao Liu N.9 [8]

This is a variant of the Yao Liu function with a range of $[-5.12, 5.12]$ and at 0, $f(x^*) = 0$. The variant combines the absolute value of the difference of squares with a linear sum of the variables. This variant will have valleys and ridges corresponding to where the square terms equal each other and increase linearly away from these points. The variant reaches its minimum when $x = y$ or $x = -y$ due to the absolute term becoming zero.

$$f(\mathbf{x}) = \sum_{i=1}^{n} \left(x_i^2 - 10\cos(2\pi x_i) + 10\right) \tag{310}$$

### 4.311 Zakharov [7, 38]

This function is known for its distinctive landscape that, despite potential complexity, contains only one global minimum without any local minimum. This function begins with the sum of the squares of all elements, to which it adds both squared and quartic terms derived from a weighted linear sum of the elements. Each element in the linear sum is scaled by its index, and this introduces a layer of complexity that depends on the values of $x$ and positions within the vector. The function operates within the domain $[-5, 10]$. This function is Continuous, Differentiable, Non-separable, Scalable, and Multimodal.

$$f(\mathbf{x}) = \sum_{i=1}^{n} x_i^2 + (\sum_{i=1}^{n} 0.5ix_i)^2 + (\sum_{i=1}^{n} 0.5ix_i)^{4"} \tag{311}$$

### 4.312 Zero Sum [8]

This function simplifies to $2x^2 + 2y^2$, a straightforward quadratic function with no cross-term, representing the sum of the individual squares of $x$ and $y$ scaled by 2. This function grows quadratically from the origin and is symmetrical. This function has a range of $[-10, 10]$, with 0 at $f(x^*) = 0$.

$$f(\mathbf{x}) = \begin{cases} 0 & \text{if } \sum_{i=1}^{n} x_i = 0 \\ 1 + \left(10000 \sum_{i=1}^{n} |x_i|\right)^{0.5} & \text{otherwise} \end{cases} \tag{312}$$

### 4.313 Zettl [8, 46]

This function features a squared term with a circular shift in $x$ and an accompanying linear term in $x$ that introduces asymmetry. The squared term primarily shapes the landscape, creating a bowl-like surface but with a distinctive offset due to the shift. This function operates within a range of $[-1, 5]$ and reaches a minimum at $(-0.0299, 0)$ with $f(x^*) = -0.003791$. This function is Continuous, Differentiable, Non-separable, Non-scalable, and Multimodal.

$$f(\mathbf{x}) = \left(x_0^2 - 2x_0 + x_1^2\right)^2 + 0.25x_0 \tag{313}$$

### 4.314 Zettl Variant

This is a variant of the Zettl function with a range of $[-5, 5]$ and at 0, $f(x^*) = 0$. This variant combines higher-order polynomial terms in $x$ with a simple cosine term in $y$. The quartic term creates complexity in the $x$-dimension (i.e., having multiple local minima and maxima), while $y$ contributes a simple parabolic growth.

$$f(\mathbf{x}) = \left(x_0^2 - 2x_0 + x_1^2\right)^2 - \cos(x_1) + 0.5x_0 \tag{314}$$

### 4.315 Zoom

This function incorporates shifts in both $x$ and $y$ for the quadratic terms, adding a sinusoidal modulation dependent on the product of $x$ and $y$. Global behavior is primarily influenced by quadratic terms, with local variations due to sinusoidal terms. This function has a range of $[-5, 5]$.

$$f(\mathbf{x}) = (x - 3.1)^2 + (y - 3.3)^2 + \sin(x \cdot y) \tag{315}$$

# 5 Top 25 Most Commonly Used Benchmark Functions

This is collected from our observations of which functions were seen to be commonly adopted in other reviews and work. Here are these functions listed alphabetically:

1. Ackley
2. Alpine
3. Beale's
4. Booth's
5. Cross-in-Tray
6. Dixon Price
7. Drop-Wave
8. Easom
9. Expanded
10. Schaffer's F6
11. Goldstein-Price
12. Griewank
13. Himmelblau's
14. Holder Table
15. Levy N.13
16. Matyas
17. Michalewicz
18. Rastrigin
19. Rosenbrock
20. Salomon's
21. Schaffer N.2
22. Schwefel
23. Sphere
24. Styblinski-Tang
25. Three-hump Camel

# 6 New Functions and Future Directions

Despite the extensive use of benchmark functions, certain areas remain underrepresented. For instance, tests and benchmarks that accurately simulate real-time decision-making environments, such as dynamic optimization problems where the landscape changes over time, are relatively scarce.

In order to bridge this gap, we present two new functions. These are called,

### The Dynamic Deceptive Basin function

This function is dynamic, meaning its landscape changes over time as it incorporates a stochastic element through a random walk in the parameter space. This dynamic nature requires algorithms to adapt continuously to find and maintain optimal solutions, which is a valuable trait for a benchmark function. The addition of noise further complicates the optimization process, testing the algorithm's ability to differentiate between true gradient changes and random fluctuations.

$$f(\mathbf{x}) = \sin(x_0 + \theta_0) \cdot \cos(x_1 + \theta_1) \cdot e^{-\|\mathbf{x}-\theta\|^2} + \text{noise}$$

Where:

- $\mathbf{x} = [x_0, x_1, \ldots, x_{n-1}]$ represents the input vector.
- $\theta = [\theta_0, \theta_1, \ldots, \theta_{n-1}]$ represents the dynamically updating state parameters.
- Noise is a normally distributed random variable, noise $\sim N(0, 0.1)$.

### The Dynamic Deceptive Basin function

This function is considerably more complex than its counterpart. It integrates polynomial, trigonometric, interaction terms, and exponential components, each adding to the function's multimodality and non-linearity. The historical influence on the function's behavior adds a layer of complexity, simulating real-world scenarios where past states affect current conditions.

$$f(\mathbf{x}) = \sum_{i=0}^{n-1} \theta_i x_i^3 + \sum_{i=0}^{n-1} \sin(x_i + \theta_i) + \sum_{i=0}^{n-1} \frac{x_i(\theta_i)^{i+n}}{0.1|x_2\theta_n|} + \sum_{i=0}^{n-1} \theta_i^2 + \text{noise}$$

Where:

- $\mathbf{x} = [x_0, x_1, \ldots, x_{n-1}]$ represents the input vector.
- $\theta = [\theta_0, \theta_1, \ldots, \theta_{n-1}]$ represents the dynamically updating state parameters.
- Noise has an amplitude dependent on the norm of $x - \sin(\text{recent effect})$ and is normally distributed, noise $\sim N(0, \text{noise amplitude})$.

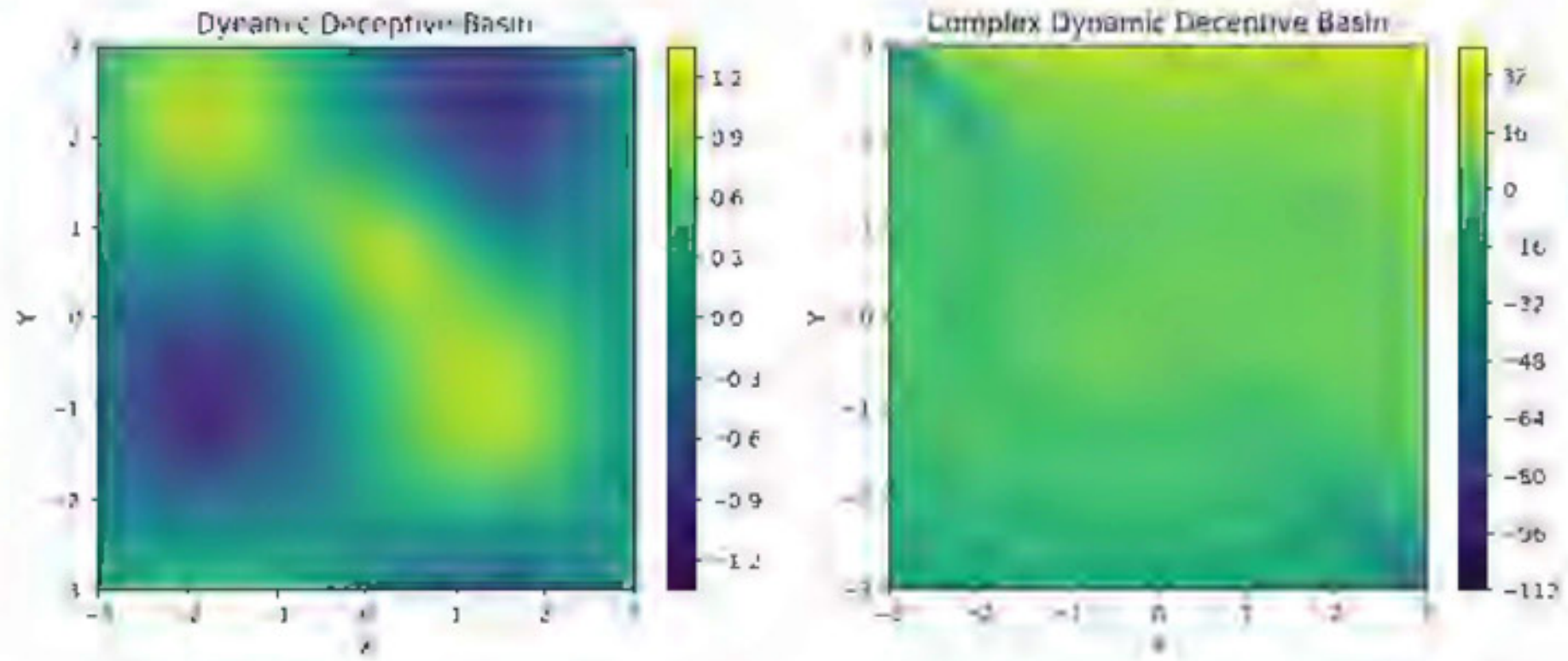

Figure 2: Illustration of the newly proposed functions

As one can see, both of these functions are non-stationary and contain multiple types of mathematical operations aimed to create numerous local minima and maxima (i.e., intricate interdependencies). They also contain noise caused by random perturbations. The functions also incorporate random noise and dynamic parameters (state_params) that change with each function evaluation. This means that the landscape of the function is constantly shifting, making it harder for optimization algorithms to converge to a stable solution. It is worth noting that the Complex Dynamic Deceptive Basin function incorporates a history of previous solutions, effectively introducing memory effects into the optimization process.

There is also a lack of functions that incorporate constraints naturally occurring in real-world scenarios, such as non-negativity, budget constraints, or integer variables. These gaps highlight the need for developing new benchmarks that more closely mirror the complexities and constraints of practical applications. Emerging trends include the optimization of systems with inherent uncertainty, like renewable energy systems, and the optimization of complex networks in telecommunications and social media analytics.

Further, integrating real-world problem scenarios into benchmark functions would help in testing how algorithms perform under practical conditions. For example, in engineering or sciences, benchmarks that simulate energy consumption patterns in smart grids or traffic flow in urban networks could provide insights into how optimization algorithms can be applied to manage and optimize real-world systems efficiently. To address these emerging trends, potential new benchmark functions arise. We invite interested researchers and future readers to think outside the box and propose standardized practices to integrate real-world scenarios into benchmark functions. This direction will enhance the reliability of optimization algorithms and broaden their applicability across different sectors.

## 7 Conclusions

This survey has cataloged and categorized over 300 benchmark functions to reveal a diverse catalog for testing and developing optimization algorithms. We highlight how these functions are applied across various domains and provide details about their mathematical descriptions, practical implications, and limitations. We hope this survey will serve as a foundational reference to aid researchers and practitioners in selecting appropriate benchmark functions to develop more robust optimization tools. We note that while this review showcases over 300 functions, it is quite possible that other functions (or variants) also exist. We invite interested readers to build upon this effort to continue to provide a venue for thorough documentation of benchmarking functions.

## 8 Data Availability

Some or all data, models, or codes that support the findings of this study are available from the corresponding author upon reasonable request.

## 9 Conflict of Interest

The authors declare no conflict of interest.

# Appendix: Visual Representation of Benchmark and Test Functions Landscapes

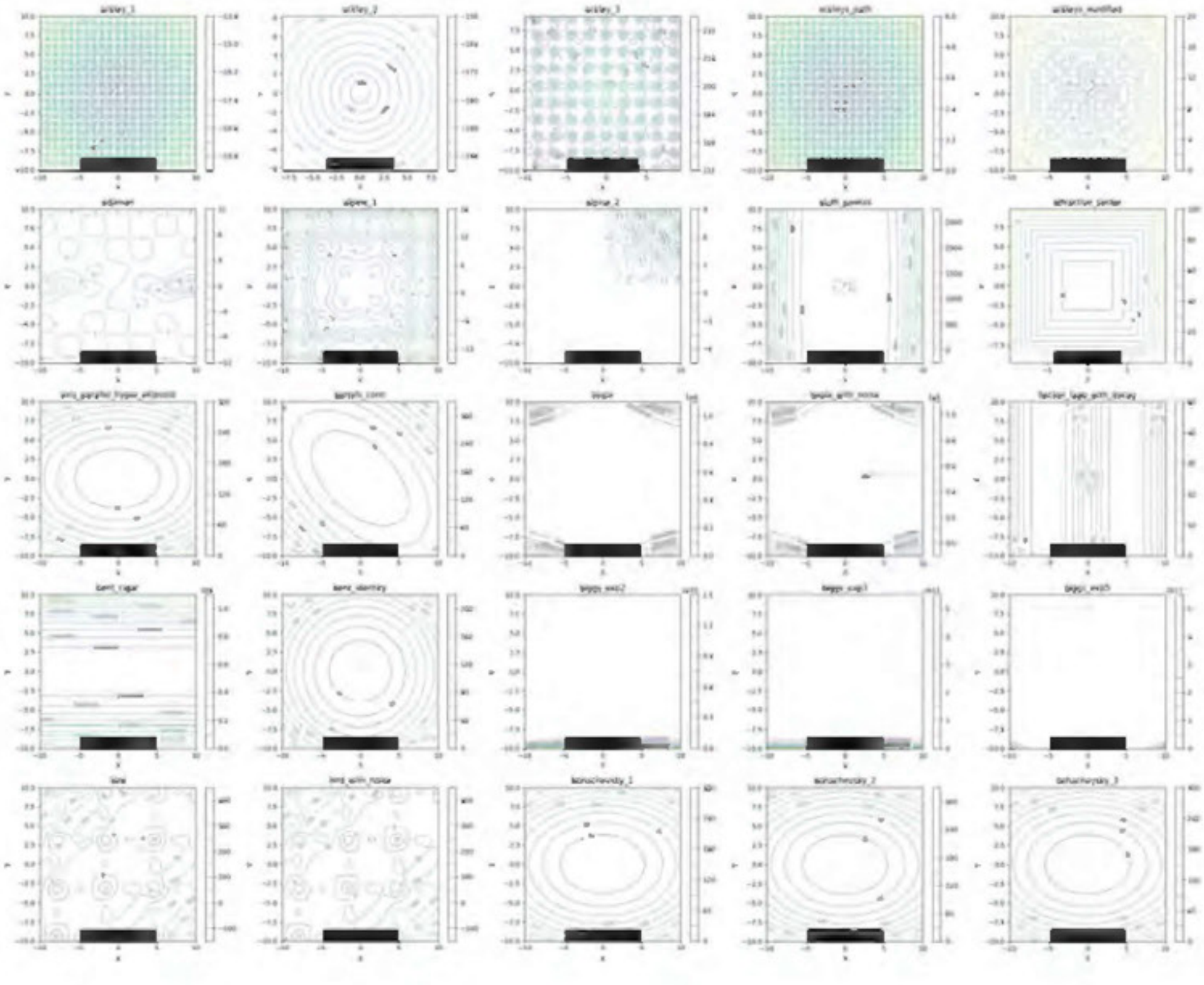

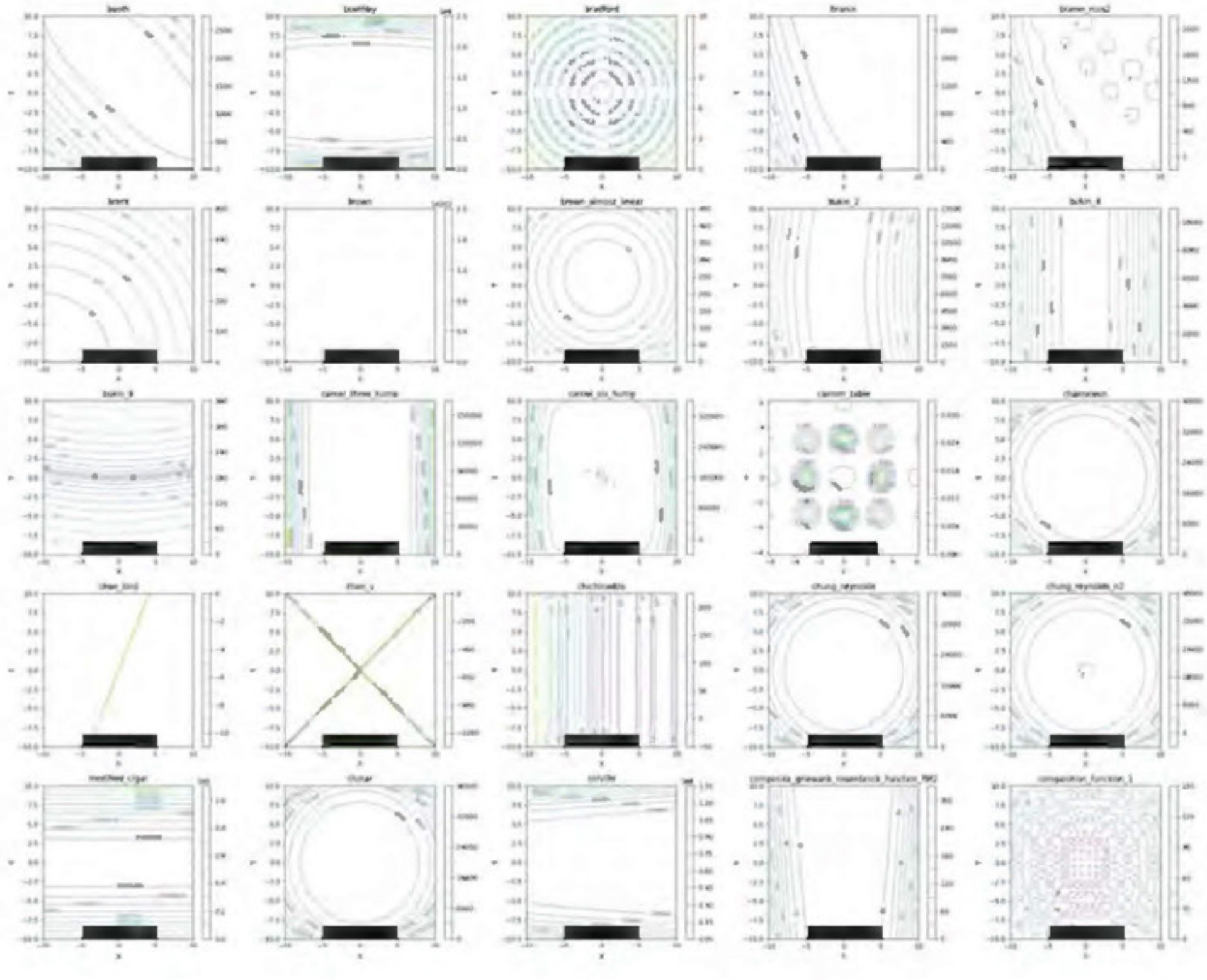

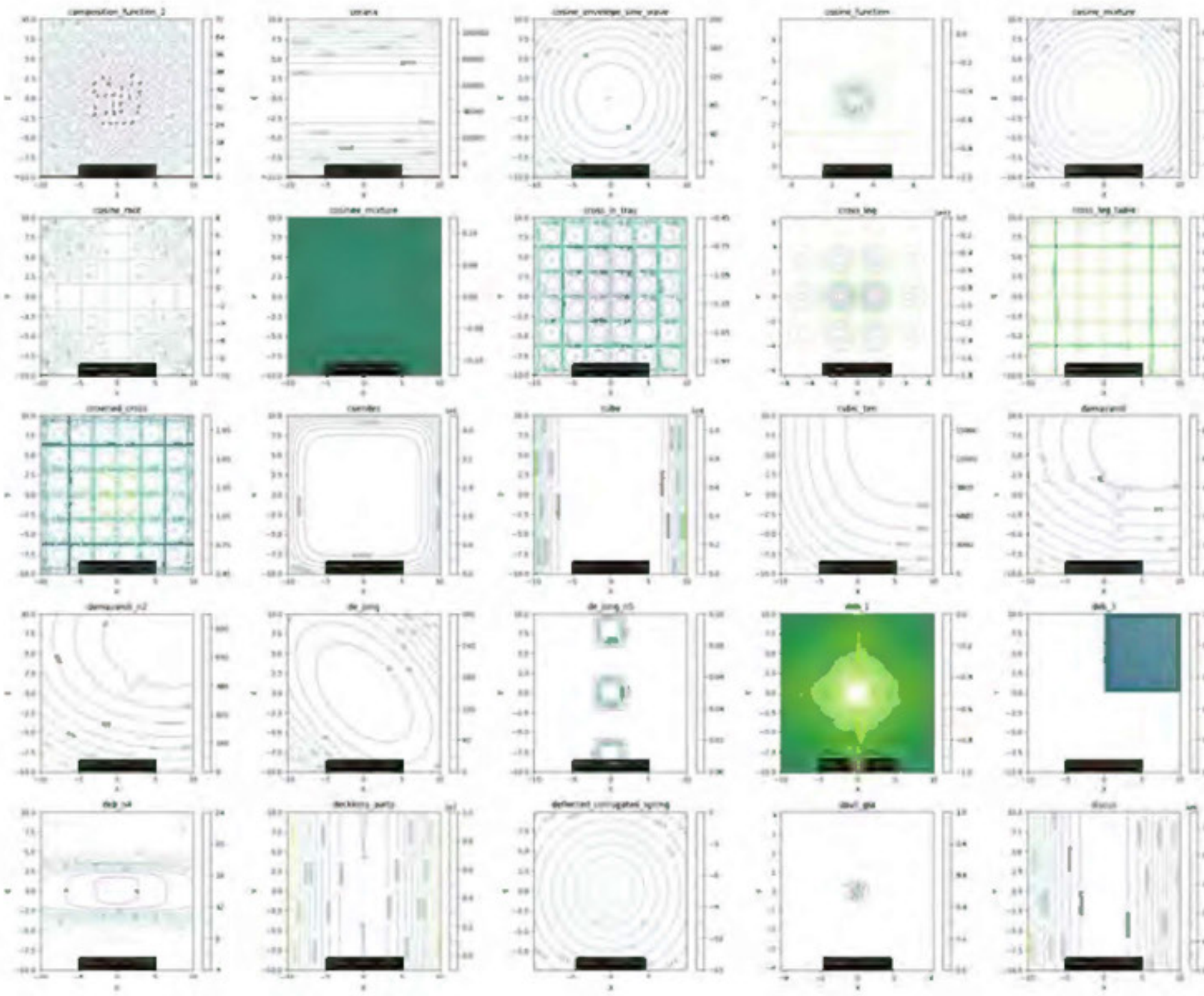

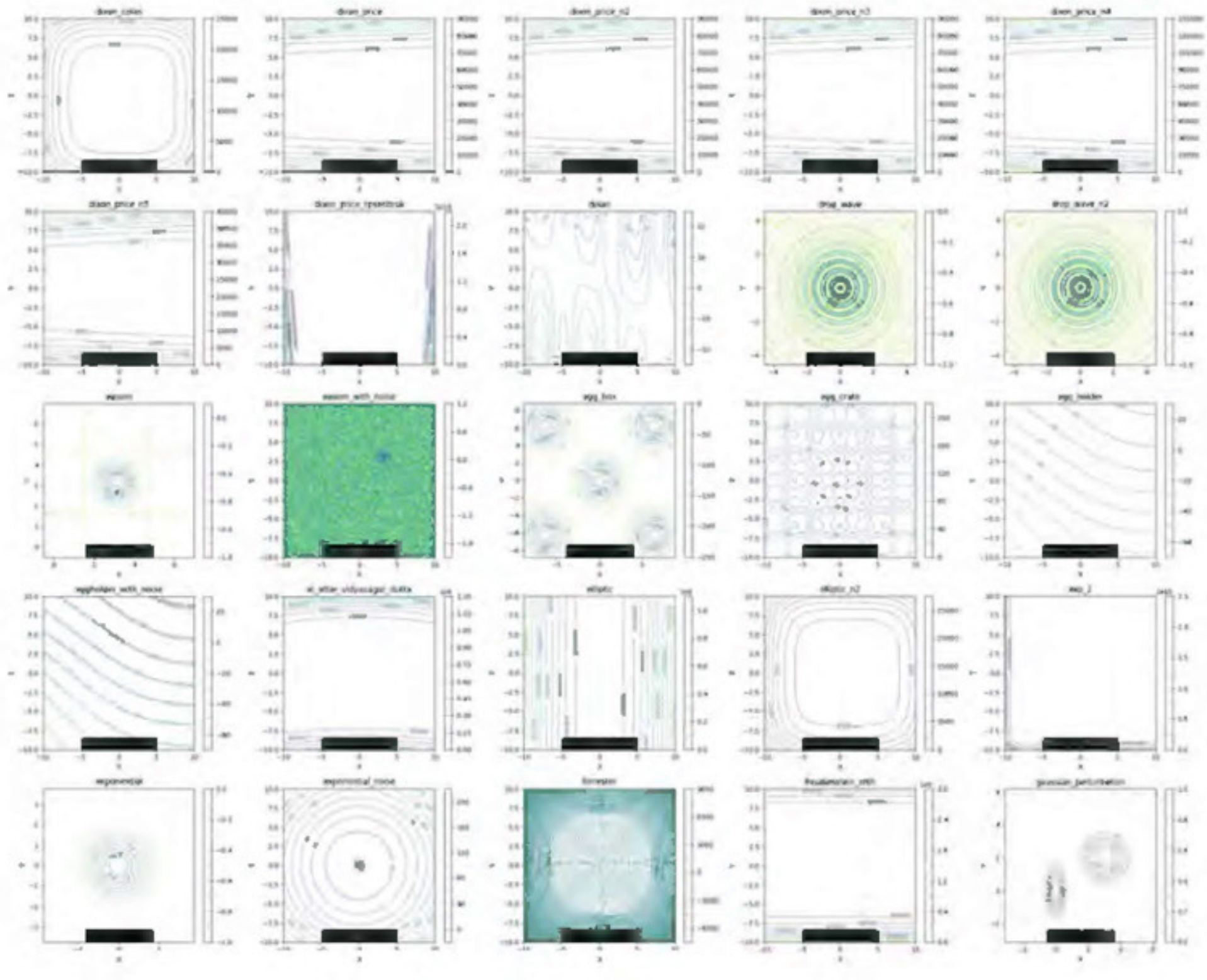

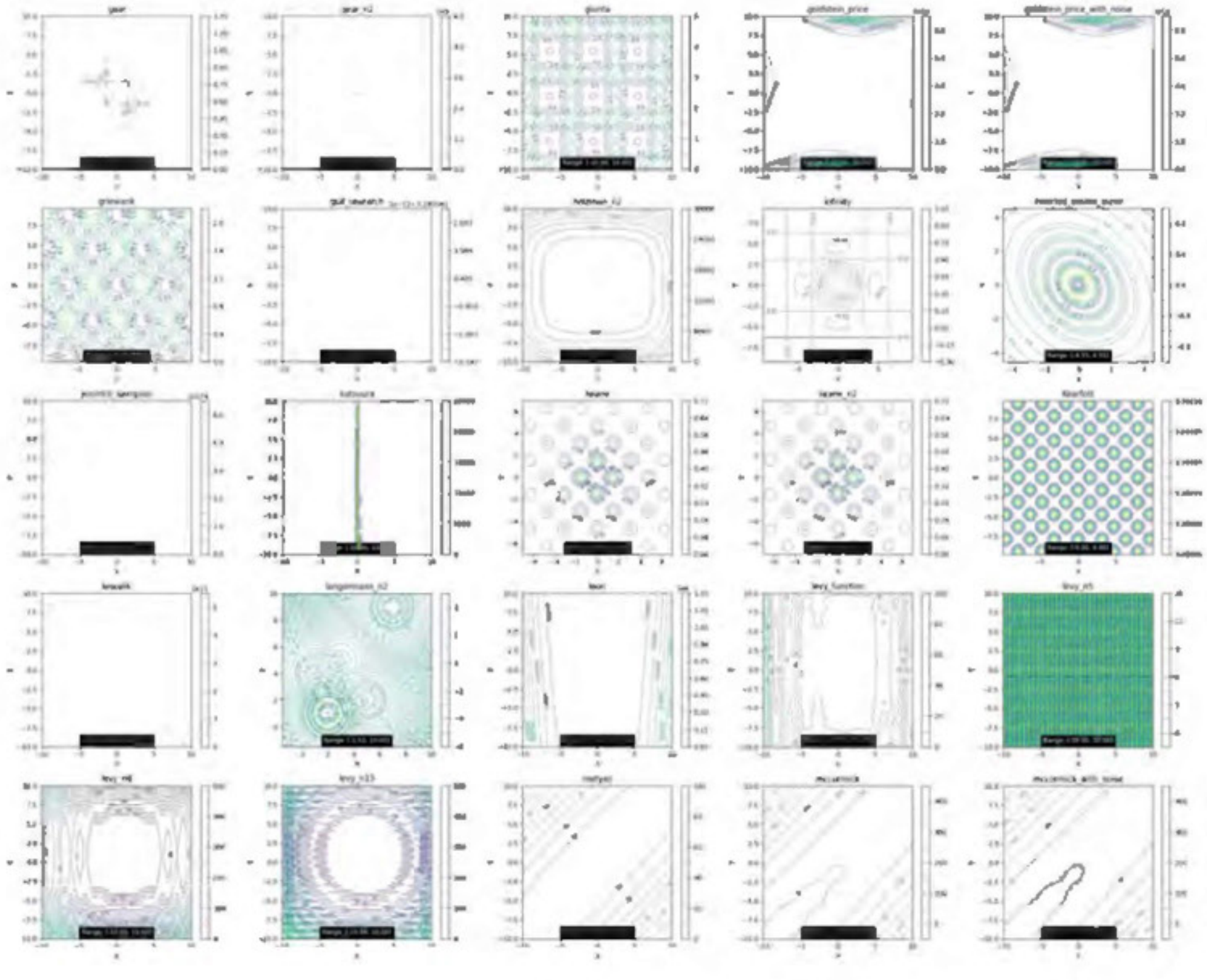

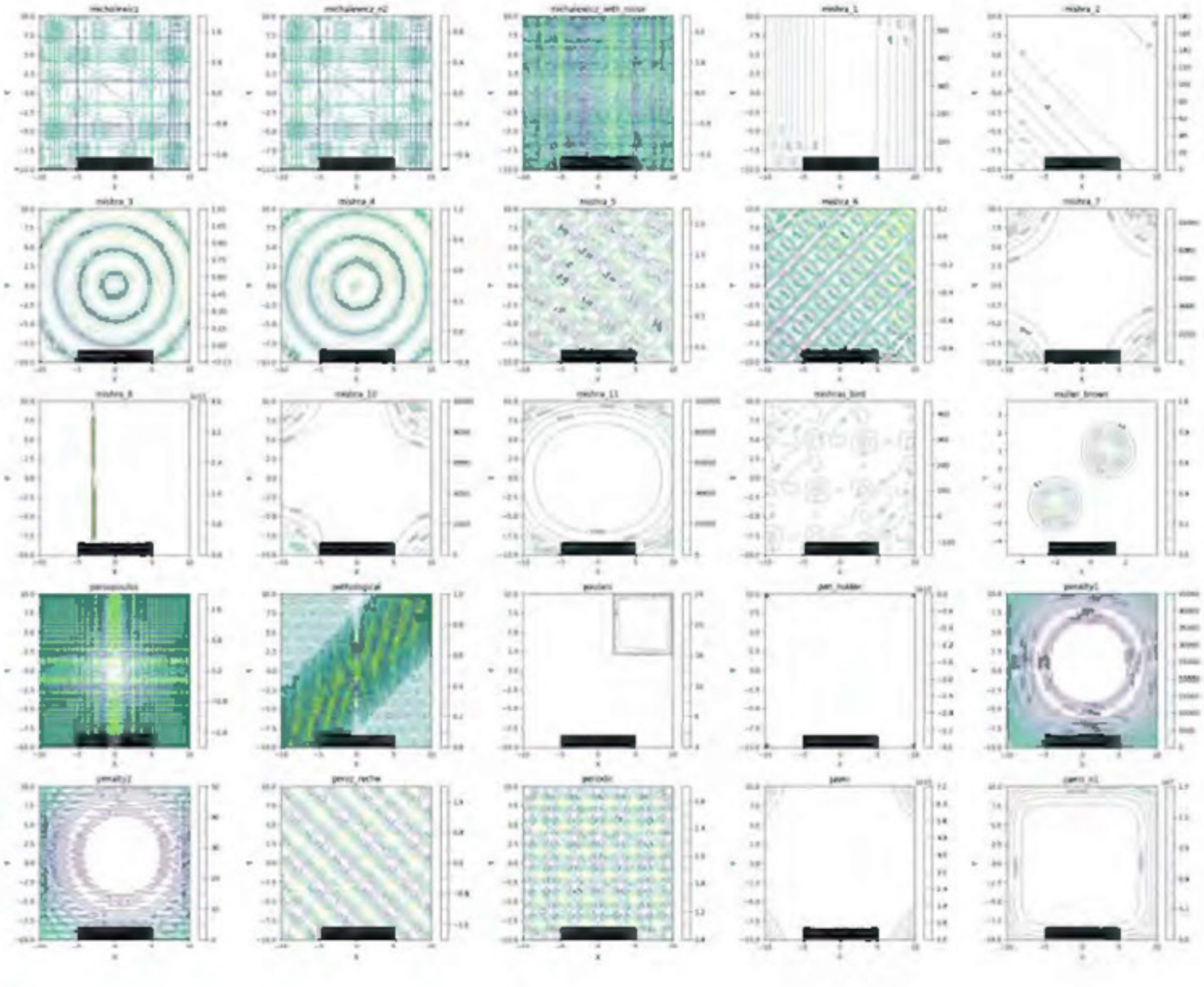

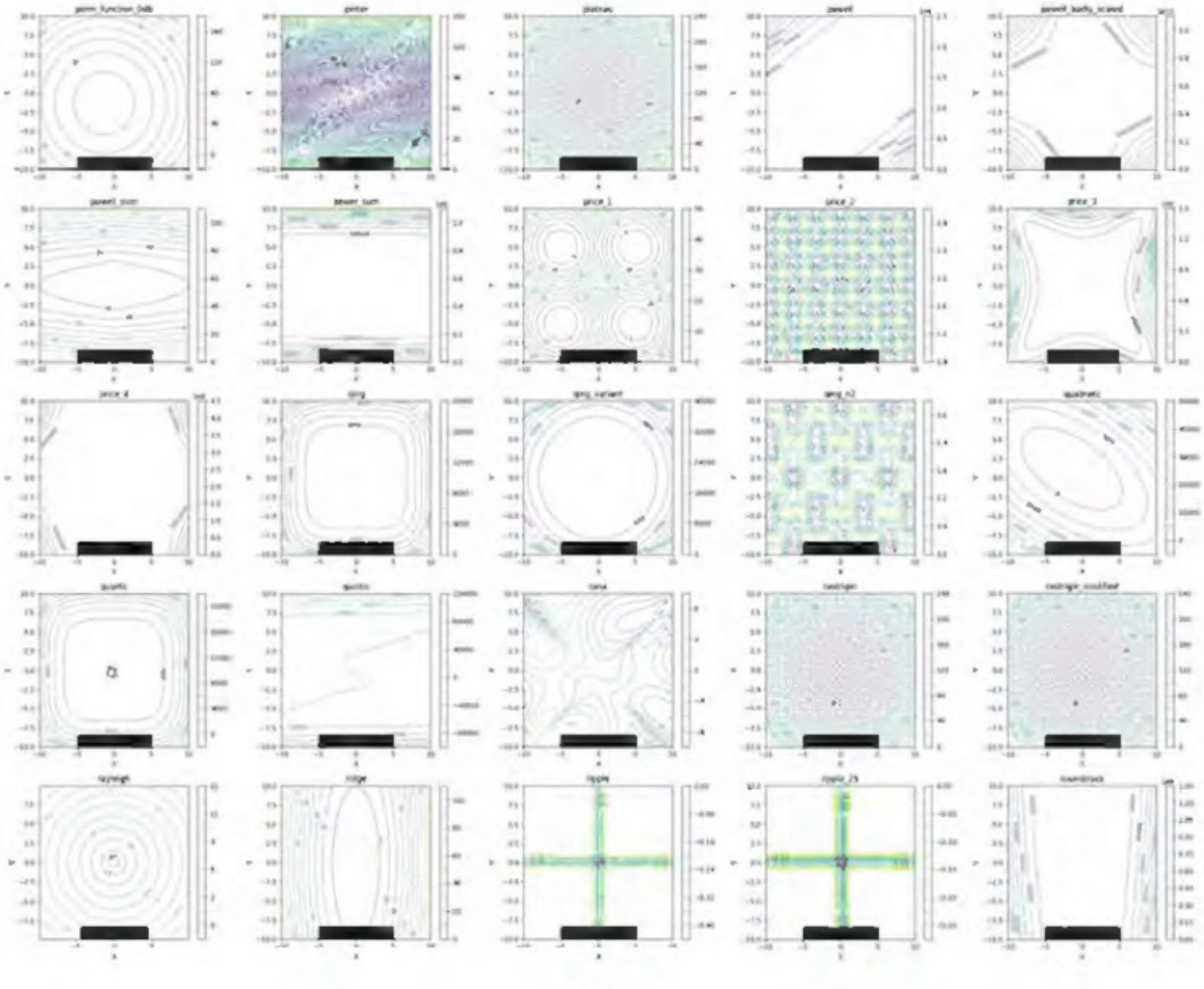

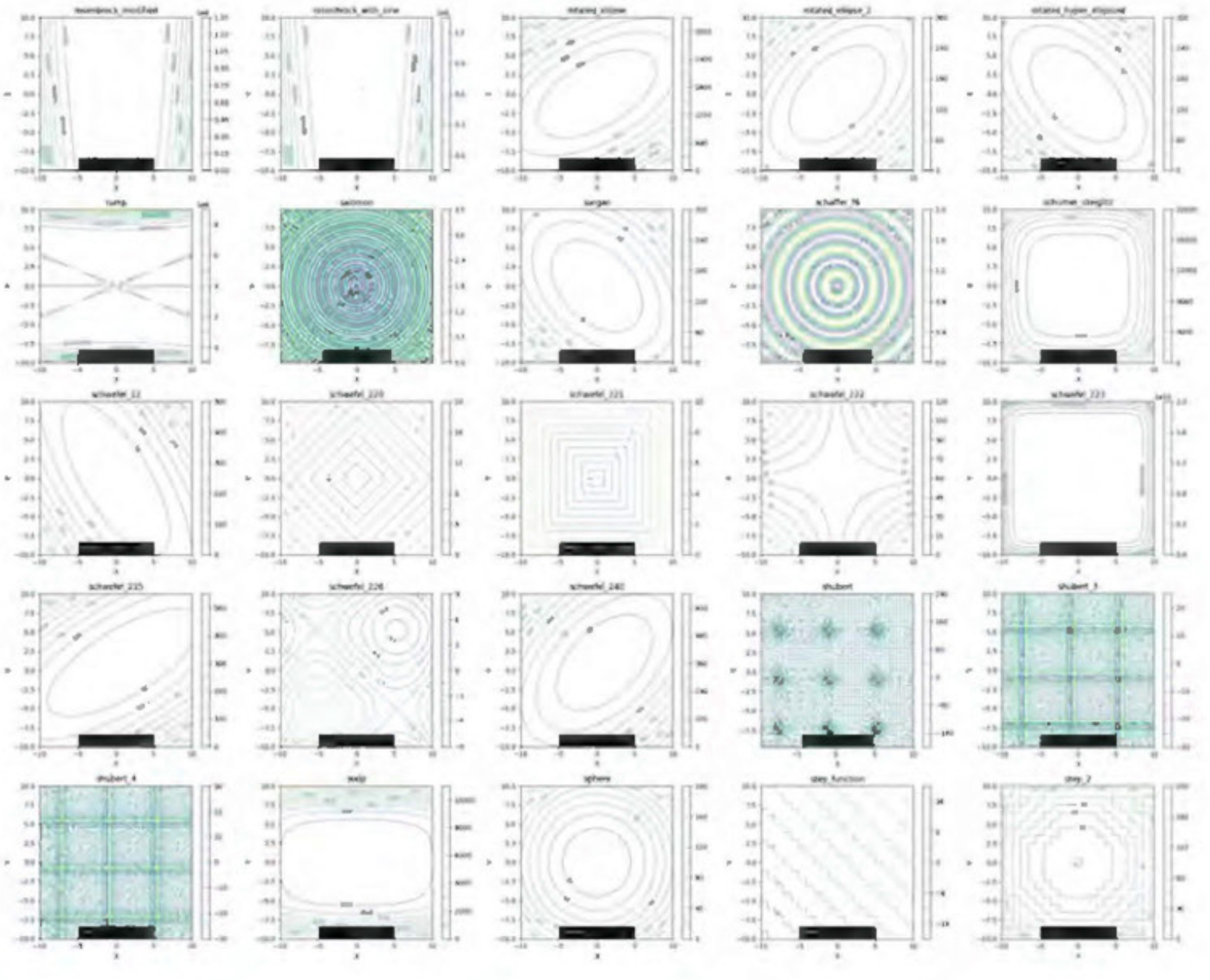

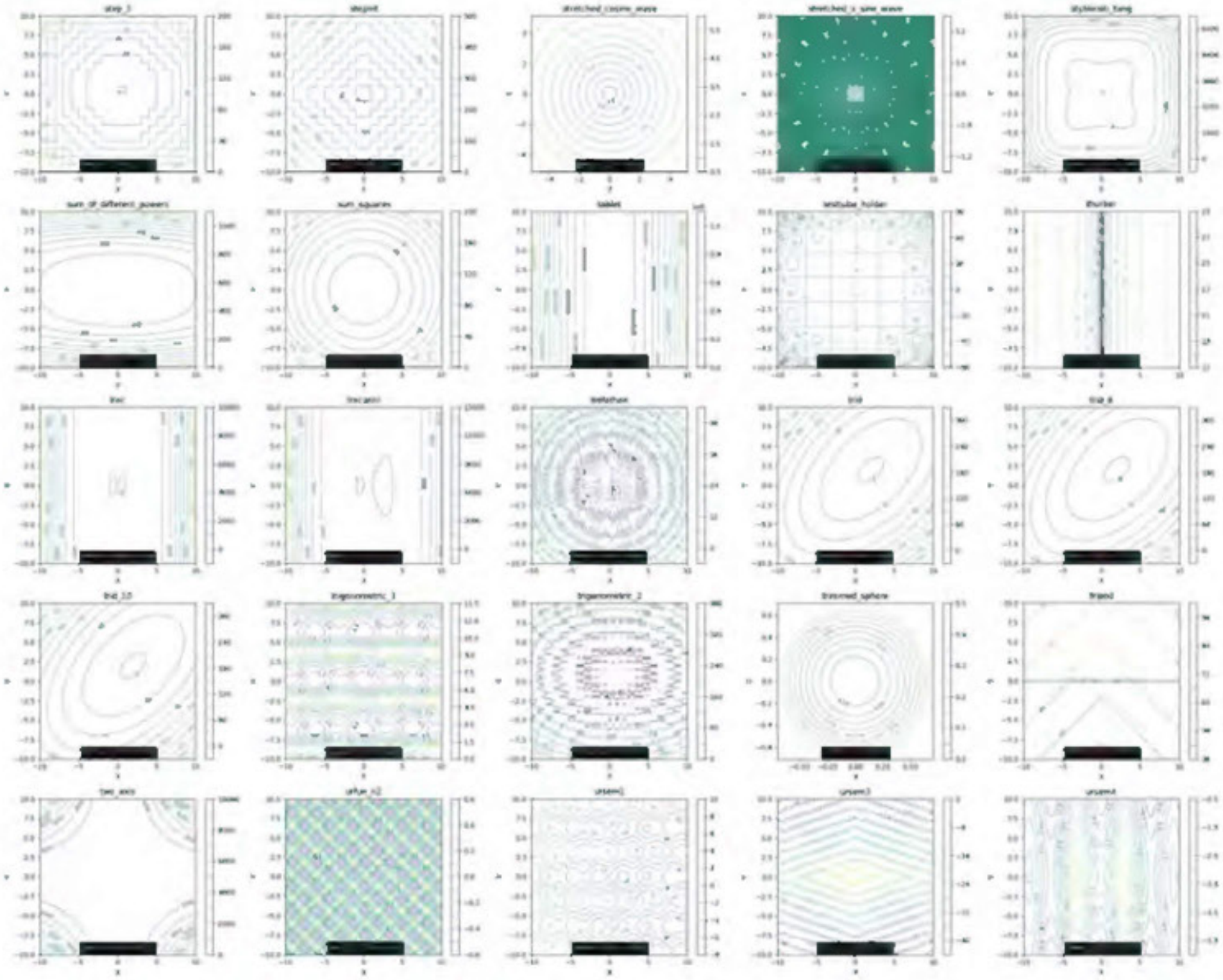

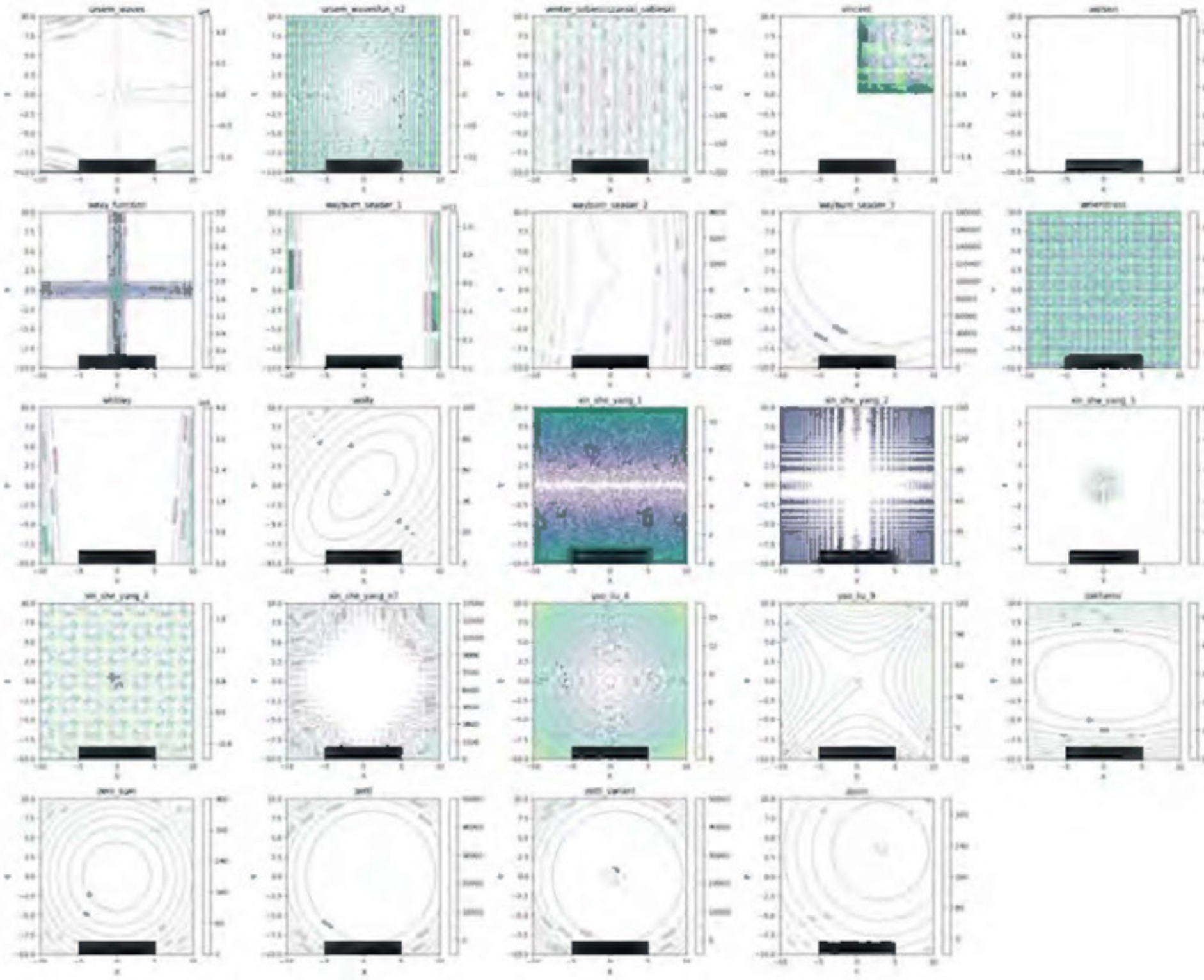

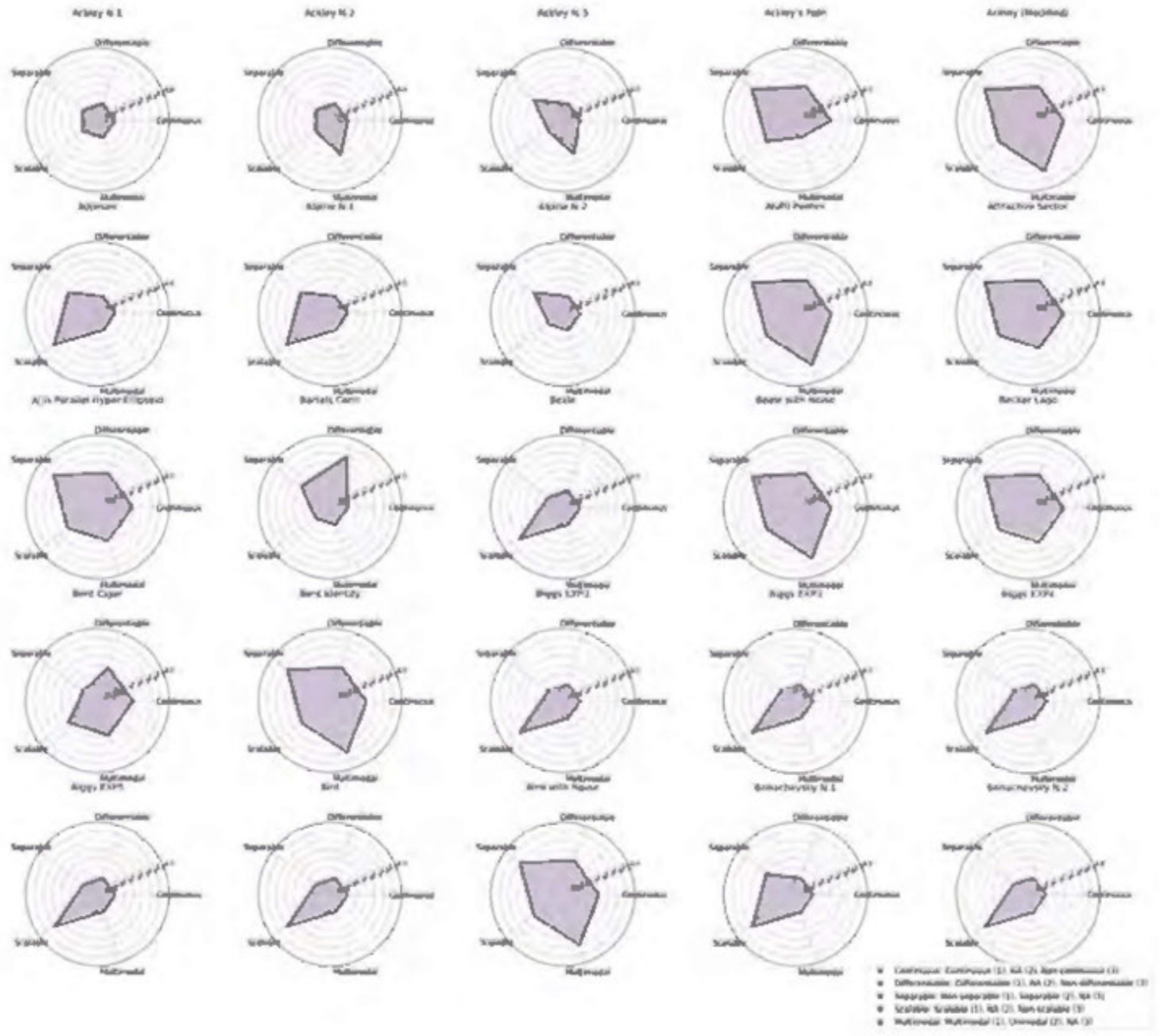

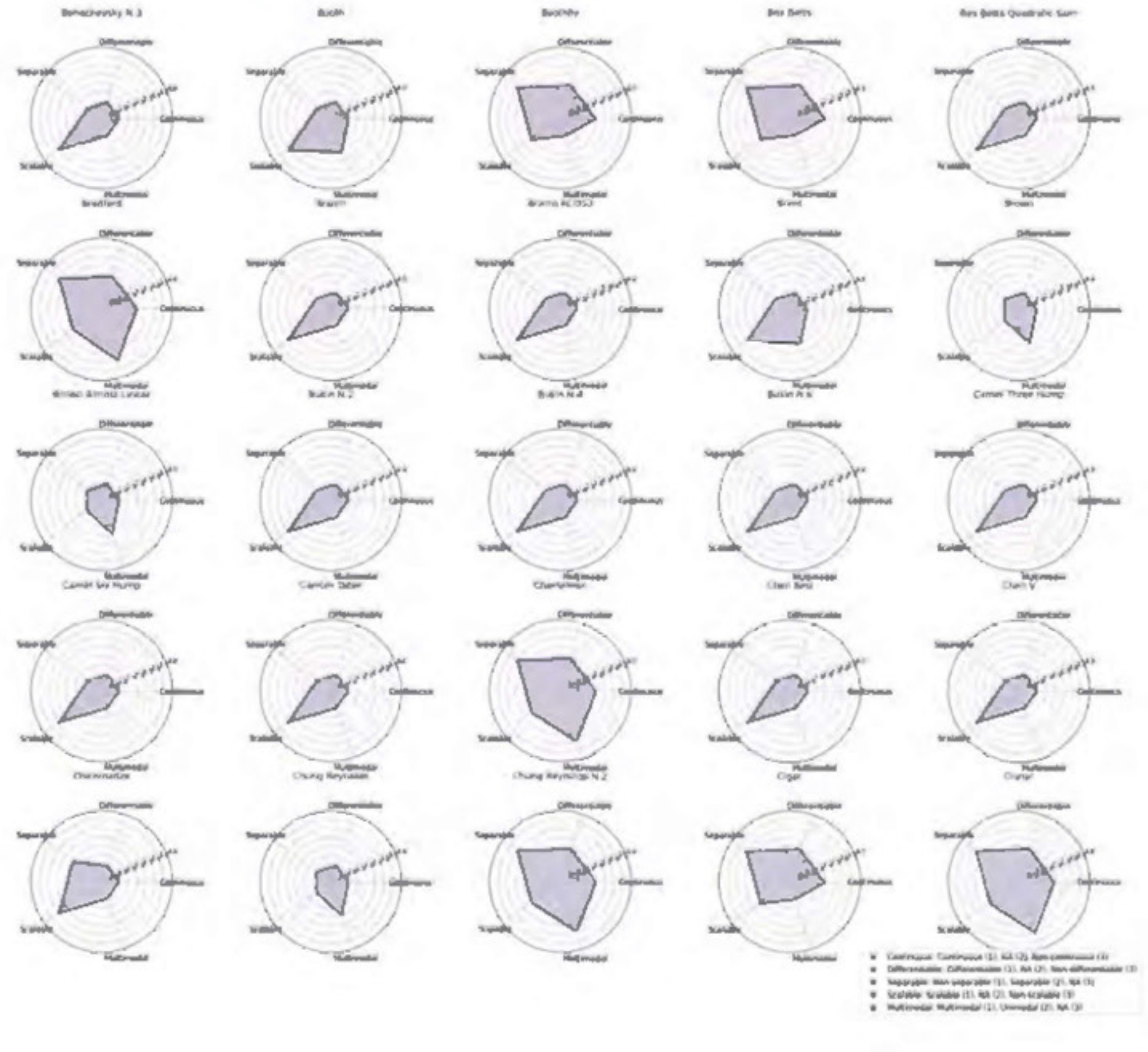

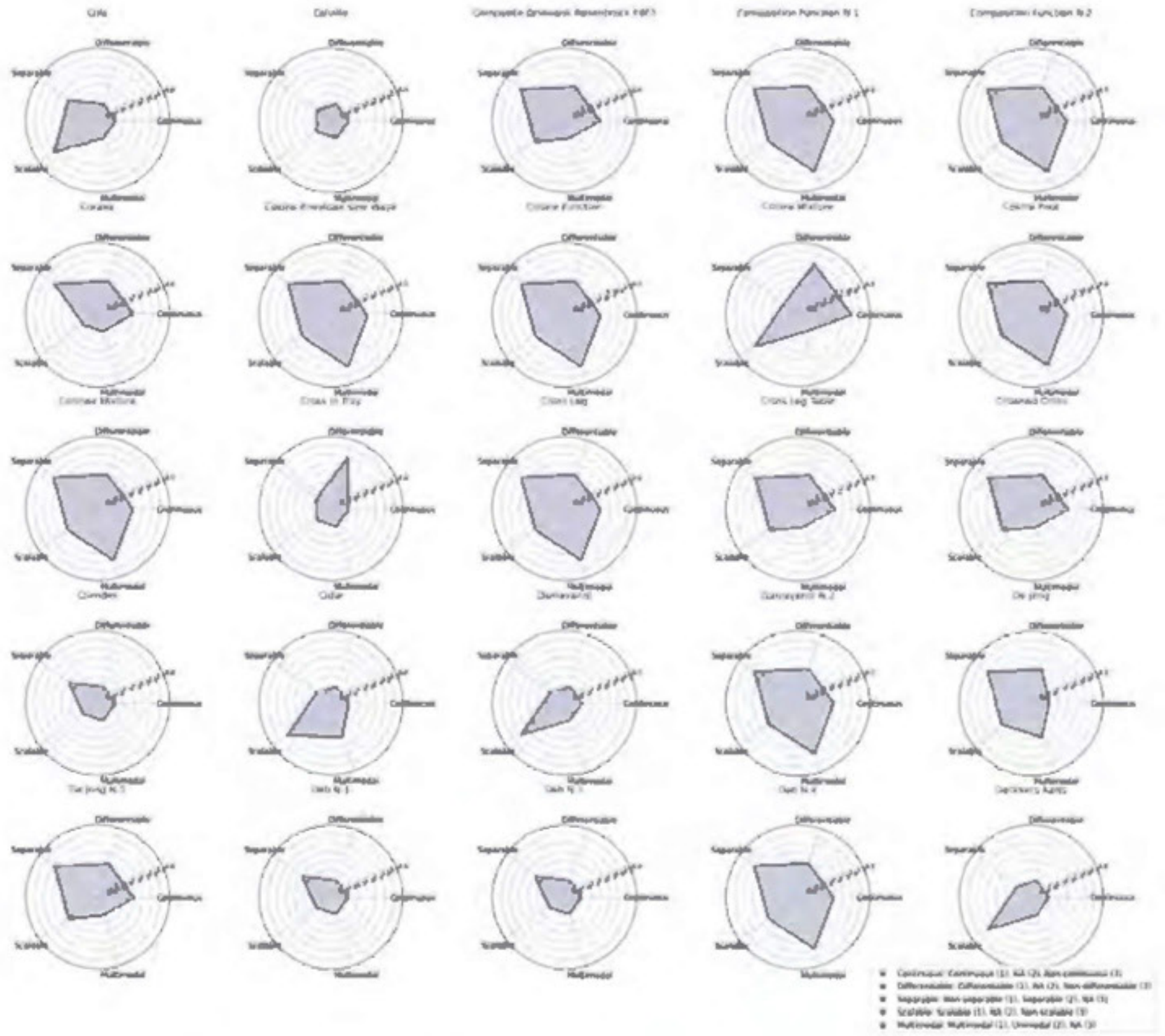

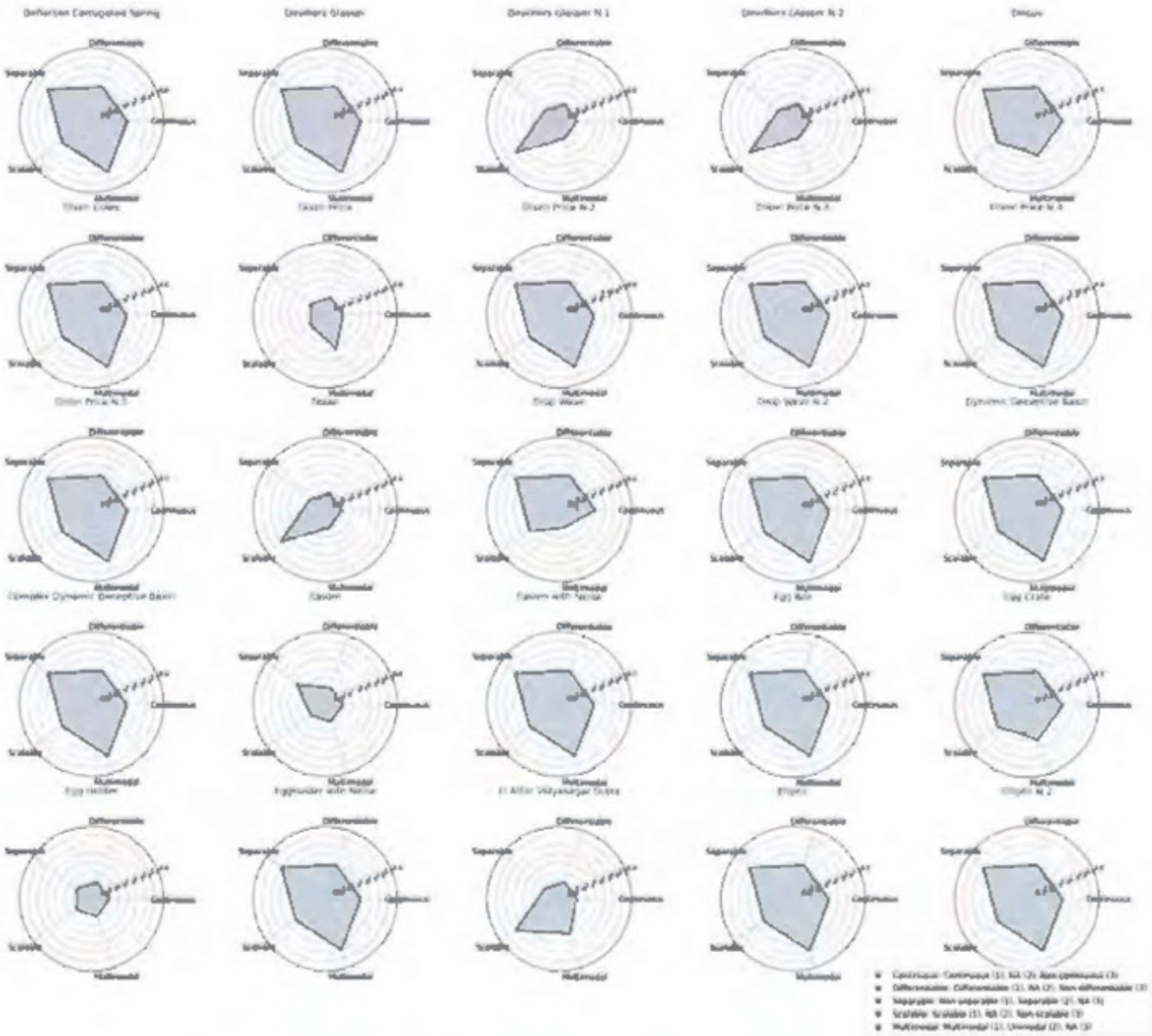
Differentiable
Separable
Continuous
Scalable
Multimodal

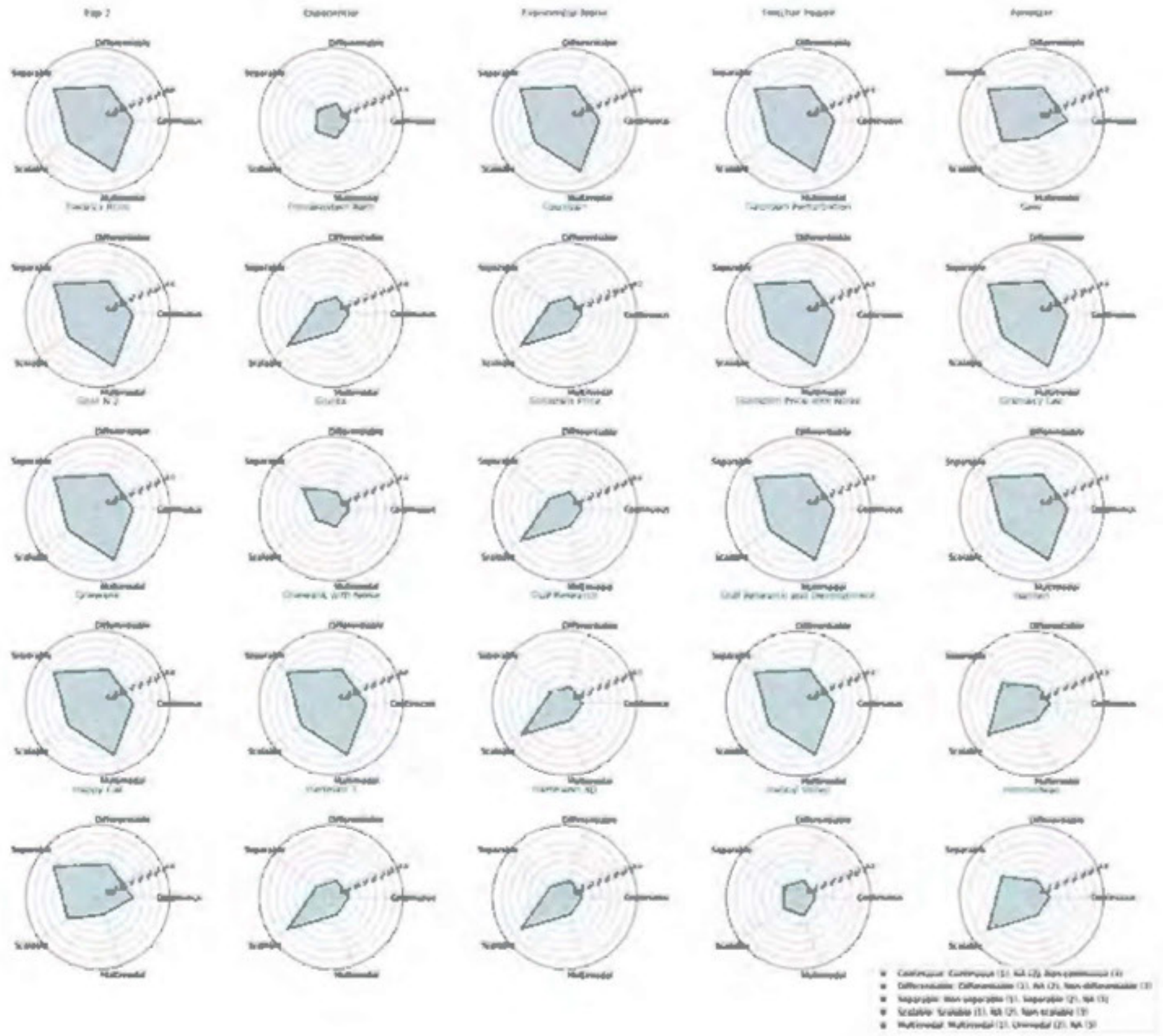

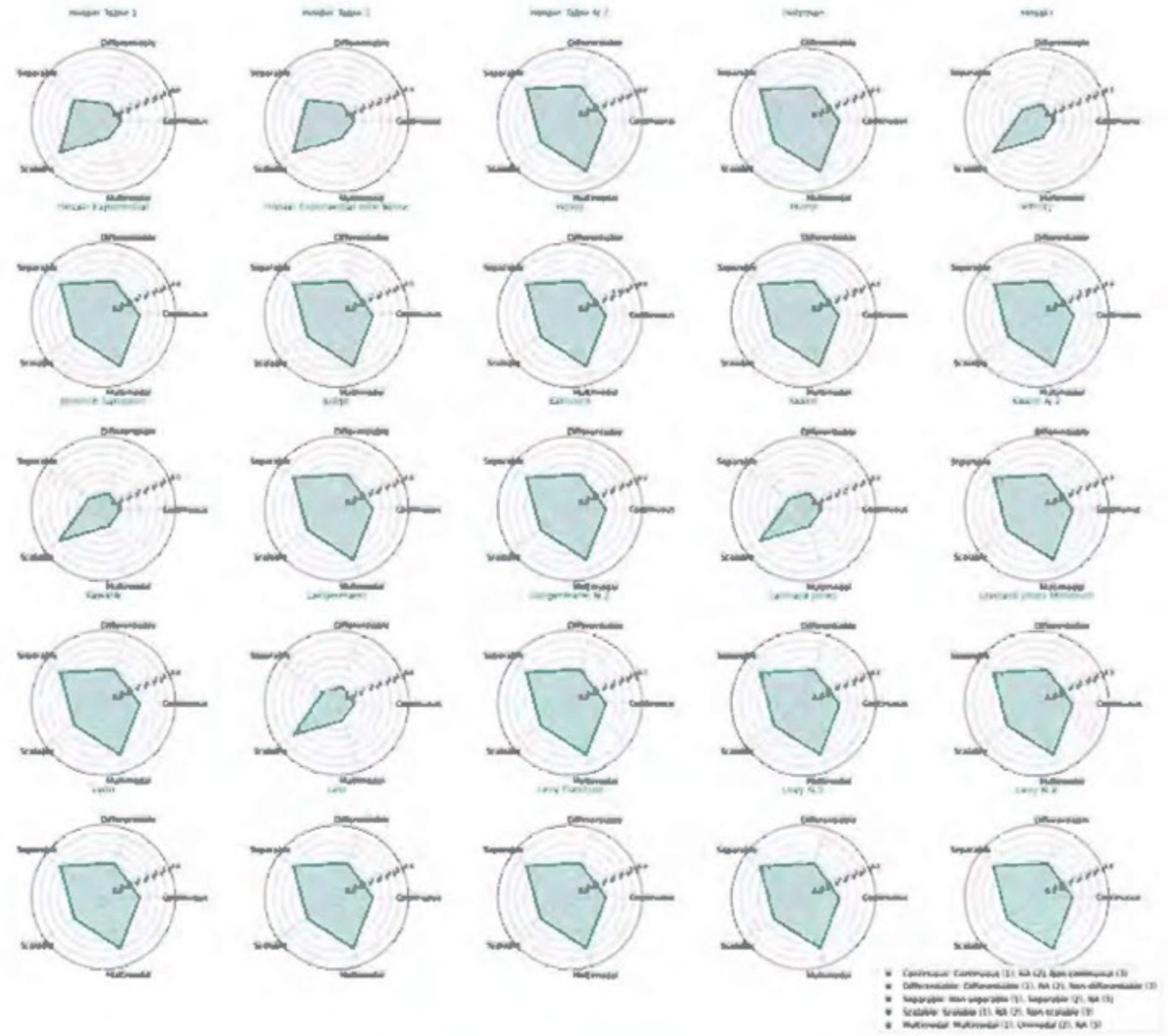

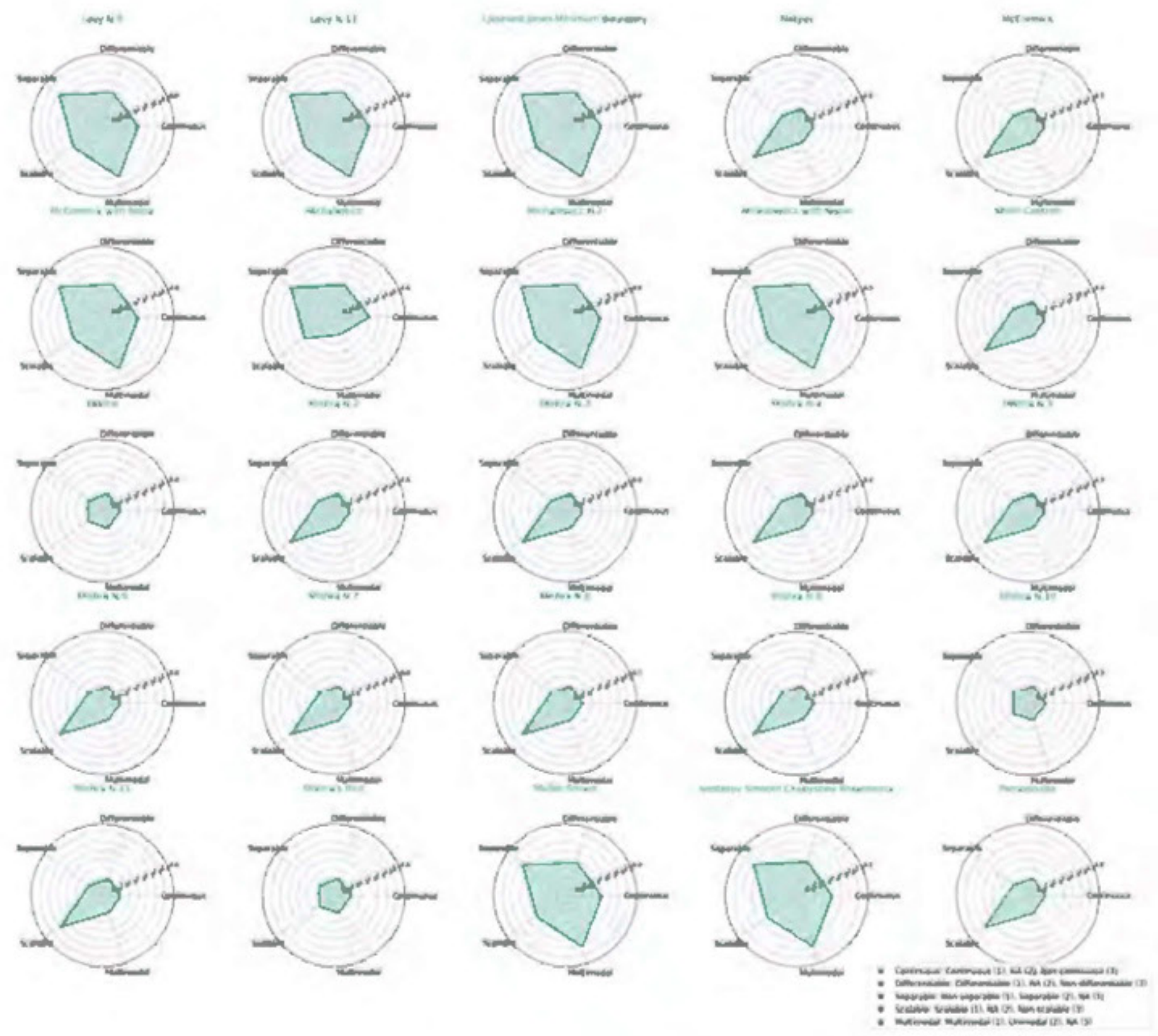
Differentiable
Separable
Continuous
Scalable
Multimodal

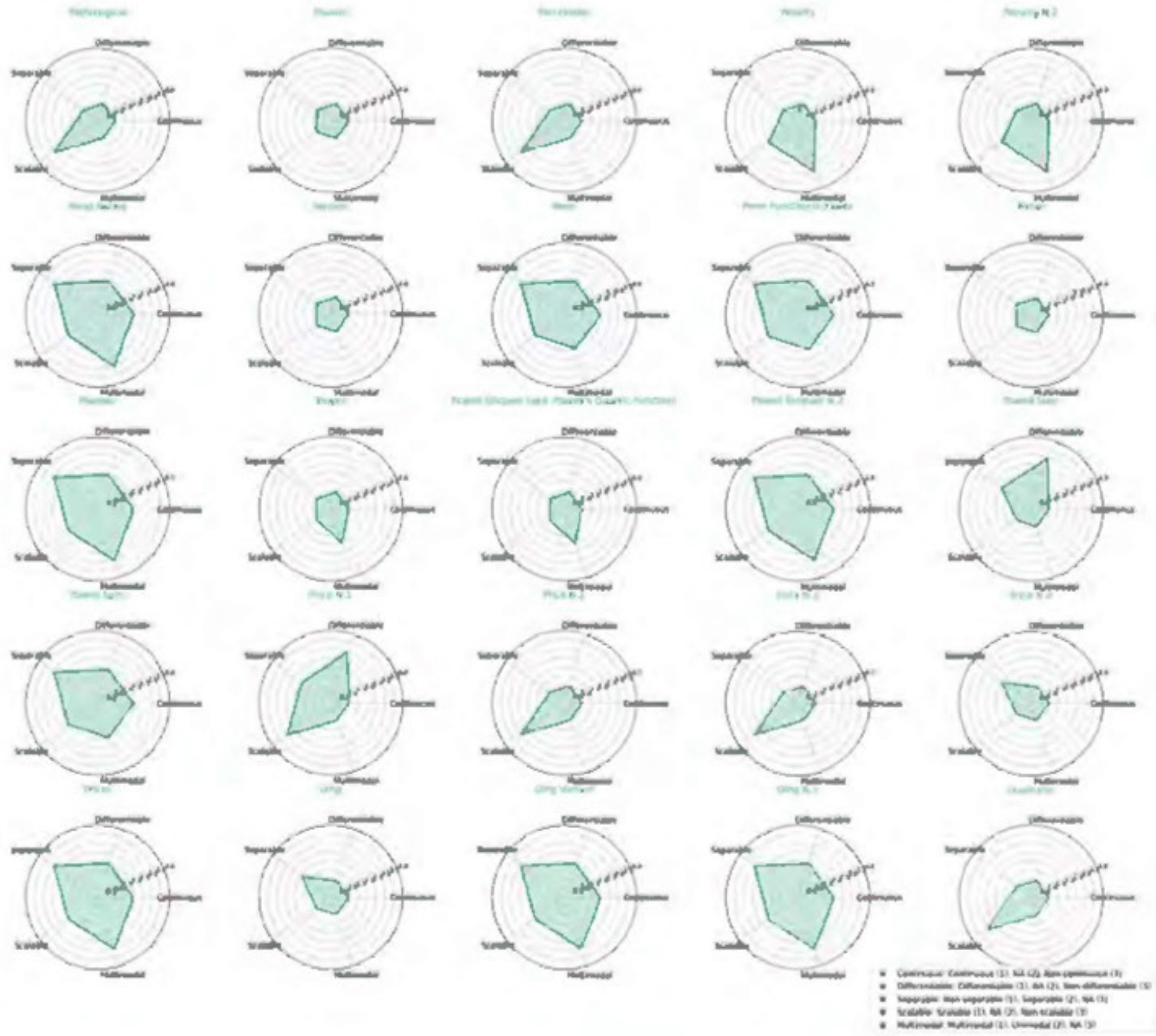
Continuous: Continuous (1), NA (2), Non-continuous (3)
Differentiable: Differentiable (1), NA (2), Non-differentiable (3)
Separable: Non-separable (1), Separable (2), NA (3)
Scalable: Scalable (1), NA (2), Non-scalable (3)
Multimodal: Multimodal (1), Unimodal (2), NA (3)

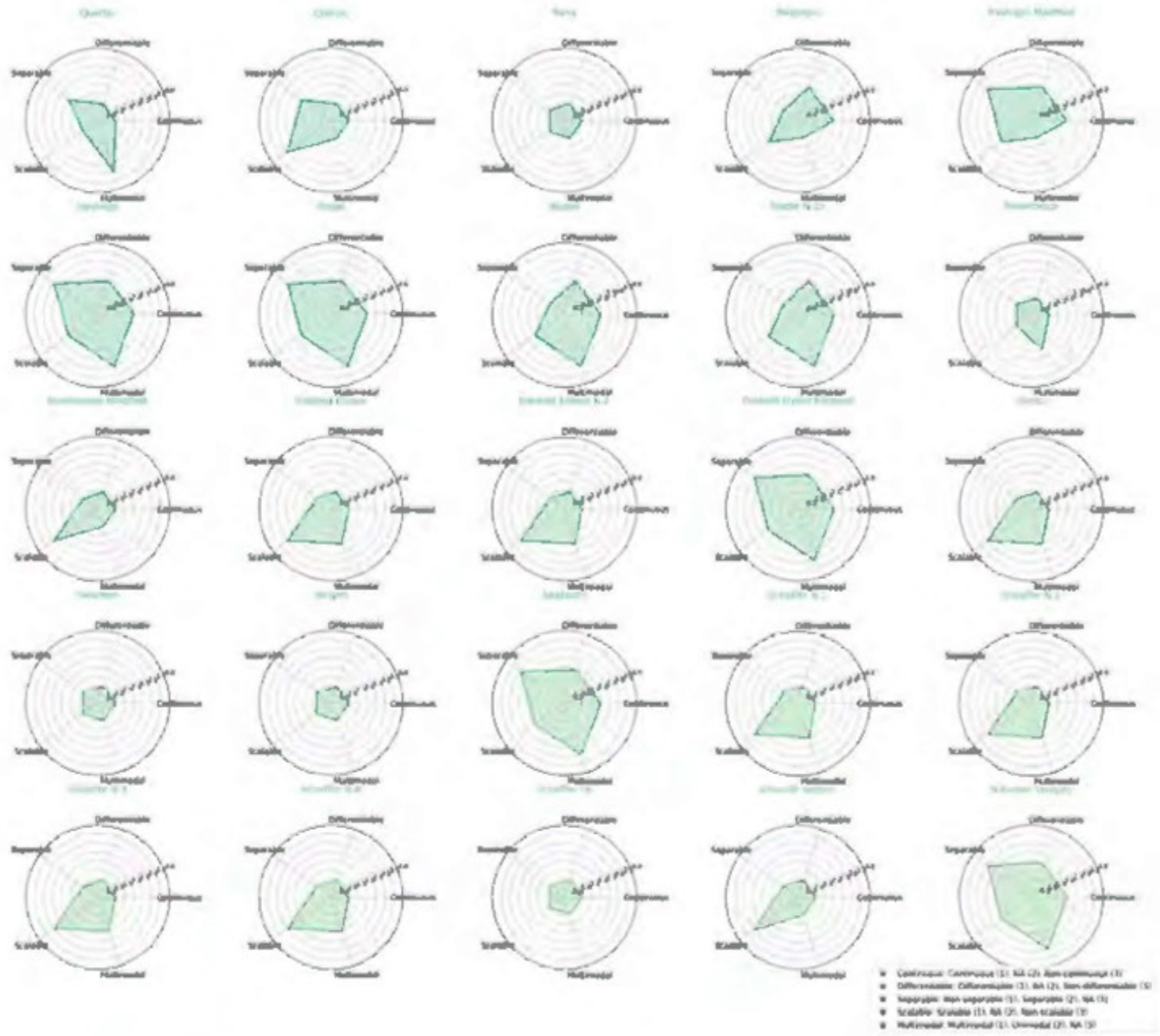

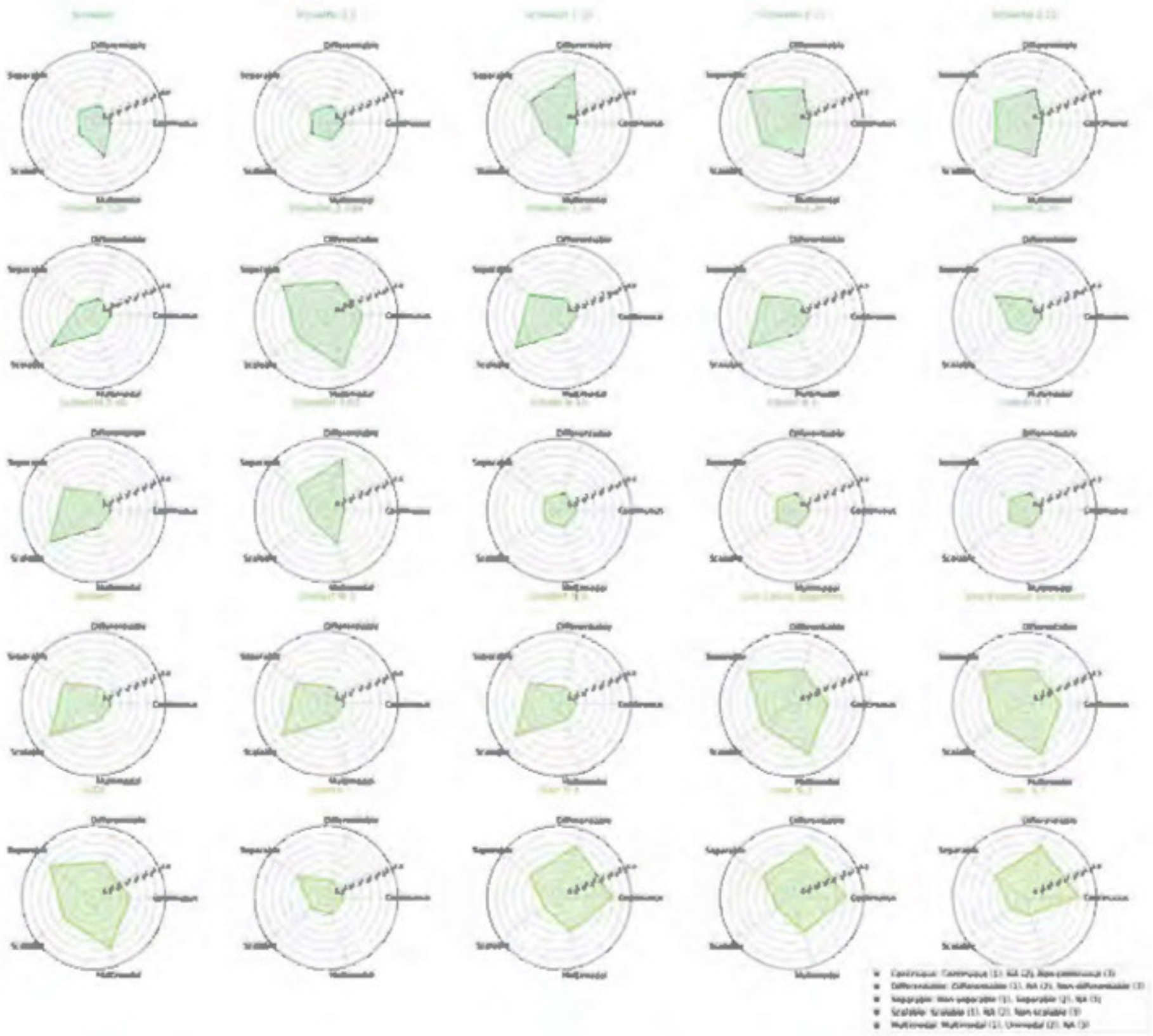
Differentiable
Separable
Continuous
Scalable
Multimodal

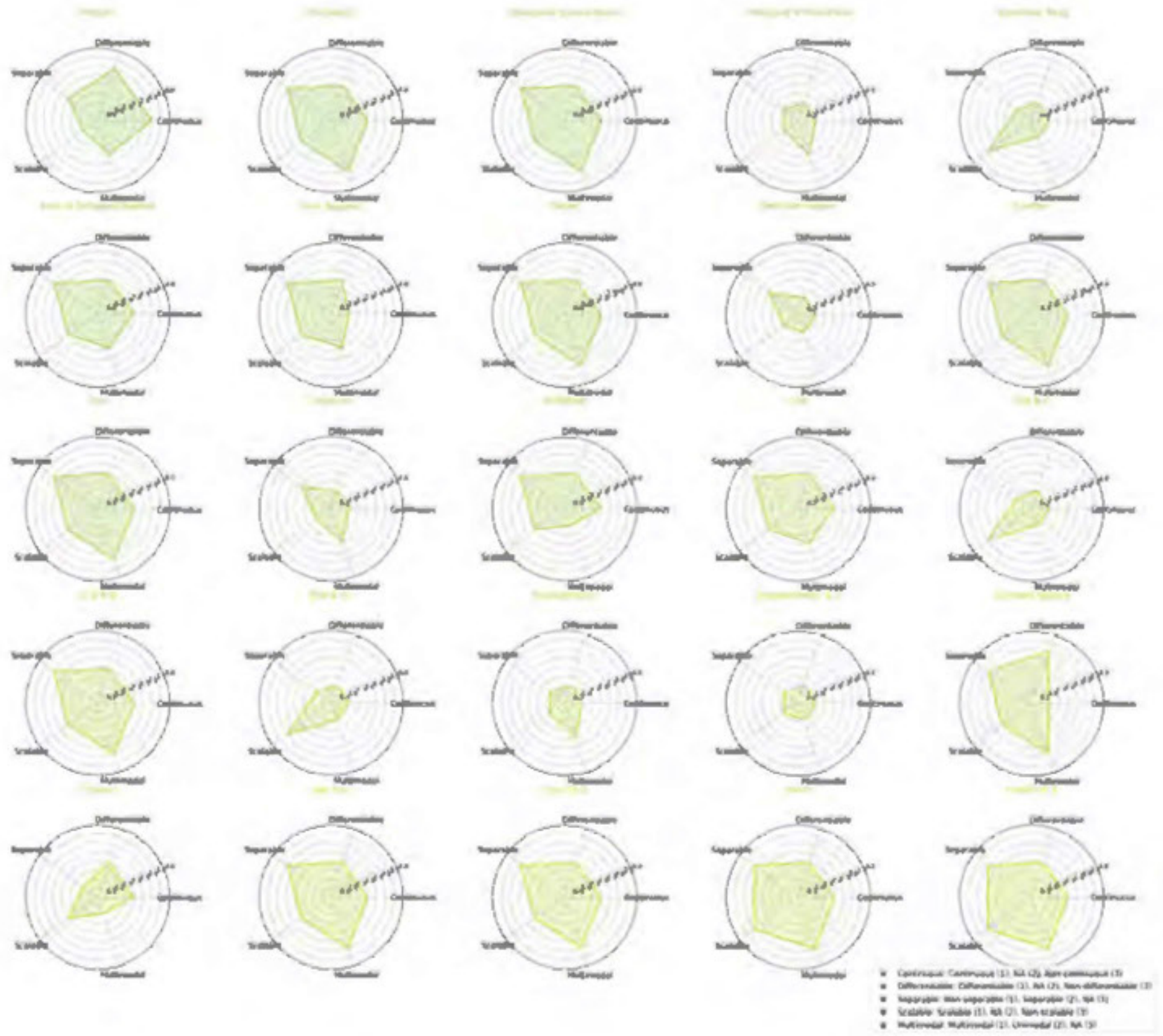
Differentiable
Separable
Continuous
Scalable
Multimodal
Continuous: Continuous (1), NA (2), Non-continuous (3)
Differentiable: Differentiable (1), NA (2), Non-differentiable (3)
Separable: Non-separable (1), Separable (2), NA (3)
Scalable: Scalable (1), NA (2), Non-scalable (3)
Multimodal: Multimodal (1), Unimodal (2), NA (3)

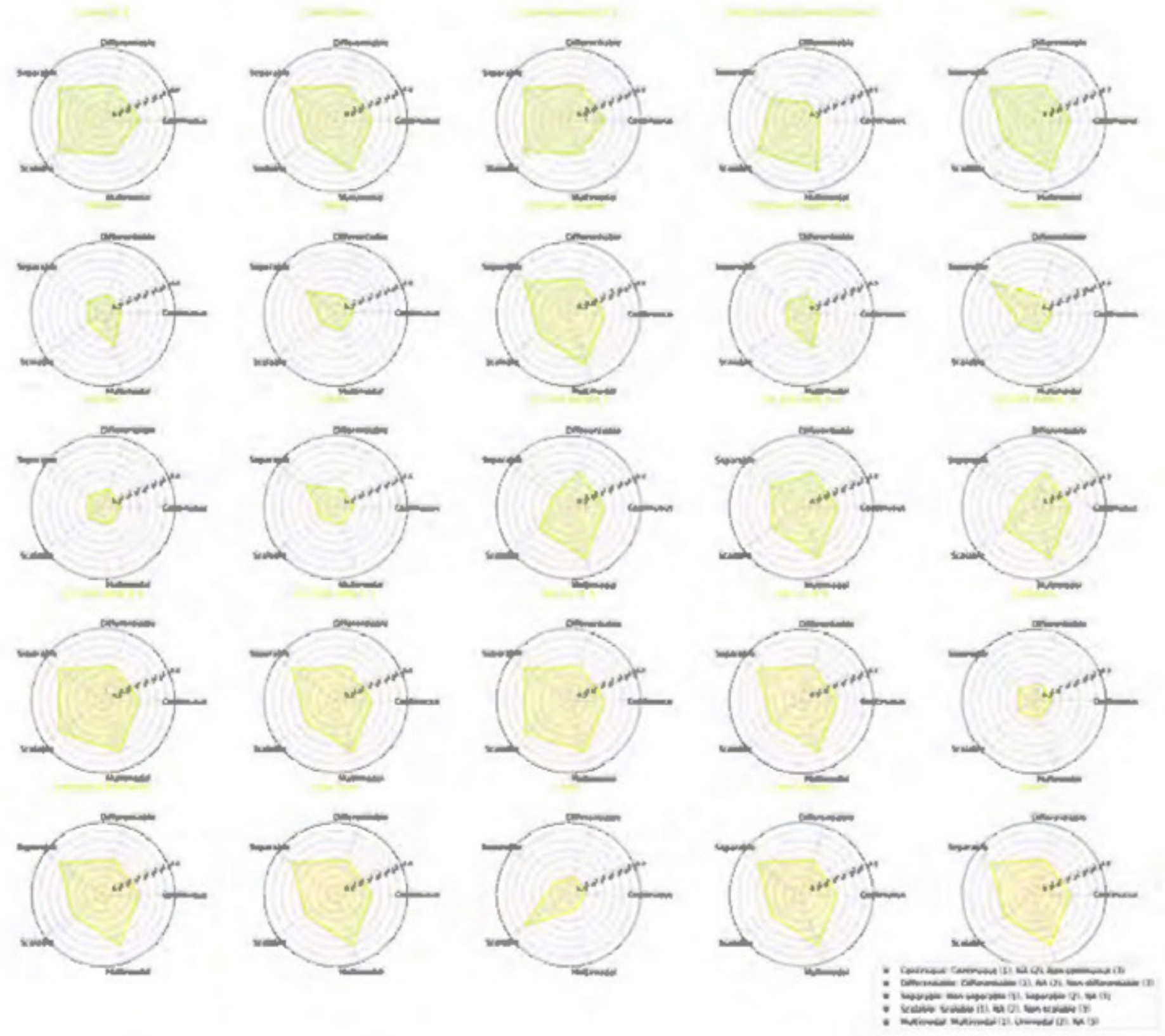
Differentiable
Continuous
Multimodal
Scalable
Separable
Continuous: Continuous (1), NA (2), Non-continuous (3)
Differentiable: Differentiable (1), NA (2), Non-differentiable (3)
Separable: Non-separable (1), Separable (2), NA (3)
Scalable: Scalable (1), NA (2), Non-scalable (3)
Multimodal: Multimodal (1), Unimodal (2), NA (3)